\documentclass[conference]{IEEEtran}
\IEEEoverridecommandlockouts
\usepackage{framed,multirow}
\usepackage{amssymb}
\usepackage{latexsym}
\usepackage{amsmath}
\usepackage{amsthm}
\usepackage{tabularx,booktabs}
\usepackage{tikz}
\usepackage[numbers]{natbib}
\usepackage[hidelinks,colorlinks=true,linkcolor=blue,citecolor=blue]{hyperref}
\usepackage{multicol}

\usepackage{siunitx}
\usepackage{tikz}
\usepackage{tabularx}
\usepackage{subfig}
\usetikzlibrary{arrows.meta}
\usetikzlibrary{shapes}
\newcolumntype{Y}{>{\centering\arraybackslash}X}
\usepackage[alpine]{ifsym}

\usetikzlibrary{decorations.pathmorphing}

\newtheoremstyle{mytheorem}
  {5pt}
  {3pt}
  {\itshape}
  {}
  {\itshape\bfseries}
  {.}
  {.5em}
  {\thmname{#1}\thmnumber{{ }#2}%
   \thmnote{ {\the\thm@notefont(#3)}}}
\makeatother
\theoremstyle{mytheorem}

\usepackage{url}
\usepackage{xcolor}
\definecolor{newcolor}{rgb}{.8,.349,.1}
\definecolor{royalazure}{rgb}{0, 0, 0}

\def\BibTeX{{\rm B\kern-.05em{\sc i\kern-.025em b}\kern-.08em
    T\kern-.1667em\lower.7ex\hbox{E}\kern-.125emX}}
    
\makeatletter
\newcommand*\titleheader[1]{\gdef\@titleheader{#1}}
\AtBeginDocument{%
  \let\st@red@title\@title
  \def\@title{%
    \bgroup\normalfont\large\centering\@titleheader\par\egroup
    \vskip1.5em\st@red@title}
}

\title{Investigating the significance of adversarial attacks and their relation to interpretability for\\radar-based human activity recognition systems}
\titleheader{}

\author{
\IEEEauthorblockN{Utku Ozbulak$^{1,\,5,*}$ \quad Baptist Vandersmissen$^{1,*}$ \quad Azarakhsh Jalalvand$^{1,\,4}$  \quad Ivo Couckuyt$^{2}$ \\ \\ \quad Arnout Van Messem$^{3,\,5}$ \quad Wesley De Neve$^{1,\,5}$}\\
\IEEEauthorblockA{\small\textit{$^1$Department of Electronics and Information Systems, Ghent University, Belgium} \\
\textit{$^2$Department of Information Technology, Ghent University--imec, Belgium} \\
\textit{$^3$Department of Applied Mathematics, Computer Science and Statistics, Ghent University, Belgium}\\
\textit{$^4$Department of Mechanical and Aerospace Engineering, Princeton University, USA}\\
\textit{$^5$Center for Biotech Data Science, Ghent University Global Campus, Republic of Korea}}
\thanks{{* Equal contribution.}}
\thanks{{Corresponding author: Utku Ozbulak\,\textendash\,utku.ozbulak@ugent.be.}}
\thanks{{Preprint. Accepted for publication on Computer Vision and Image Understanding, Special issue on Adversarial Deep Learning in Biometrics \& Forensics, Elsevier, 2020.}}
\thanks{{DOI:} \href{https://doi.org/10.1016/j.cviu.2020.103111}{https://doi.org/10.1016/j.cviu.2020.103111}.}
}

\begin{document}
\maketitle

\begin{abstract}
Given their substantial success in addressing a wide range of computer vision challenges, Convolutional Neural Networks (CNNs) are increasingly being used in smart home applications, with many of these applications relying on the automatic recognition of human activities. In this context, low-power radar devices have recently gained in popularity as recording sensors, given that the usage of these devices allows mitigating a number of privacy concerns, a key issue when making use of conventional video cameras. Another concern that is often cited when designing smart home applications is the resilience of these applications against cyberattacks. 
It is, for instance, well-known that the combination of images and CNNs is vulnerable against adversarial examples, mischievous data points that force machine learning models to generate wrong classifications during testing time.
In this paper, we investigate the vulnerability of radar-based CNNs to adversarial attacks, and where these radar-based CNNs have been designed to recognize human gestures. Through experiments with four unique threat models, we show that radar-based CNNs are susceptible to both white- and black-box adversarial attacks. 
We also expose the existence of an extreme adversarial attack case, where it is possible to change the prediction made by the radar-based CNNs by only perturbing the padding of the inputs, without touching the frames where the action itself occurs. Moreover, we observe that gradient-based attacks exercise perturbation not randomly, but on important features of the input data. We highlight these important features by making use of Grad-CAM, a popular neural network interpretability method, hereby showing the connection between adversarial perturbation and prediction interpretability.
\end{abstract}

\section{Introduction}
\label{Introduction}
Recent advancements in the field of computer vision, natural language processing, and audio analysis enabled the deployment of intelligent systems in homes in the form of assistive technologies. These so-called \textit{smart homes} come with a wide range of functionality such as voice- and gesture-controlled appliances, security systems, and health-related applications. Naturally, multiple sensors are needed in these smart homes to capture the actions performed by household residents and to act upon them.

\textbf{Sensors for smart home applications}\,\textemdash\,Microphones and video cameras are currently two of the most commonly used sensors in smart homes. The research in the domain of video-oriented computer vision is extensive, and the combined usage of a video camera and computer vision enables a wide range of assistive technologies, including applications related to security (e.g., intruder detection) and applications incorporating gesture-controlled functionalities~\citep{video_camera_is_strong}. However, one of the major drawbacks of using video cameras is their privacy intrusiveness~\citep{rajpoot2015video}. These privacy-related concerns are, at an increasing rate, being covered by media articles~\citep{media-privacy}. Furthermore, a largely overlooked aspect of video-assisted technologies in smart homes is that video cameras are able to capture both smart home residents and visitors. Therefore, residents of smart homes need to be aware of the statutory restrictions on privacy invasion.

Low-power radar devices, as complementary sensors, are capable of alleviating the privacy concerns raised over the usage of video cameras. In that regard, the main advantages of radar devices over video cameras are as follows: (1) better privacy preservation, (2) a higher effectiveness in poor capturing conditions (e.g., low light, presence of smoke), and (3) through-the-wall sensing~\citep{through_wall_sensing}.

\begin{figure}[t]
\centering
\rotatebox[origin=l]{90}{\phantom{--}Frame: 1}
\subfloat{{{\includegraphics[width=2.45cm, height=1.60cm]{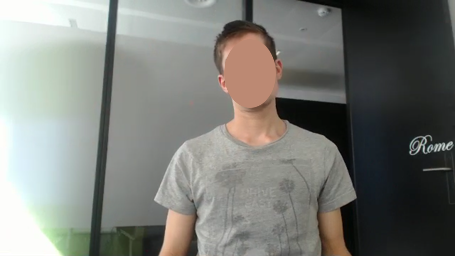}}}}\quad
\subfloat{{{\includegraphics[width=2.45cm, height=1.60cm]{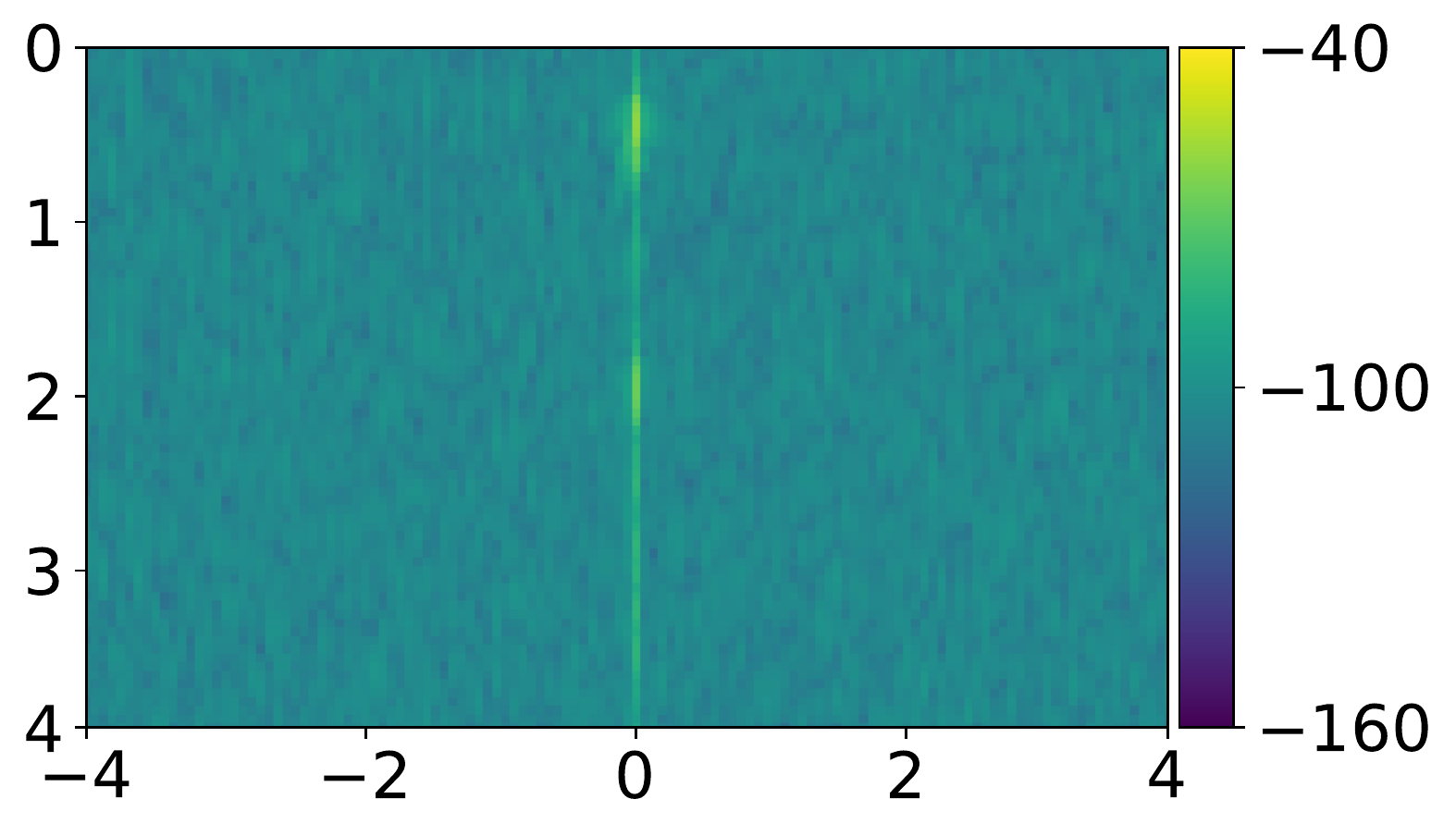}}}}\quad
\subfloat{{{\includegraphics[width=2.45cm, height=1.60cm]{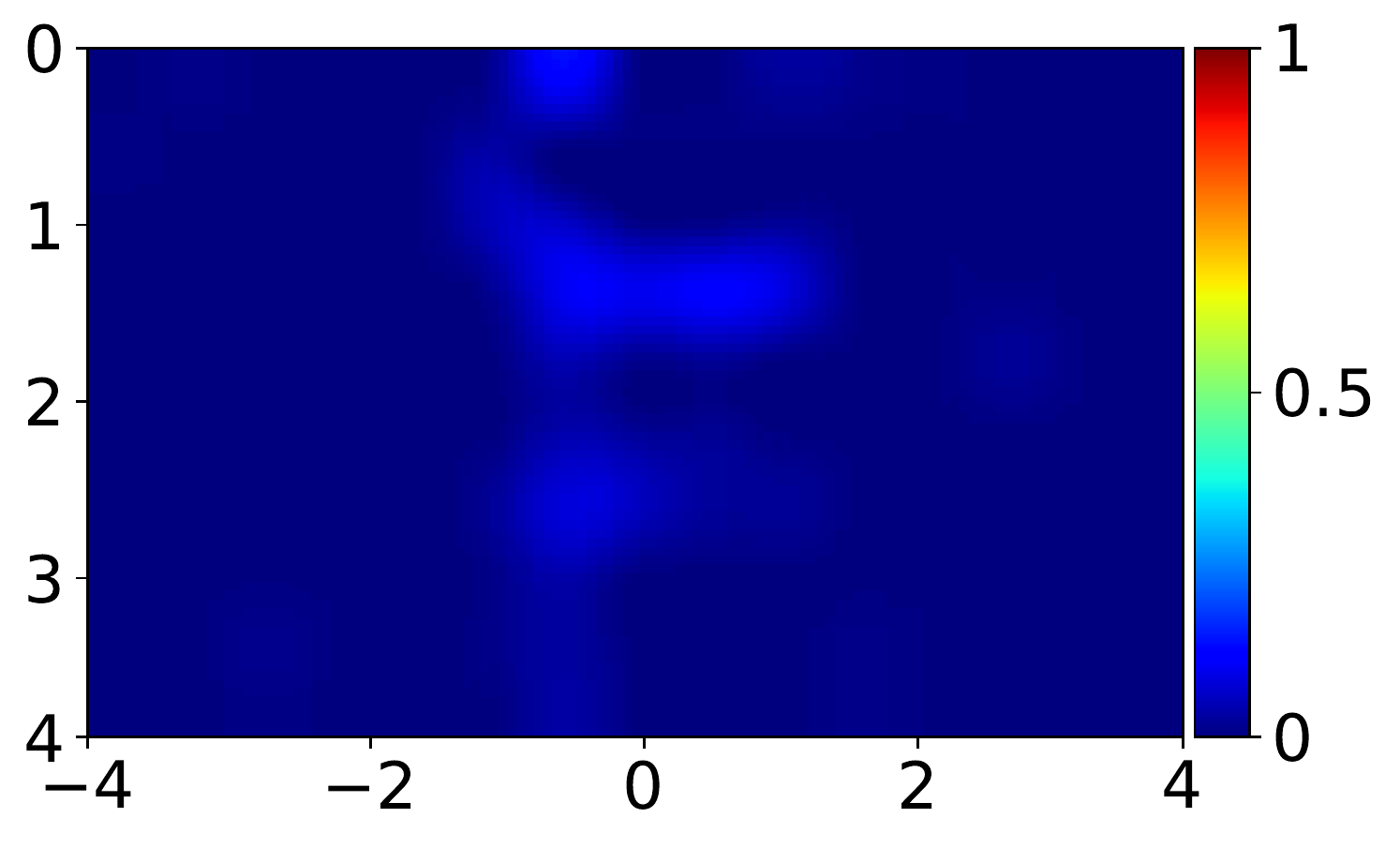}}}}
\vspace{1em}
\\
\rotatebox[origin=l]{90}{\phantom{-}Frame: 10}
\subfloat{{{\includegraphics[width=2.45cm, height=1.6cm]{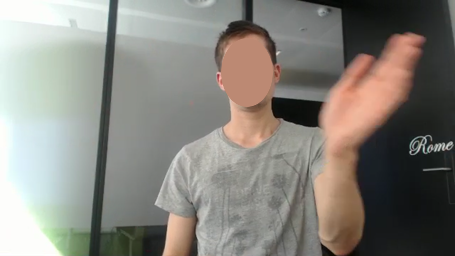}}}}\quad
\subfloat{{{\includegraphics[width=2.45cm, height=1.6cm]{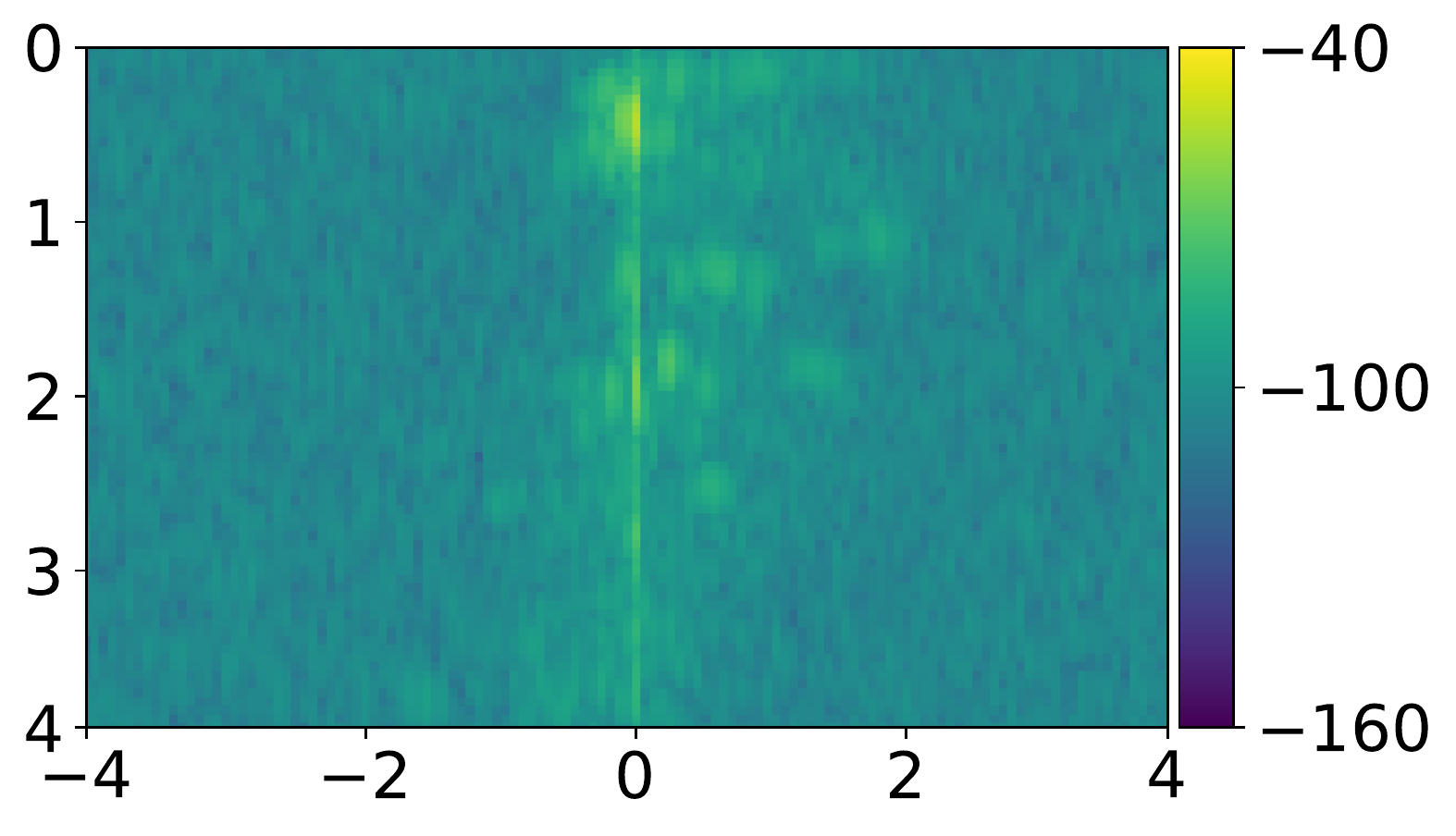}}}}\quad
\subfloat{{{\includegraphics[width=2.45cm, height=1.6cm]{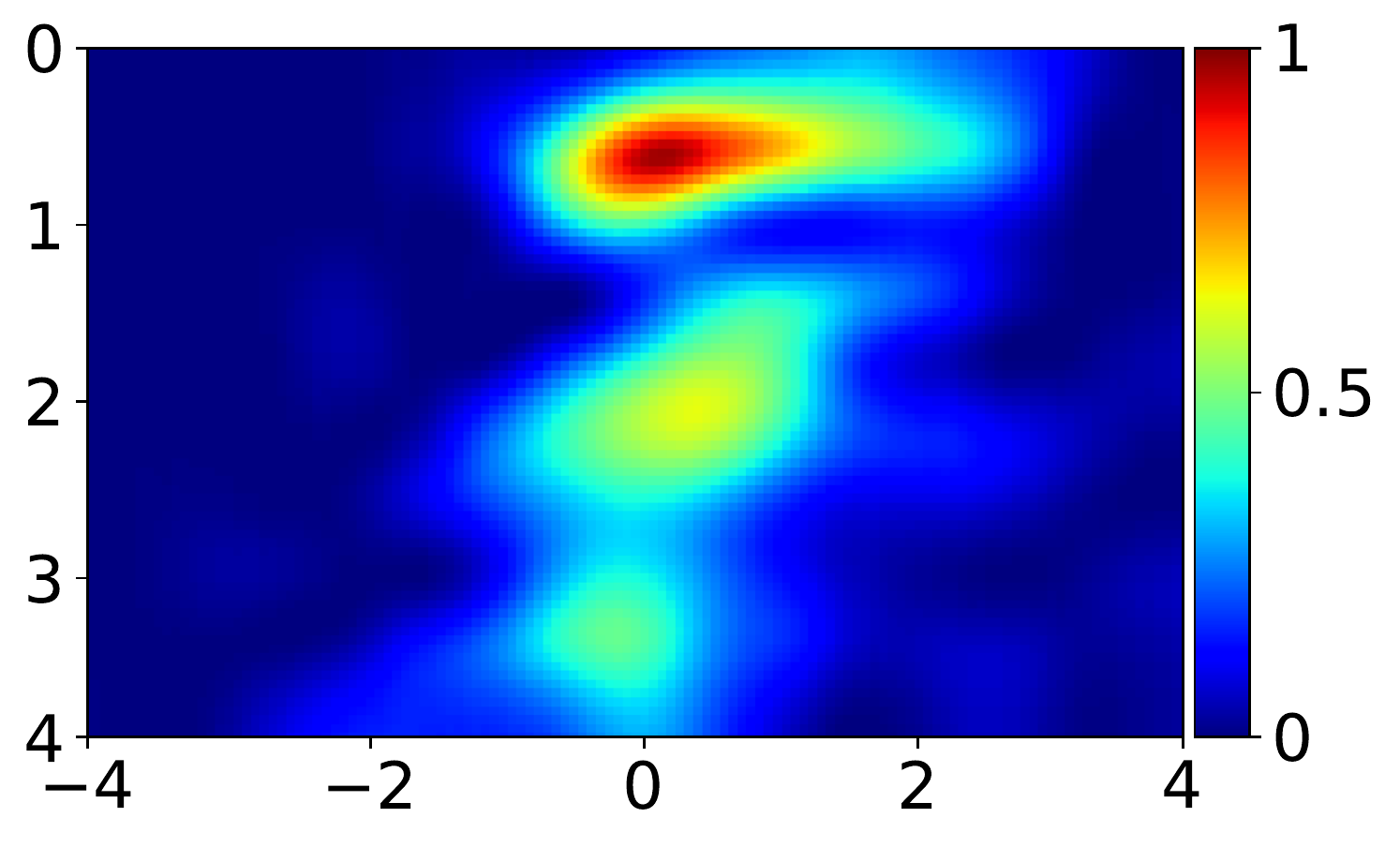}}}}
\\
\vspace{1em}
\rotatebox[origin=l]{90}{\phantom{-}Frame: 13}
\subfloat{{{\includegraphics[width=2.45cm, height=1.6cm]{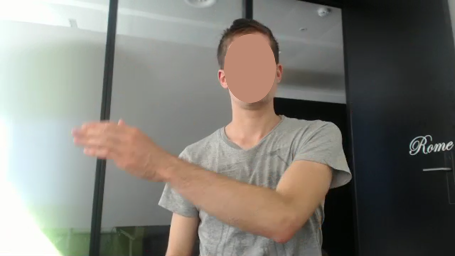}}}}\quad
\subfloat{{{\includegraphics[width=2.45cm, height=1.6cm]{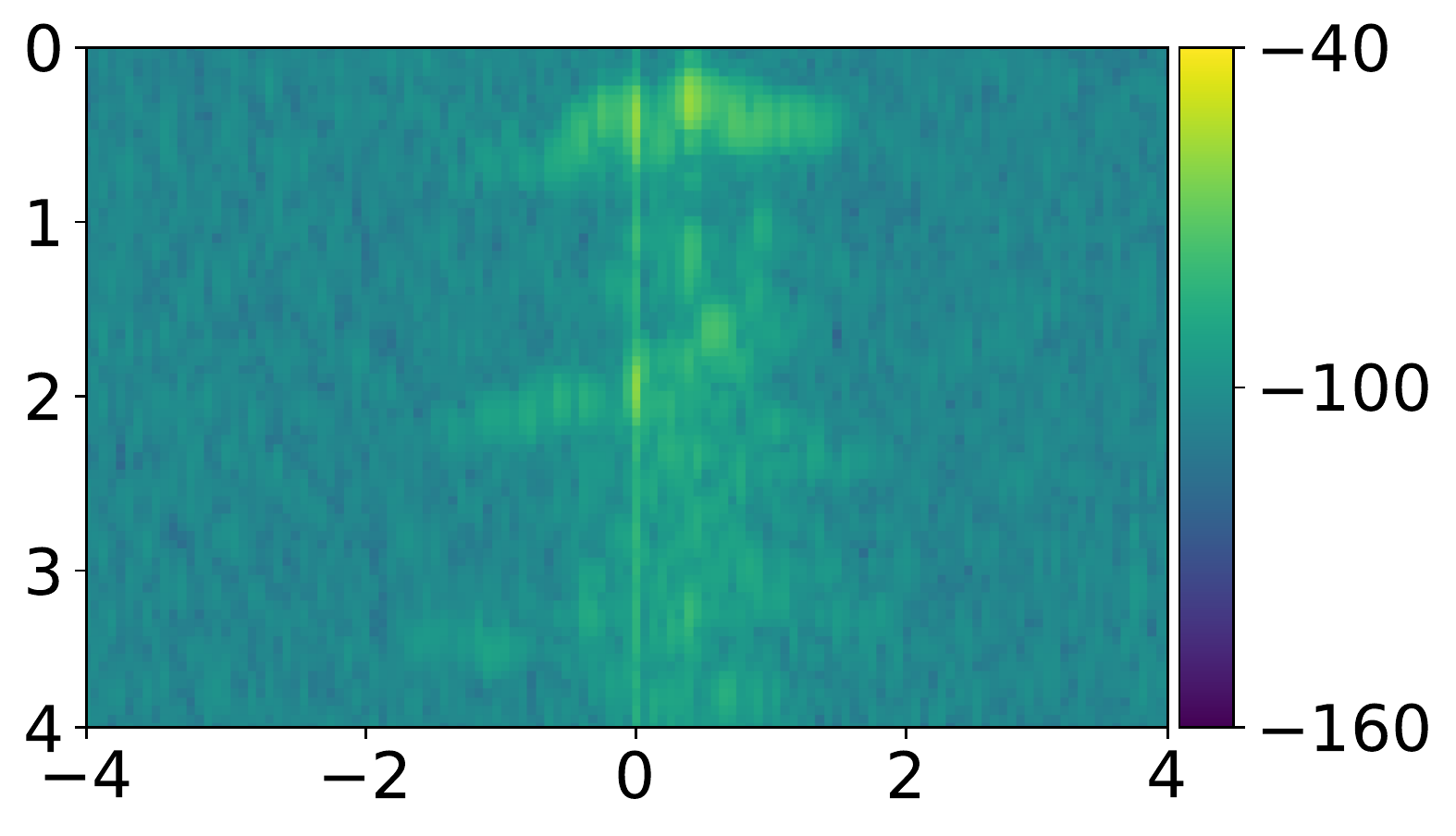}}}}\quad
\subfloat{{{\includegraphics[width=2.45cm, height=1.6cm]{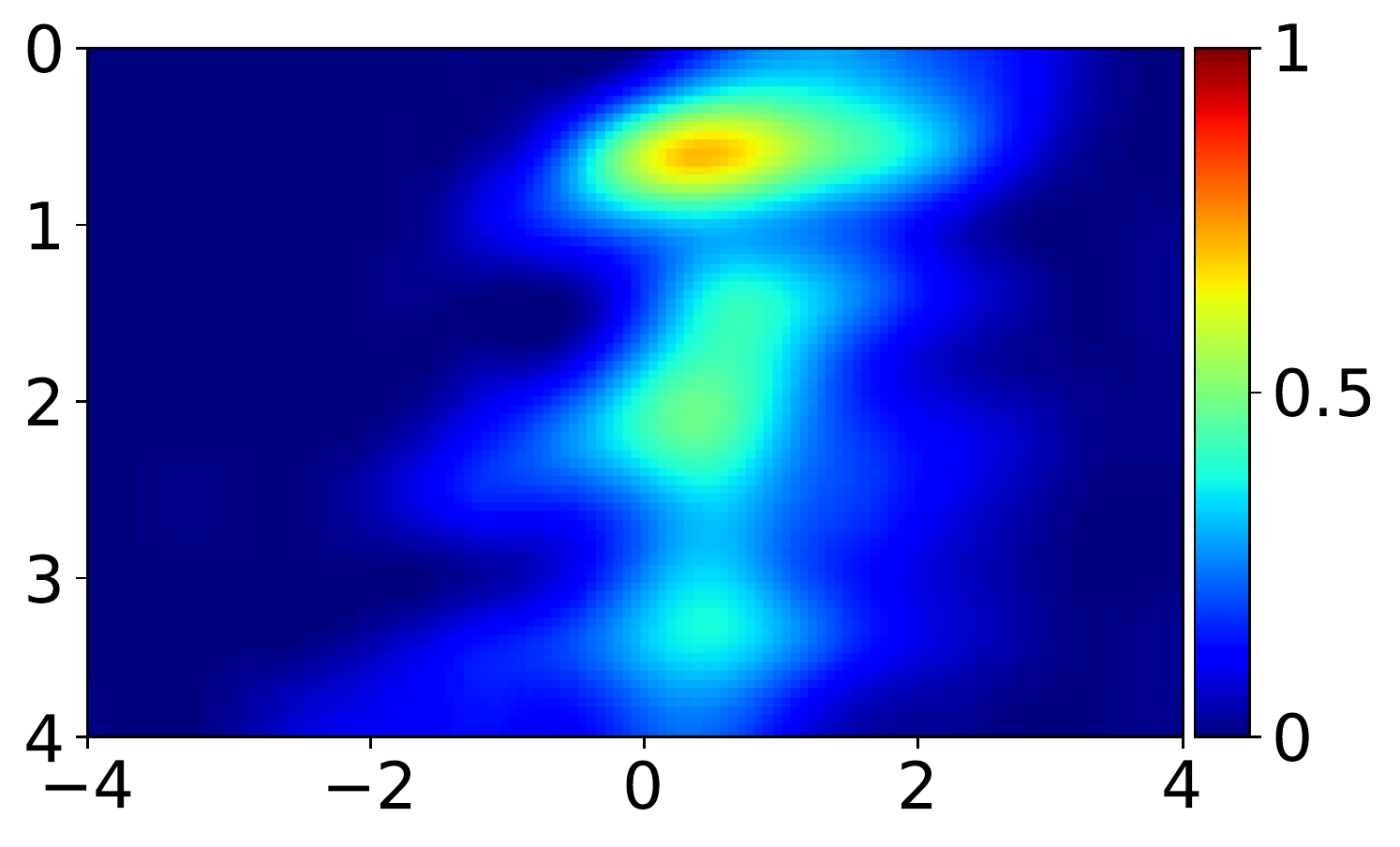}}}}

\caption{(1) RD frames, (2) video frames, and (3) Grad-CAM heatmaps for the action \textit{swipe left}. The $X$-axis and $Y$-axis of the RD frames and the Grad-CAM images, which are omitted for visual clarity, correspond to range and velocity, respectively.}
\label{fig:gracam_swipe_left}
\centering
\end{figure}

Frequency-modulated continuous-wave (FMCW) radars capture the environment by transmitting an electromagnetic signal over a certain line-of-sight. The reflections of this transmitted signal are then picked up by one or more receiving antennas and converted into range-Doppler (RD) and micro-Doppler (MD) frames~\citep{2014Chen}. These frames contain velocity and range information about all the objects in the line-of-sight (for the duration of the recording). Recent studies show that with the help of (deep) neural networks, it is possible to leverage these RD and MD frames to recognize multiple individuals~\citep{ Reservoir_Radar2019,vandersmissen2018indoor} or to detect human activities with high precision~\citep{gesture_rec_radar}. The aforementioned studies represent these RD and MD frames in the form of a sequence of mono-color images which are supplied as an input to deep CNNs. Three example RD frames and their corresponding video frames for the gesture \textit{swiping left} are given in Fig.~\ref{fig:gracam_swipe_left}.

\textbf{Convolutional neural networks}\,\textemdash\,Even though CNNs were applied extensively in the work of~\citet{lecun1998gradient} on the MNIST data set, they only became popular after some revolutionary results were obtained in the study of~\citet{Alexnet} on the ImageNet data set~\citep{ILSVRC15:rus}. That work was further improved by many researchers, showing the efficacy of deep CNNs~\citep{resnet,VGG,inceptionv3}. One of the main benefits of CNNs is their ability to automatically learn features, thus making it possible to forgo the cumbersome process of having to define hand-engineered expert features. Since CNN architectures are end-to-end differentiable, the features can be learned by an optimization method such as gradient descent. This property allows CNNs to be applied to various types of data beyond images and video sequences, such as text (i.e., natural language processing), speech, and radar data~\citep{graves2013speech, md_hand_gesture}.

Nevertheless, CNNs usually also come with a number of detrimental properties, namely, (1) a high training complexity, (2) difficulty of interpretation, and (3) vulnerability to adversarial attacks.

\textbf{Adversarial examples}\,\textemdash\,The study of~\citet{LBFGS} showed that the predictions made by CNNs may change drastically with small changes in the pixel values of the input images. Specifically, when the input data at hand are modified with a gradient-based method to maximize the likelihood of another class (i.e., targeted adversarial attacks) or to minimize the likelihood of the initially correct class (i.e., untargeted adversarial attacks), it is possible to create malicious data samples that are called \textit{adversarial examples}. These adversarial examples are shown to exist not only in the digital domain but also in the real world~\citep{IFGS}, and are thus recognized as a major security threat for models that are used in a real environment (that is, an environment where the input is not strictly administered).

Research on adversarial attacks on CNNs gained traction after the seminal studies conducted by~\citet{LBFGS} and~\citet{Goodfellow-expharnessing}, with~\citet{biggio2013evasion} providing an in-depth discussion of adversarial attacks on machine learning models deployed in security-sensitive applications. Since then, the susceptibility to adversarial attacks is recognized as one of the major drawbacks of deep learning, along with the lack of clear interpretability of these models. Some studies even go so far as to suggest that these two issues should be investigated as a single topic~\citep{etmann2019connection,ross2018improving,tao2018attacks}. 

On the other side of this story, many defense techniques to prevent adversarial attacks have been proposed, only to be found ineffective~\citep{athalye2018_obfuscated,DBLP:journalsCarliniW17}. As it currently stands, there are no defense mechanisms available that reliably prevent all adversarial attacks. 

\textbf{Aim of this study}\,\textemdash\, Given the extensive research in the machine learning community on techniques to prevent adversarial attacks, we analyze the vulnerability of radar-based CNNs to adversarial examples, with the goal of assessing their significance as a security threat in smart homes. For this analysis, we consider the human activity recognition task presented in~\citet{vandersmissen2019indoor}, in which the proposed models were able to identify human gestures with high precision using a low-power FMCW radar. Our analysis of adversarial attacks covers a wide range of scenarios, from white-box attacks, in which the adversary is assumed to have all the knowledge about the underlying system, to localized attacks on radar frames under strict conditions, in which the adversary is assumed to have limited knowledge. Furthermore, we also attempt to analyze the connection between adversarial attacks and neural network interpretability by investigating the connection between prediction, perturbation amount, and Grad-CAM~\citep{grad_cam}, a popular deep neural network (DNN) interpretability technique.

This paper is organized as follows. In Section~\ref{Framework}, we describe the mathematical notation, the data set, and the machine learning models used. In Section~\ref{Threat model}, we cover the details of the different threat models considered in this study, followed by a discussion of the experiments performed and the results obtained in Section~\ref{Threat Model Evaluation}. Next, we provide a number of additional experiments in Section~\ref{Relation of Adversarial Attacks to Interpretability}, exposing the relation between adversarial examples and model interpretability. In Section~\ref{Conclusion and Future Work}, we conclude our paper and provide directions for future work.

\section{Framework and Notation}
\label{Framework}
In this section, we outline our mathematical notation, also providing details on the data and the models used in this study.

\begin{itemize}
\item $\mathbf{X}$ : an arbitrary RD frame represented as a 3-D tensor (frame count $\times$ height $\times$ width), with values in the range of $[0, 1]$.

\item $\mathbf{y} = g(\theta, \mathbf{X})$ : a classification function that links an input $\mathbf{X}$ to an output vector $\mathbf{y}$ of size $M$, containing predictions made by a neural network that comes with parameters $\theta$ and that does not contain a final softmax layer. $M$ denotes the total number of classes used. The $k$-th element of this vector is referred to as $g(\theta, \mathbf{X})_k$.

\item $J(g(\theta, \mathbf{X})_{c}) =  - \log \left( \dfrac{e^{g( \theta, \mathbf{X})_{c} }}{ \sum_{m=1}^{M} e^{g( \theta , \mathbf{X})_{m}} } \right)$ : the cross-entropy (CE) function, which calculates the negative logarithmic loss of the softmax prediction made by a neural network for a class $c$.

\item $\nabla_{x} g(\theta, \mathbf{X})$ : the partial derivative of a neural network $g$ with respect to an input $\mathbf{X}$. 

\end{itemize}

\begin{table}[t]
\centering
\caption{An overview of all the activities in the selected data set.}
\begin{tabular}{lcc}
	\cmidrule[1pt]{1-3} 
    (Class ID)  Activity  & Samples & Avg. duration (Std.)  \\ 
    \cmidrule{1-3}  
($0$) Drumming & $390$	& $2.92s$ ($\pm0.94$) \\
($1$) Shaking  	& $360$	& $3.03s$  ($\pm0.97$) \\
($2$) Swiping Left 	& $436$	& $1.60s$ ($\pm0.27$)\\
($3$) Swiping Right  & $384$	& $1.71s$ ($\pm0.31$) \\
($4$) Thumb Up	 & $409$	& $1.85s$ ($\pm0.37$)\\
($5$) Thumb Down	 & $368$	& $2.06s$ ($\pm0.42$)\\
\cmidrule[1pt]{1-3}
\end{tabular}
\label{tbl:activities}
\end{table}

\textbf{Data}\,\textemdash\,Our experiments are conducted on a data set of human gestures, containing six different hand-based actions performed in an indoor environment, published with our previous study~\citep{vandersmissen2019indoor}. The different gestures, along with the number of samples per gesture and their average duration, are listed in Table~\ref{tbl:activities}. These activities vary from dynamic and clear movements (e.g., {\it swiping left}) to static (e.g., {\it thumbs up}) and subtle (e.g., {\it drumming}) motions.

The samples are recorded using nine different subjects, with each subject repeating each activity several times, and with each subject performing different activities at different speeds and with different pause intervals. This recording approach results in less generic and more diverse activities, given that the length of the activities is not predetermined, nor is their order. The gestures are performed in front of both a radar sensor and an RGB camera, with both devices recording in a synchronized manner. As shown in Table~\ref{tbl:activities}, the extent of time in which each activity is performed differs significantly per activity class. The data set contains 2347 activities in total, with an average duration of \SI{2.16}{\second} per activity, thus making it one of the larger radar data sets concerning human actions~\citep{jokanovic2016radar_cite_data_is_small,kim2015humanr_cite_data_is_small,seyfiouglu2018deep_cite_data_is_small}.

In order to implement a number of threat scenarios, which are discussed in more detail in Section~\ref{Threat model}, and in order to work with a scenario that better reflects real-world settings, we apply a data set split different from the random stratified split used in~\cite{vandersmissen2019indoor}. Instead, we use a subject-specific split, ensuring that the data of a single subject are only present in either the training, testing, or validation set.

\begin{itemize}
    \item \textbf{$(S_{+})$}\,:\, A subject-specific split, with the training set consisting of samples obtained from subjects $2$, $6$, $8$, and $9$. Samples originating from subjects $1$ and $7$ are used for the validation set and samples obtained from subjects $3$, $4$, and $5$ are used for the test set. This approach leads to a total of $1050$, $572$, and $725$ samples for the training, testing, and validation set, respectively.
    
    \item \textbf{$(S_{-})$}\,:\, This subject-specific split is the opposite of $(S_{+})$, which means that the training set contains samples obtained from subjects $1$, $3$, $4$, $5$, and $7$. Likewise, samples obtained from the subjects $2$ and $9$ are used for the testing set, while subjects $6$ and $8$ provide samples for the validation set. This approach leads to a total of $1297$, $404$, and $646$ samples for the training, testing, and validation set, respectively.
\end{itemize}

In line with our previous study~\citep{vandersmissen2019indoor}, we consider a fixed sample length of $50$ frames, which matches the average length of the majority of the activity samples. Samples that are shorter than $50$ frames are padded with the median RD frame. This median frame is calculated by using all of the samples in the data set in order to acquire a padding frame that does not disturb the prediction (i.e., that is not an out-of-distribution sample). 
For the samples that possess more than $50$ frames, only the middle $50$ frames are considered.

\textbf{Models}\,\textemdash\,In this study, we use three substantially different architectures in order to solve the multi\-class classification problem of human activity recognition. The first architecture is the 3D-CNN architecture used in~\citet{vandersmissen2019indoor}, which we will refer to as $\mathcal{A}$. The second architecture is a variant of ResNeXt~\citep{resnext}, a relatively new architecture that achieved the second place in ILSVCR~2017~\citep{ILSVRC15:rus}. The design of this architecture is heavily inspired by VGG~\citep{VGG} and ResNet~\citep{resnet}. To handle data with a temporal dimension (e.g., radar or video data), we use a modified version of this architecture (adopted from~\citet{resnext_video}). In the remainder of this paper, we will refer to the ResNeXt architecture used as $\mathcal{R}$.

Furthermore, a recent trend in the field of activity recognition is the usage of CNN-LSTM architectures~\citep{residual_lstm_2,residual_lstm_3,residual_lstm_1}, leveraging the underlying CNN as the feature extractor and employing a Long Short-Term Memory (LSTM)~\citep{LSTM} layer in order to discover temporal relations. Apart from the usage of the previously explained fully convolutional architectures, we also employ a similarly capable CNN-LSTM architecture ($\mathcal{L}$) in order to discover differences between fully convolutional and CNN-LSTM architectures in terms of adversarial robustness. Architectural details of all models, as well as their performance on the selected dataset, can be found in the supplementary materials.

With $8.1$ million trainable parameters, the employed ResNeXt model ($\mathcal{R}$) is significantly more complex than the 3D-CNN model ($\mathcal{A}$), which contains approximately $647$ thousand trainable parameters. Consequently, the size of the two models is also considerably different: ResNeXt occupies about $32.9$MB of memory, whereas 3D-CNN only takes about $2.9$MB. Although the space occupied by each of the models does not make a significant difference for many of the current commercial products, on the same hardware, the 3D-CNN model is up to $7$ times faster than the ResNeXt model in terms of inference speed. When considering edge-computing and real-time applications in the context of smart homes, this means that deploying the ResNeXt model will naturally cost more than deploying the 3D-CNN model.

By employing fully-convolutional models that are significantly different in terms of both architecture and the number of trainable parameters, we are able to study the impact of adversarial examples generated by an \textit{advanced} model on a \textit{simpler} model, and vice versa. Moreover, by evaluating the adversarial examples generated by these fully-convolutional models on the CNN-LSTM model, we are able to analyze the effectiveness of non-LSTM adversarial examples on LSTM architectures. 

The accuracies of $\mathcal{A}$, $\mathcal{R}$, and $\mathcal{L}$ are provided in Table~\ref{tab:architecture_split_results} in the supplementary materials, as obtained for the evaluation splits. As can be observed from this table, although the number of trainable parameters is significantly different for each architecture, they achieve comparable results. Note that the models trained on the evaluation splits $S_{\{-,+\}}$ achieve slightly lower test and validation accuracies than the models presented in our previous work. This can be attributed to the reduced amount of training data we intentionally assigned to these splits, with the goal of covering a wide range of threat models (see Section~\ref{Threat model}). Throughout this paper, we adopt the notation $\mathcal{A}_{\text{Dataset split}}$ in order to describe a trained model. For instance, $\mathcal{A}_{S_+}$ means that the model is of architecture $\mathcal{A}$ and that this model has been trained on the training  set of $S_{+}$. As will be described in the next section, our approach towards selecting models and creating evaluation splits makes it possible to evaluate a wide range of white- and black-box attack scenarios.

\begin{figure}[t]
\centering
\hspace{-2em}
\begin{tikzpicture}[thick,scale=0.6, every node/.style={scale=0.9}]
\centering
    \draw[-{Latex[width=1.5mm,lightgray]},dashed, line width=0.5mm, lightgray] (0, 0) -- (9, 4.5);
    \node[align=center,rotate=0] at (9.65, 4.5) {\scriptsize Increasing\\\scriptsize difficulty};
    \draw[-{Latex[width=1.5mm]}, line width=0.5mm] (0, 0) -- (9.5, 0);
    \draw[-{Latex[width=1.5mm]}, line width=0.5mm] (0, 0) -- (0, 4.25);
    \node[align=center,rotate=0] at (10.65, 0) {\scriptsize Decreasing\\\scriptsize knowledge};
    \node[align=center,rotate=0] at (0, 4.6) {\scriptsize Increasing\\ \scriptsize complexity};
    \node[align=center,rotate=0] at (-1.4, 0.5) {\scriptsize Confidence};
    \node[align=center,rotate=0] at (-1.4, 0.15) {\scriptsize reduction};
    \node[align=center,rotate=0] at (-1.4, 1.45) {\scriptsize Misclassification};
    \node[align=center,rotate=0] at (-1.4, 2.5) {\scriptsize Targeted};
    \node[align=center,rotate=0] at (-1.4, 2.15) {\scriptsize misclassification};
    \node[align=center,rotate=0] at (-1.4, 3.5) {\scriptsize Localized attack};
    \node[align=center,rotate=0] at (1, -0.55) {\scriptsize Training data};
    \node[align=center,rotate=0] at (1, -0.95) {\scriptsize and};
    \node[align=center,rotate=0] at (1, -1.35) {\scriptsize model};
    \node[align=center,rotate=0] at (2.75, -0.55) {\scriptsize Only};
    \node[align=center,rotate=0] at (2.75, -0.95) {\scriptsize trained};
    \node[align=center,rotate=0] at (2.75, -1.35) {\scriptsize model};
    \node[align=center,rotate=0] at (4.5, -0.55) {\scriptsize Only};
    \node[align=center,rotate=0] at (4.5, -0.95) {\scriptsize training};
    \node[align=center,rotate=0] at (4.5, -1.35) {\scriptsize data};
    \node[align=center,rotate=0] at (6.5, -0.55) {\scriptsize Only};
    \node[align=center,rotate=0] at (6.5, -0.95) {\scriptsize architecture};
    \node[align=center,rotate=0] at (8.25, -0.75) {\scriptsize Surrogate};
    \draw (0.1, 0.5) -- (-0.1, 0.5);
    \draw (0.1, 1.5) -- (-0.1, 1.5);
    \draw (0.1, 2.5) -- (-0.1, 2.5);
    \draw (0.1, 3.5) -- (-0.1, 3.5);
    \draw (1, -0.1) -- (1, 0.1);
    \draw (2.75, -0.1) -- (2.75, 0.1);
    \draw (4.5, -0.1) -- (4.5, 0.1);
    \draw (6.5, -0.1) -- (6.5, 0.1);
    \draw (8.25, -0.1) -- (8.25, 0.1);
    \draw (0, -1.7) -- (3.5, -1.7);
    \draw (0, -1.7) -- (0, -1.5);
    \draw (3.5, -1.7) -- (3.5, -1.5);
    \draw (4, -1.7) -- (9, -1.7);
    \draw (4, -1.7) -- (4, -1.5);
    \draw (9, -1.7) -- (9, -1.5);
    \node[align=center,rotate=0] at (1.85, -2) {\scriptsize White-box};
    \node[align=center,rotate=0] at (6.5, -2) {\scriptsize Black-box};
    \node[] at (3, 2.5)  {WB};
    \node[] at (4.75, 2.5)  {B:1};
    \node[] at (6.5, 2.5)  {B:2};
    \node[] at (8.2, 3.2)  {attack};
    \node[] at (8.2, 3.6)  {Padding};
    \node[] at (8.2, 2.5)  {B:3};
\end{tikzpicture}
\caption{Threat configurations and their taxonomy for adversarial examples. Scenarios evaluated in this study are highlighted with their abbreviation.}
\label{fig:threat_model}
\end{figure}
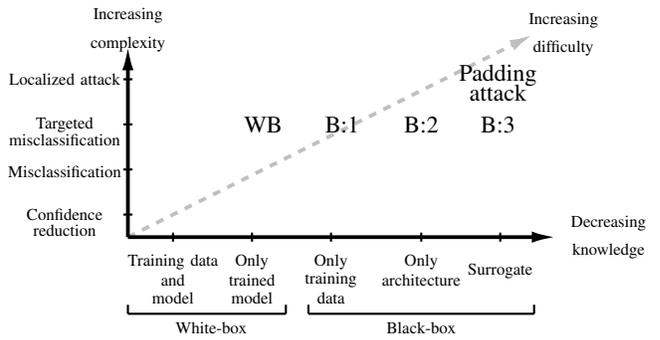

\section{Threat Model}
\label{Threat model}
In this section, we discuss the threat models evaluated in this paper. To that end, recall that activities performed in smart homes cause either a global response, meaning that the assistance of a third party is required (e.g., calling the police to prevent an intruder from entering a home or calling an ambulance for a health-related emergency situation), or a local, in-house response, meaning that the request of a household resident is related to a functionality confined to the house (e.g., turning on the lights). In this study, we evaluate threat scenarios concerning an adversarial attack to the neural network that is part of the decision making mechanism, which may affect both in-house and out-house functionalities. Naturally, there are also other types of security-related topics that need to be analyzed when a smart home system requests aid from outside the house. Such topics are mainly related to home network security and are deemed out of scope for this study.

Given the context described above, an activity and a corresponding flow of events taking place in a smart home environment are visualized in Fig.~\ref{fig:visual_summary} in the supplementary materials, as well as in the graphical abstract of this paper. We consider three entry points for a possible adversarial attack: (1) when the radar frames are generated, (2) when radar frames are being transferred from the detector to the on-site server, and (3) right before the inference phase, where frames are sent as an input to the underlying CNN.

In order to be consistent with past research that studied adversarial examples, we use a taxonomy similar to the one outlined in~\citet{JSMA}. In this context, the two main aspects of a threat model are (1) the knowledge of the adversary about the underlying system and (2) the complexity of the attack. Multiple levels of (1) and (2) are shown in Fig.~\ref{fig:threat_model}, with the labels within the figure also illustrating the different scenarios assessed in this paper. Among all possible combinations of the attacks listed in Fig.~\ref{fig:threat_model}, we only evaluate the two most restrictive cases, which are targeted misclassification and targeted misclassification with a localized attack. A detailed description of all attacks can be found in the supplementary materials.

Based on the different types of attacks and the level of knowledge of the adversary about the underlying system, we evaluate the following scenarios as threat models:

\begin{itemize}
\item \textbf{White-box threat model} (WB)\,:\, The adversary has access to the underlying trained model (including the trained weights) that performs the classification when a radar activity is performed (e.g., adversarial examples generated by $\mathcal{A}_{S_+}$ and tested on $\mathcal{A}_{S_+}$). 
\item \textbf{Black-box threat models}\,:\, The adversary does not have access to the underlying trained model (including the trained weights), but the adversary does have access to the following specifications of the underlying decision-making system:
\begin{itemize}
\item Scenario $1$ (B:1)\,:\, The adversary has access to (1) the architecture of the underlying model that performs the classification (without the trained weights) and (2) data from a similar distribution that the underlying model was trained with (e.g., adversarial examples generated by $\mathcal{A}_{S_+}$ and tested on $\mathcal{A}_{S_-}$).
\item Scenario $2$ (B:2)\,:\, The adversary has access to the training data used to train the underlying model, but not to the exact specifics of this model such as the weights, layers, and nodes (e.g., adversarial examples generated by $\mathcal{A}_{S_+}$ and tested on $\mathcal{R}_{S_+}$).
\item Scenario $3$ (B:3)\,:\, The adversary neither has access to the underlying model nor the training data used to train this model. However, the adversary has access to data obtained from a similar distribution and a model that is akin to the underlying model (e.g., adversarial examples generated by $\mathcal{A}_{S_+}$ and tested on $\mathcal{R}_{S_-}$).
\end{itemize}
\end{itemize}

In the upcoming section, we evaluate the threat models described above using various adversarial attacks.

\section{Threat Model Evaluation}
\label{Threat Model Evaluation}

In this section, we analyze the robustness of radar-based CNNs against adversarial examples using the threat models described in Section~\ref{Threat model}. To that end, we investigate the significance of commonly used adversarial attacks, as well as additional attacks that are only possible because of the existence of the temporal domain in the employed input.

\subsection{Evaluating Common Adversarial Attacks for Radar Data}
\label{Evaluating Common Adversarial Attacks for Radar Data}

Since the inception of adversarial attacks against machine learning models, a wide range of attack methods has been proposed in the literature, with each attack method focusing on a different aspect of the adversarial optimization. Fast Gradient Sign (FGS)~\citep{Goodfellow-expharnessing} is a fast way of generating adversarial examples without any iteration. Adversarial Patch~\citep{brown2017adversarial_patch} produces visible patches that, when added to the input, reliably change the prediction of the model. Basic Iterative Method (BIM)~\citep{IFGS} is an extension of FGS that generates adversarial examples in an iterative manner. Universal Perturbation, as proposed by~\citet{moosavi2017universal}, shows that it is possible to generate adversarial examples with a pre-selected perturbation pattern. Finally, the Carlini-Wagner Attack (CW)~\citep{CW_Attack} produces \textit{durable} adversarial examples that are resistant to defense systems. Next to these, there are also other black-box attacks that do not use a surrogate model to produce adversarial examples~\citep{guo2019simple_black_black_box,ilyas2018black_black_black_box}. 

Recently published studies typically aim for improving the \textit{strength} of the produced adversarial examples, making it easier to evade deployed defense systems~\citep{PGD_attack}. In the literature, BIM and CW are often selected as methods to evaluate newly proposed defense mechanisms against adversarial examples, given the diversity of their properties. Similarly, we use BIM and CW in order to investigate the vulnerability of radar-based CNNs.

\begin{table*}[t]
\centering
\caption{Median value (interquartile range) of the $L_2$ and $L_{\infty}$ distances obtained for $1000$ adversarial optimizations, as well as their success rate, for the models and data sets described in Section~\ref{Framework}. For easier comprehension, the threat models are listed from most permissive to least permissive. $L_{\infty}$ values less than $0.003$ are rolled up to $0.003$ (this is approximately the smallest amount of perturbation required to change a pixel value by $1$, so to make discretization possible).}
\begin{tabular}{ccccccccc}
 \cmidrule[1pt]{1-9}
   \multirow{2}{*}{\shortstack{Threat\\model}}  & \multirow{2}{*}{\shortstack{Source\\model}} & \multirow{2}{*}{\shortstack{Target\\model}} & \multicolumn{3}{c}{BIM} & \multicolumn{3}{c}{CW}  \\
 \cmidrule[0.25pt]{4-9}
  ~ &   ~ & ~ & $L_2$ & $L_{\infty}$ & Success \% & $L_2$ & $L_{\infty}$ & Success \% \\
\cmidrule[0.4pt]{1-9}
\multirow{4}{*}{WB} & \multirow{2}{*}{$\mathcal{A}_{\{S_+,S_-\}}$} & \multirow{2}{*}{$\mathcal{A}_{\{S_+,S_-\}}$}	 & $0.79$ & $0.003$ & \multirow{2}{*}{$100\%$} & $0.50$ & $0.02$ & \multirow{2}{*}{$97\%$}\\
~ & ~ & ~ & $(0.72)$ & $(0.003)$ & ~ &  $(0.42)$ & $(0.02)$\\
 \cmidrule[0.25pt]{2-9}
~ & \multirow{2}{*}{$\mathcal{R}_{\{S_+,S_-\}}$} & \multirow{2}{*}{$\mathcal{R}_{\{S_+,S_-\}}$}	 & $0.19$ & $0.003$ &  \multirow{2}{*}{$100\%$} & $0.17$ & $0.01$ &  \multirow{2}{*}{$99\%$} \\
~ & ~ & ~ & $(0.11)$ & $(0.003)$ & ~ & $(0.11)$ & $(0.02)$ & ~ \\
\cmidrule[0.4pt]{1-9}
 
\multirow{4}{*}{BB:1} & \multirow{2}{*}{$\mathcal{A}_{\{S_+,S_-\}}$} & \multirow{2}{*}{$\mathcal{A}_{\{S_-,S_+\}}$}	 & $2.27$ & $0.006$ & \multirow{2}{*}{$79\%$} & $0.72$ & $0.02$ & \multirow{2}{*}{$54\%$}\\
~ & ~ & ~ & $(3.17)$ & $(0.011)$ & ~ &  $(1.17)$ & $(0.09)$\\
 \cmidrule[0.25pt]{2-9}
~ & \multirow{2}{*}{$\mathcal{R}_{\{S_+,S_-\}}$} & \multirow{2}{*}{$\mathcal{R}_{\{S_-,S_+\}}$} & $0.24$ & $0.006$ &  \multirow{2}{*}{$35\%$} & $0.20$ & $0.01$ &  \multirow{2}{*}{$31\%$} \\
~ & ~ & ~ & $(0.08)$ & $(0.003)$ & ~ & $(0.17)$ & $(0.03)$ & ~ \\
\cmidrule[0.4pt]{1-9}

\multirow{4}{*}{BB:2} & \multirow{2}{*}{$\mathcal{A}_{\{S_+,S_-\}}$}  & \multirow{2}{*}{$\mathcal{R}_{\{S_+,S_-\}}$} 	 & $2.24$ & $0.009$ & \multirow{2}{*}{$35\%$} & $0.92$ & $0.03$ & \multirow{2}{*}{$30\%$}\\
~ & ~ & ~ & $(4.77)$ & $(0.012)$ & ~ & $(0.97)$ & $(0.11)$ \\
 \cmidrule[0.25pt]{2-9}
~ & \multirow{2}{*}{$\mathcal{R}_{\{S_+,S_-\}}$}  &  \multirow{2}{*}{$\mathcal{A}_{\{S_+,S_-\}}$} & $0.24$ & $0.006$ &  \multirow{2}{*}{$28\%$} & $0.22$ & $0.03$ &  \multirow{2}{*}{$27\%$} \\
~ & ~ & ~ & $(0.12)$ & $(0.003)$ & ~ & $(0.18)$ & $(0.02)$ & ~ \\ 
\cmidrule[0.4pt]{1-9}
 
\multirow{4}{*}{BB:3} & \multirow{2}{*}{$\mathcal{A}_{\{S_+,S_-\}}$}& \multirow{2}{*}{$\mathcal{R}_{\{S_-,S_+\}}$}	 & $3.12$ & $0.012$ & \multirow{2}{*}{$30\%$} & $0.87$ & $0.03$ & \multirow{2}{*}{$21\%$}\\
~ & ~ & ~ & $(5.14)$ & $(0.07)$ & ~ &  $(1.03)$ & $(0.13)$ \\
 \cmidrule[0.25pt]{2-9}
~ &\multirow{2}{*}{$\mathcal{R}_{\{S_+,S_-\}}$} & \multirow{2}{*}{$\mathcal{A}_{\{S_-,S_+\}}$}	 & $0.22$ & $0.006$ &  \multirow{2}{*}{$19\%$} & $0.21$ & $0.02$ &  \multirow{2}{*}{$18\%$} \\
~ & ~ & ~ & $(0.10)$ & $(0.005)$ & ~ & $(0.27)$ & $(0.04)$ & ~ \\ 

 \cmidrule[1pt]{1-9} 
\end{tabular}
\label{tbl:attack-table}
\end{table*}

\textbf{Basic Iterative Method}\,\textemdash\,This attack is an extension of FGS, which uses the signature of the cross-entropy loss in order to generate adversarial examples. On the one hand, BIM is seen as a method for generating \textit{weak} adversarial examples that heavily perturb the input. On the other hand, BIM is highly efficient because it generates adversarial examples much faster than CW. It is defined as follows:
\begin{align} \label{eq:ifgs}
\mathbf{X}_{n+1} =  Clip_{\mathbf{X}, \epsilon}(  \mathbf{X}_{n} - \alpha \ \text{sign} (\nabla_x J(g(\theta,\mathbf{X}_n)_c)  )) \,,
\end{align}
where the $Clip$ function ensures that the adversarial example $\mathbf{X}_{n+1}$ is a valid input (i.e., an image) and where $\alpha$ denotes the perturbation multiplier.
In this study, we use $\alpha = 15\times10^{-4}$, meaning that a single iteration of perturbation will change values by half a pixel value (i.e., $0.5/255$), reporting results based on the use of this perturbation multiplier. 

\textbf{Carlini-Wagner Attack}\,\textemdash\,The Carlini-Wagner attack is proposed as a method to generate \textit{strong} adversarial examples \citep{CW_Attack}. This attack uses multi-target optimization and maximizes the prediction likelihood of both the target class and second-most-likely class in order to deceive the underlying machine learning model. CW is criticized for its high computational complexity, which is primarily due to its extensive search for \textit{strong} perturbations~\citep{Goodfellow:2018:MML:3234519.3134599}. It is defined as follows:
\begin{align} \label{eq:CW_2}
& \text{miminize} \quad  ||\mathbf{X} - (\mathbf{X} + \delta)||_{2}^{2} +  \; \ell(\mathbf{X} + \delta)\,, \\
& \ell (\mathbf{X}') =  \max \big(  \max \{ g(\theta, \mathbf{X}')_i : i\neq c\} - g(\theta, \mathbf{X}')_c, -  \kappa \big) \,
\end{align}
where $\delta$ is the perturbation added to the image and $\ell$ a loss function. This equation aims at maximizing the prediction likelihood of the target class $c$ and the second-most likely class $i$, with $\kappa$ controlling the logit difference between both classes. We adhere to the study of \citet{CW_Attack} and set $\kappa = 20$.

\textbf{Constraints on Adversarial Attacks}\,\textemdash\,During the generation of adversarial examples, when not considering constraints for the generated adversarial examples, (1) the optimization may result in an adversarial example that does not represent a valid input for the targeted neural network or (2) the attack may not be representative of a real-world scenario. In order to avoid such scenarios, we impose a box constraint, a time constraint, and a discretization constraint on the way adversarial examples are generated. A detailed description of these constraints can be found in the supplementary materials.

\textbf{Experiments}\,\textemdash\, In Table~\ref{tbl:attack-table}, we present the experimental results obtained in terms of white-box and black-box transferability success, for the adversarial examples created with BIM and CW. Specifically, Table~\ref{tbl:attack-table} details the success rate obtained for $1000$ adversarial examples that originate from unseen data points by their respective models during training time, as well as median $L_2$ and $L_{\infty}$ distances between adversarial examples and their initial data points, giving an idea of the minimum amount of perturbation necessary to change the prediction of a model by both attacks. Since we use the $L_2$ and $L_{\infty}$ distances between adversarial examples when they transfer successfully from a source model to a target model, we also provide the interquartile range in order to gain insight into the spread of the $L_2$ and $L_{\infty}$ distances. We use median and interquartile range over mean and standard deviation in order to mitigate the influence of outliers when the success rate of the attacks is low. Based on this experiment, we make the following observations:

\begin{itemize}
\item Unsurprisingly, the success rate of the adversarial attacks decreases as the knowledge of the adversary on the underlying system decreases. As opposed to this trend, the minimal required perturbation to change the prediction of the target model often increases as the knowledge of the adversary decreases.

\item More often than not, adversarial attacks with BIM are more successful than the ones with CW, even though the latter is considered a more advanced attack. Using additional experiments, we observed this can be primarily attributed to the time constraint imposed on the optimization. Since CW is computationally more expensive than BIM, given its extensive search for a minimum amount of perturbation, generating an adversarial example with CW within the time limit imposed becomes challenging.

\item Although BIM is more successful in generating adversarial examples than CW, the adversarial examples generated by BIM come with much stronger perturbations in terms of $L_2$ distance than those generated by CW. On the other hand, thanks to the $sign(\cdot)$ function flattening the gradients to an equal level, adversarial examples created with BIM come with much less perturbation in terms of $L_{\infty}$ distance. This finding for radar data is also in line with the observations we have made in the image domain~\citep{ozbulak2020perturbation}.

\item Even though ResNeXt models are able to find adversarial examples with less perturbation, the time limit set on the generation of adversarial examples also affects ResNeXt models more than 3D-CNN models, since it takes them longer to perform a prediction, as well as to calculate the gradients for adversarial example generation, ultimately resulting in lower success rates.

\item For all black-box cases, the ResNeXt architecture is able to find adversarial examples with much less perturbation than 3D-CNN. Our initial interpretation of this finding was that the adversarial examples generated from \textit{stronger} models transfer with less perturbation when attacking similar or \textit{weaker} models. However, recent results in the area of adversarial research suggest that residual models that contain skip-connections allow the generation of adversarial examples with much less perturbation~\citep{skip_connections_easier}. Our experiments also confirm this observation.

\item Experimental results obtained for the CNN-LSTM architecture (presented in Table~\ref{tbl:attack-table-lstm} in the supplementary materials) show that adversarial examples generated by fully convolutional architectures are capable of adversarial transferability. Moreover, we observed that the CNN-LSTM architecture employed in this study provides no additional security compared to fully convolutional models. A detailed discussion of these results is presented in the supplementary materials.

\end{itemize}

Detailed visual examples of the degree of perturbation needed and the perturbation visibility can be found in Fig.~\ref{fig:l2_comparison} in the supplementary materials.

\subsection{Adversarial Padding for Radar Data}
\label{Adversarial Padding for Radar Data}

\begin{table}[t]
\centering
\caption{Median value (interquartile range) of the $L_2$ and $L_{\infty}$ distances obtained for $1000$ adversarial optimizations, as well as their success rate, for the models and data sets described in Section~\ref{Framework}, hereby using the padding attack described in Section~\ref{Adversarial Padding for Radar Data}.}
\begin{tabular}{ccccc}
 \cmidrule[1pt]{1-5}
    \multirow{2}{*}{\shortstack{Source\\model}} & \multirow{2}{*}{\shortstack{Target\\model}} & \multicolumn{3}{c}{Padding attack}  \\
\cmidrule[0.25pt]{3-5}
   ~ & ~ & $L_2$ & $L_{\infty}$ & Success \% \\
\cmidrule[0.4pt]{1-5}
\multirow{2}{*}{$\mathcal{A}_{\{S_+,S_-\}}$} & \multirow{2}{*}{$\mathcal{A}_{\{S_+,S_-\}}$}	 & $3.34$ & $0.012$ & \multirow{2}{*}{$84\%$} \\
 ~ & ~ & $(0.56)$ & $-$ & ~ \\
  \cmidrule[0.25pt]{1-5}
\multirow{2}{*}{$\mathcal{R}_{\{S_+,S_-\}}$} & \multirow{2}{*}{$\mathcal{R}_{\{S_+,S_-\}}$}	 & $3.07$ & $0.012$ &  \multirow{2}{*}{$74\%$}  \\
~ & ~ & $(0.28)$ & $-$ & ~ \\
\cmidrule[0.4pt]{1-5}
 
\multirow{2}{*}{$\mathcal{A}_{\{S_+,S_-\}}$} & \multirow{2}{*}{$\mathcal{A}_{\{S_-,S_+\}}$}	 & $3.47$ & $0.011$ & \multirow{2}{*}{$68\%$} \\
 ~ & ~ & $(0.82)$ & $-$ & ~ \\
  \cmidrule[0.25pt]{1-5}
 \multirow{2}{*}{$\mathcal{R}_{\{S_+,S_-\}}$} & \multirow{2}{*}{$\mathcal{R}_{\{S_-,S_+\}}$} & $4.19$ & $0.010$ &  \multirow{2}{*}{$51\%$} \\
 ~ & ~ & $(1.07)$ & $-$ & ~ \\
\cmidrule[0.4pt]{1-5}

\multirow{2}{*}{$\mathcal{A}_{\{S_+,S_-\}}$}  & \multirow{2}{*}{$\mathcal{R}_{\{S_+,S_-\}}$} 	 & $5.19$ & $0.011$ & \multirow{2}{*}{$48\%$} \\
 ~ & ~ & $(0.95)$ & $-$ & ~ \\
  \cmidrule[0.25pt]{1-5}
 \multirow{2}{*}{$\mathcal{R}_{\{S_+,S_-\}}$}  &  \multirow{2}{*}{$\mathcal{A}_{\{S_+,S_-\}}$} & $5.56$ & $0.012$ &  \multirow{2}{*}{$28\%$} \\
 ~ & ~ & $(0.87)$ & $-$ & ~ \\ 
\cmidrule[0.4pt]{1-5}
 
 \multirow{2}{*}{$\mathcal{A}_{\{S_+,S_-\}}$}& \multirow{2}{*}{$\mathcal{R}_{\{S_-,S_+\}}$}	 & $4.67$ & $0.010$ & \multirow{2}{*}{$41\%$}\\
 ~ & ~ & $(1.21)$ & $-$ & ~  \\
  \cmidrule[0.25pt]{1-5}
\multirow{2}{*}{$\mathcal{R}_{\{S_+,S_-\}}$} & \multirow{2}{*}{$\mathcal{A}_{\{S_-,S_+\}}$}	 & $5.47$ & $0.012$ &  \multirow{2}{*}{$25\%$} \\
~ & ~ & $(0.80)$ & $-$ & ~  \\ 

 \cmidrule[1pt]{1-5} 
\end{tabular}
\label{tbl:pad-attack-table}
\end{table}

Our experiments show that, in the most restrictive case, the success rate of the adversarial attacks falls as low as $18\%$. The reason behind the low success rate can again be mainly attributed to the time constraint imposed on the attacks. However, an attacker may already be in possession of a pattern of adversariality that is ready to be deployed without needing any additional computation, thus nullifying the time constraint set on generating an adversarial example. In the literature, this type of attacks is called \textit{universal perturbations}~\citep{moosavi2017universal}. The main focus of these studies is to generate a universal perturbation pattern in advance and use it during inference time.

\citet{moosavi2017universal} demonstrated that universal perturbations exist for DNNs and that these universal perturbations may even be diverse in nature (i.e., more than one universal perturbation may be available). Instead of evaluating the techniques proposed in \citet{moosavi2017universal}, which allow for finding a universal perturbation eventually, we experiment with the idea of generating adversarial padding. This approach can then be employed for samples in the data set that contain less than 50 frames, facilitating prediction without requiring additional optimization. In other words, adversarial padding allows an adversary to change the prediction without even having to modify the frames in which the activity of interest takes place. Inspired by the work of~\citet{moosavi2017universal}, we use the approach described below to generate the adversarial padding.

Let $\mathbf{P}\in [0,1]^{80\times126}$ be the median padding used in this study, as described in Section~\ref{Framework}, and let $\mathbf{X} \in [0,1]^{\ell\times80\times126}$ be an $\ell$-frame-long activity for some $\ell\in\{1,\ldots,50\}$. We define $\mathbf{M}_{1} = [ \mathbf{P} \, \ldots \, \mathbf{P}] \in [0,1]^{50\times80\times126}$ as the initial adversarial padding pattern made up of multiple $\mathbf{P}$s and denote by $\mathbf{M}^i$, $i \in \{1,\ldots,50\}$, the $i$-th frame of $\mathbf{M}$. In order to generate the adversarial padding, we use the following approach:
\begin{align}
&\bar{\mathbf{X}} = [\mathbf{X} \, \underbrace{\mathbf{P} \ldots \mathbf{P}}_{50-\ell}] \,, \quad
\hat{\mathbf{X}} = [\mathbf{X} \, \underbrace{\mathbf{M}^{\ell+1} \ldots \mathbf{M}^{50}}_{50-\ell}] \,,\\
&\text{minimize} \,\,  ||\, \bar{\mathbf{X}} \, - \, \hat{\mathbf{X}} \,||_{2} \,, \\
&\text{such that} \,\,  \arg \max \big(g(\theta, \hat{\mathbf{X}})  \big) = c,\,\,\, \arg \max \big(g(\theta, \bar{\mathbf{X}})  \big) \neq c
\end{align}

where $c$ is the target class. We then calculate $\mathbf{M}$ in an iterative manner as follows:
\begin{align}
&\mathbf{M}_{n+1}  = \mathbf{M}_n +\big( \alpha \, \nabla_x \, \big( g(\theta, [\mathbf{X}\,\,\mathbf{M}^{\ell+1}_n \ldots \mathbf{M}^{50}_n] ) \big )_c \, \odot \, \mathbf{K} \big) \,,\\
&\mathbf{K}^i  =
    \begin{cases}
      1^{80\times126}, & \text{if}\ i \in \{i_1, i_2, i_3\} \\
      0^{80\times126}, & \text{otherwise} \\
    \end{cases} \,, \\ 
&\text{with}  \,\, ||\mathbf{M}^{\{i_1,i_2,i_3\}}_n - \mathbf{P}||_2 < ||\mathbf{M}^{i_u}_n - \mathbf{P}||_2 \,,\\ & \quad \quad \, \forall i_u \in \{1,\ldots,50\}\setminus\{i_1, i_2, i_3\} \,,
\end{align}
where $\odot$ is the element-wise tensor multiplication. The idea behind using $\mathbf{K}$ is to force the optimization to modify the least modified padding frames (in this case, the three frames $i_1$, $i_2$, and $i_3$). This leads to a more uniform distribution of the perturbation over the padding frames, rather than having perturbation that is concentrated in just a few frames. Indeed, we observed that the optimization focuses on just a few frames rather than all frames when $\mathbf{K}$ is not incorporated, resulting in an adversarial example that is not able to reliably change the prediction. We also experimented with selecting more than three frames: although results were comparable, selecting three frames produced the best results in terms of adversarial transferability.

By following the aforementioned procedure, we are able to create padding sequences that convert model predictions to the targeted class, without even having to change the frames in which the activity occurs. In order to demonstrate the effectiveness of this attack, we provide Table~\ref{tbl:pad-attack-table}, containing details about the success rate of the padding attack, obtained under the same conditions as the results presented in Table~\ref{tbl:attack-table}. Note that the $L_{\infty}$ interquartile range is not presented in Table~\ref{tbl:pad-attack-table} because the padding attack aims at spreading the adversarial perturbation equally over all padding frames, thus keeping the $L_{\infty}$ distance between genuine data points and their adversarial counterparts for multiple data points approximately the same.

Given Table~\ref{tbl:pad-attack-table} in the main text and Table~\ref{tbl:pad-attack-table-lstm} in the supplementary materials, the first observation is that the padding attack cannot achieve a success rate upwards of $80\%$ for white-box cases. This is because certain samples, which usually belong to activities $0$ and $1$, are either not padded or padded with very few frames (see Table~\ref{tbl:activities} for average activity duration). Thus, it is very challenging, or downright impossible, for the padding attack to change the prediction. Trivially, the shorter activities are more affected by the padding attack. Furthermore, we can observe that the padding attack achieves higher success rates in most black-box cases than the attacks presented in Table~\ref{tbl:attack-table}, albeit by incorporating stronger perturbations.

\begin{figure*}[t]
\centering
\rotatebox[origin=l]{90}{\phantom{----}Frame: 36}\hspace{0.4em}
\subfloat{{\includegraphics[width=4.2cm, height=2.6cm]{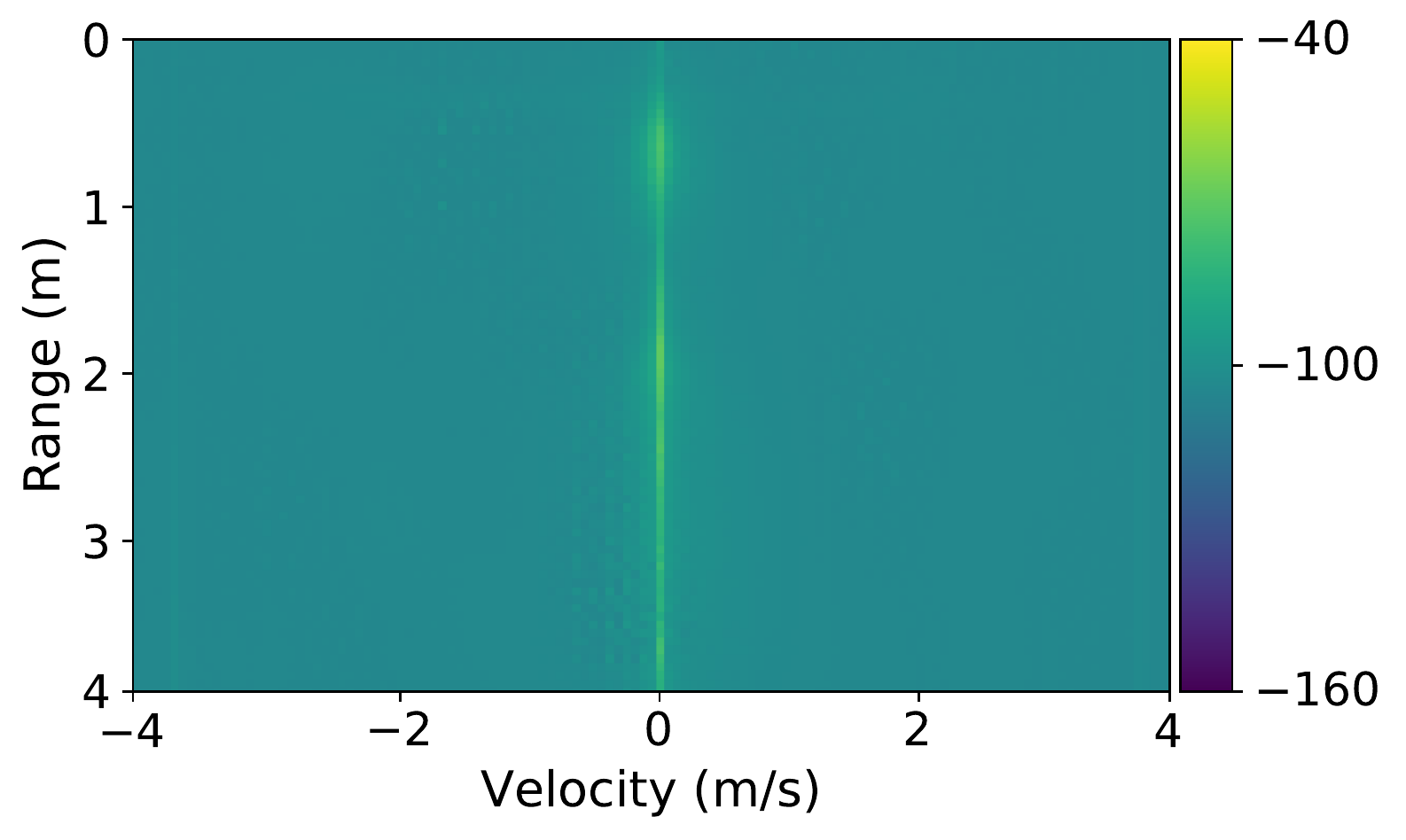}}}
\hspace{0.3em}
\rotatebox[origin=l]{90}{\phantom{----}Frame: 41}\hspace{0.4em}
\subfloat{{\includegraphics[width=4.2cm, height=2.6cm]{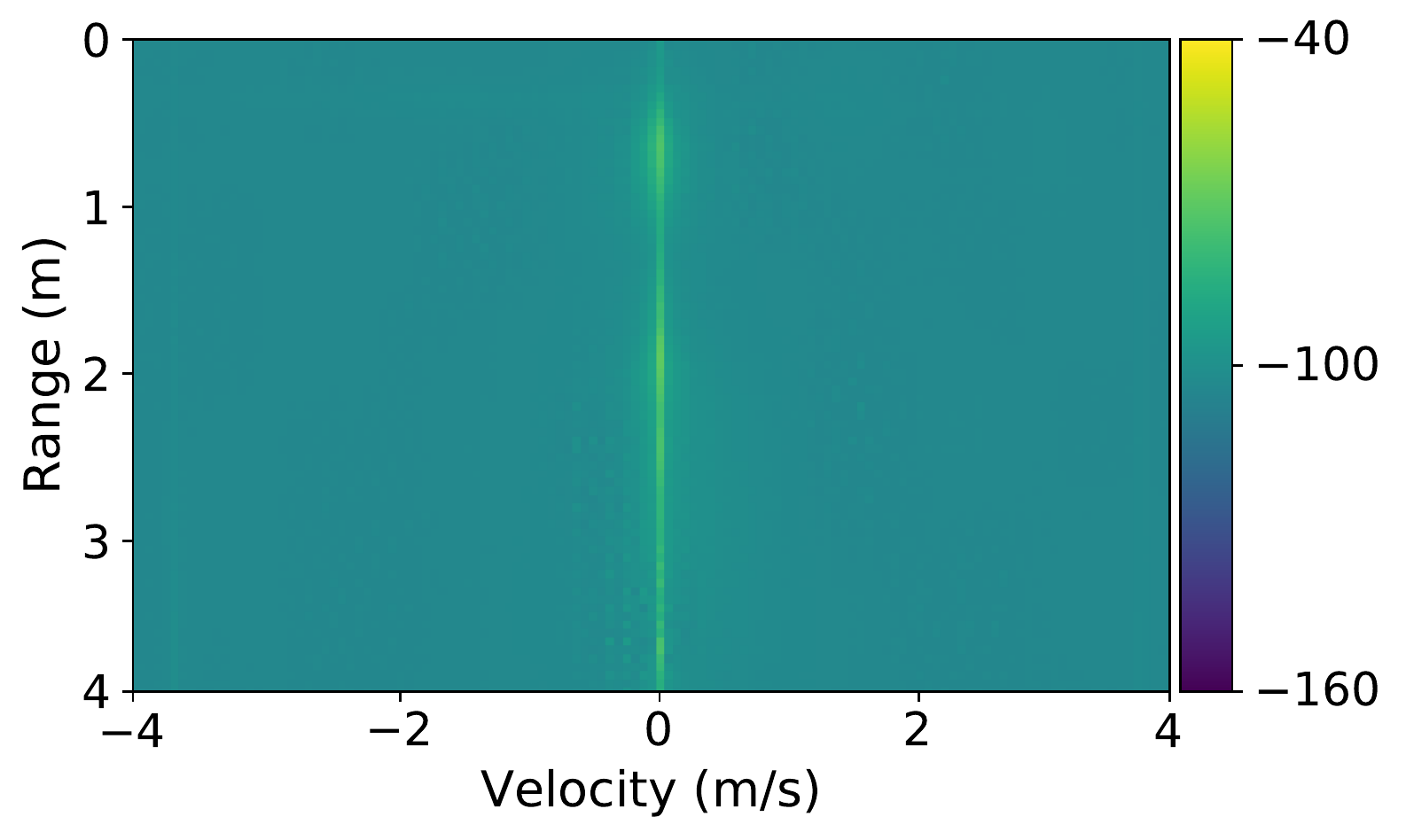}}}
\hspace{0.3em}
\rotatebox[origin=l]{90}{\phantom{----}Frame: 46}\hspace{0.4em}
\subfloat{{\includegraphics[width=4.2cm, height=2.6cm]{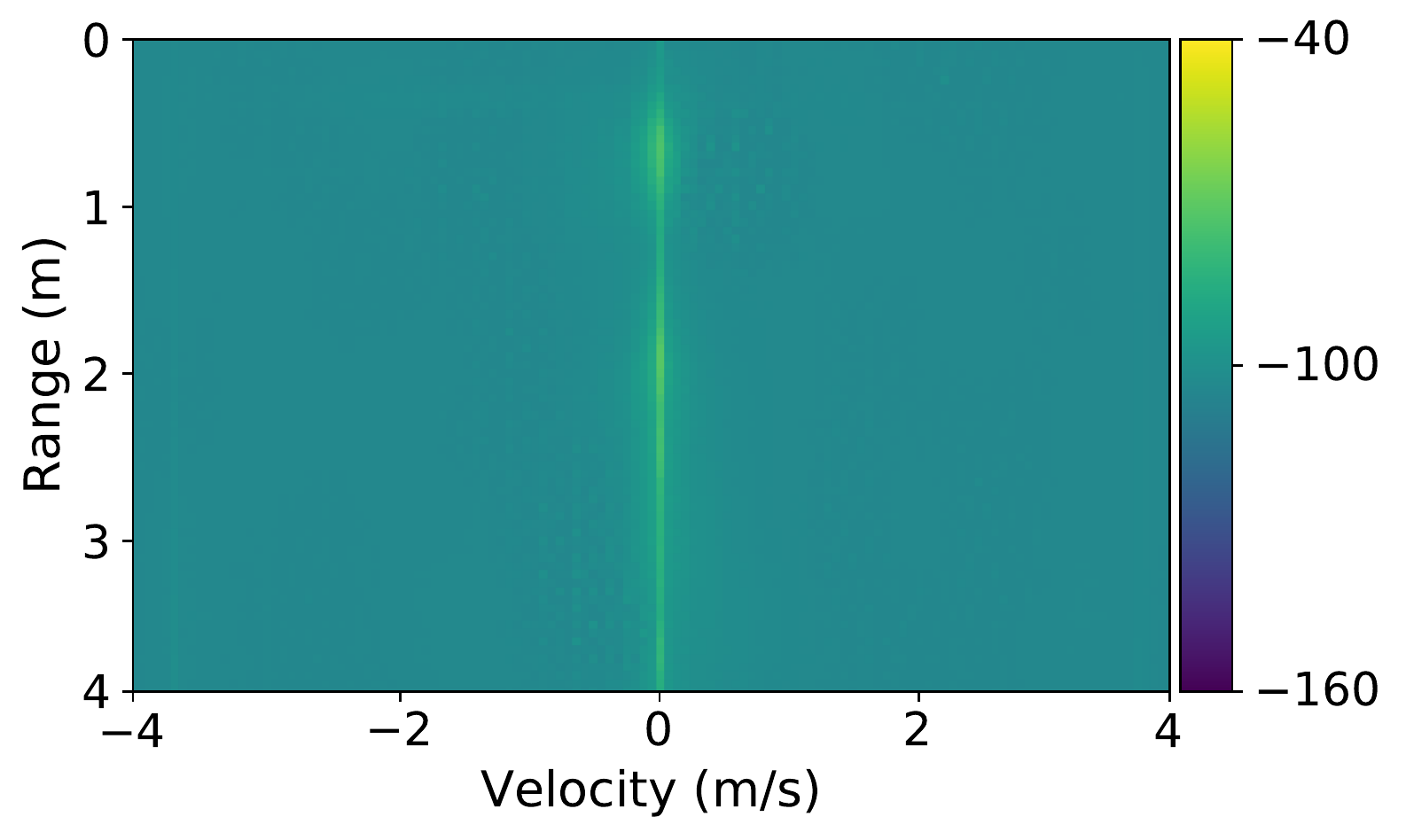}}}
\\\vspace{0.5em}
\rotatebox[origin=l]{90}{\phantom{----}Frame: 37}\hspace{0.4em}
\subfloat{{\includegraphics[width=4.2cm, height=2.6cm]{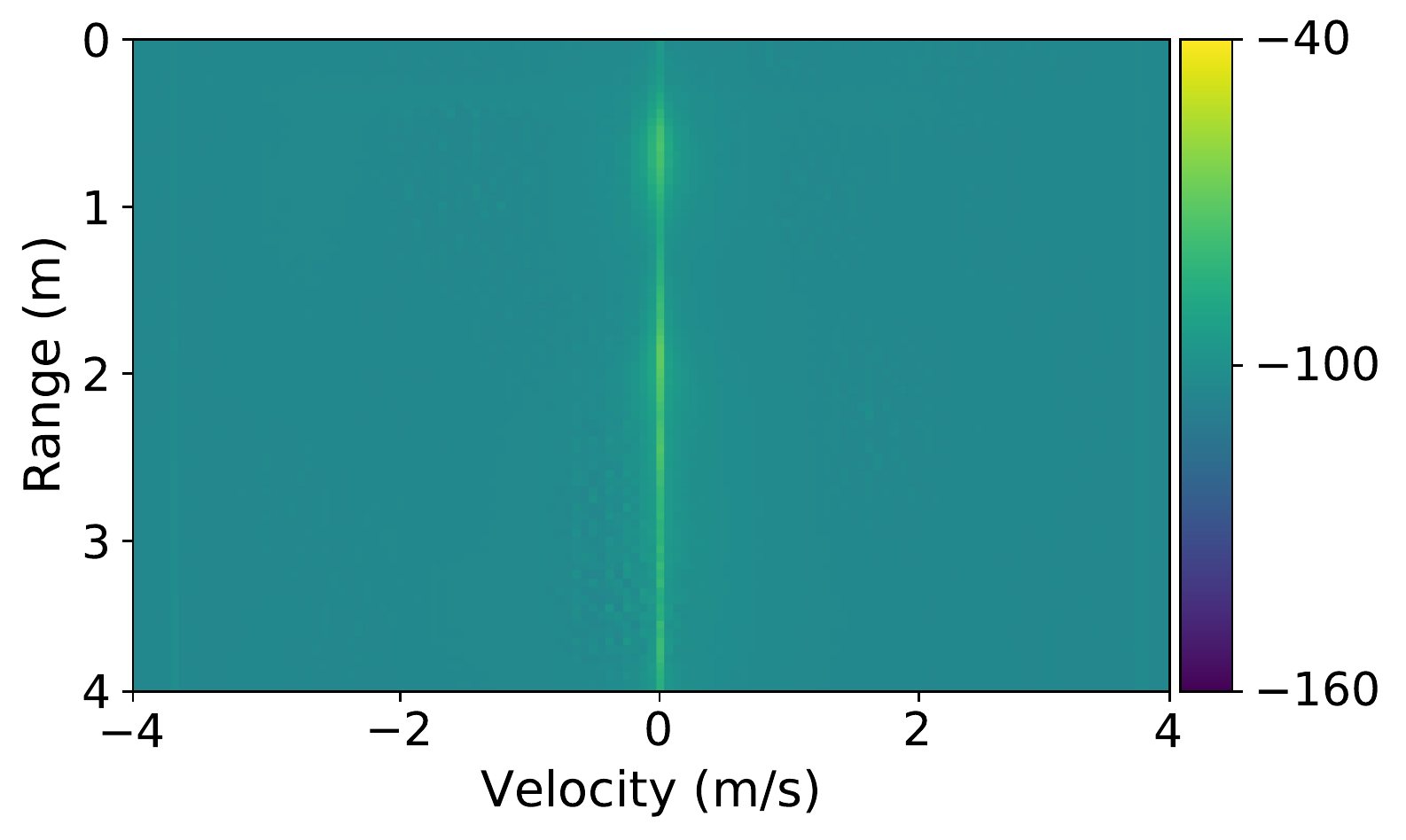}}}
\hspace{0.3em}
\rotatebox[origin=l]{90}{\phantom{----}Frame: 42}\hspace{0.4em}
\subfloat{{\includegraphics[width=4.2cm, height=2.6cm]{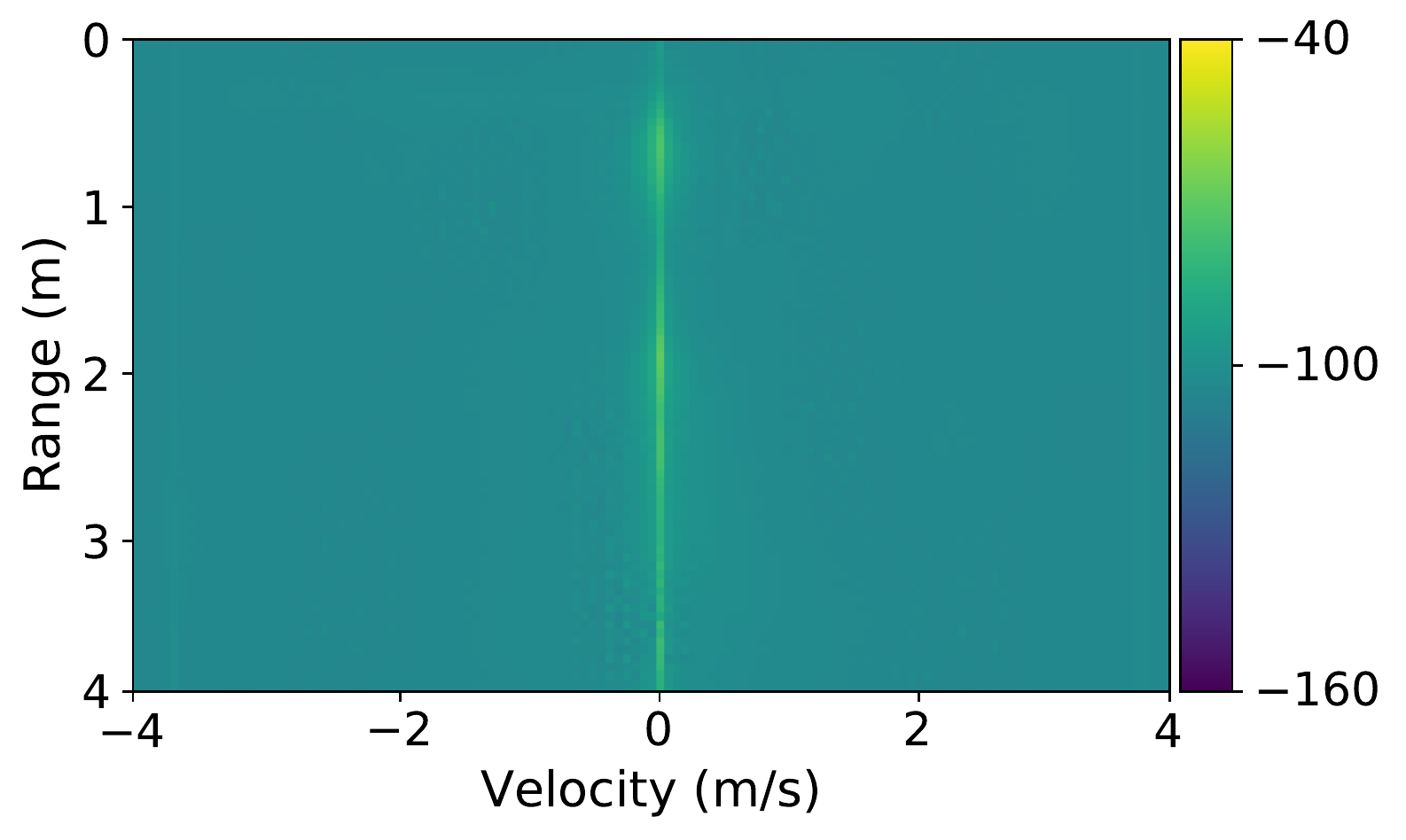}}}
\hspace{0.3em}
\rotatebox[origin=l]{90}{\phantom{----}Frame: 47}\hspace{0.4em}
\subfloat{{\includegraphics[width=4.2cm, height=2.6cm]{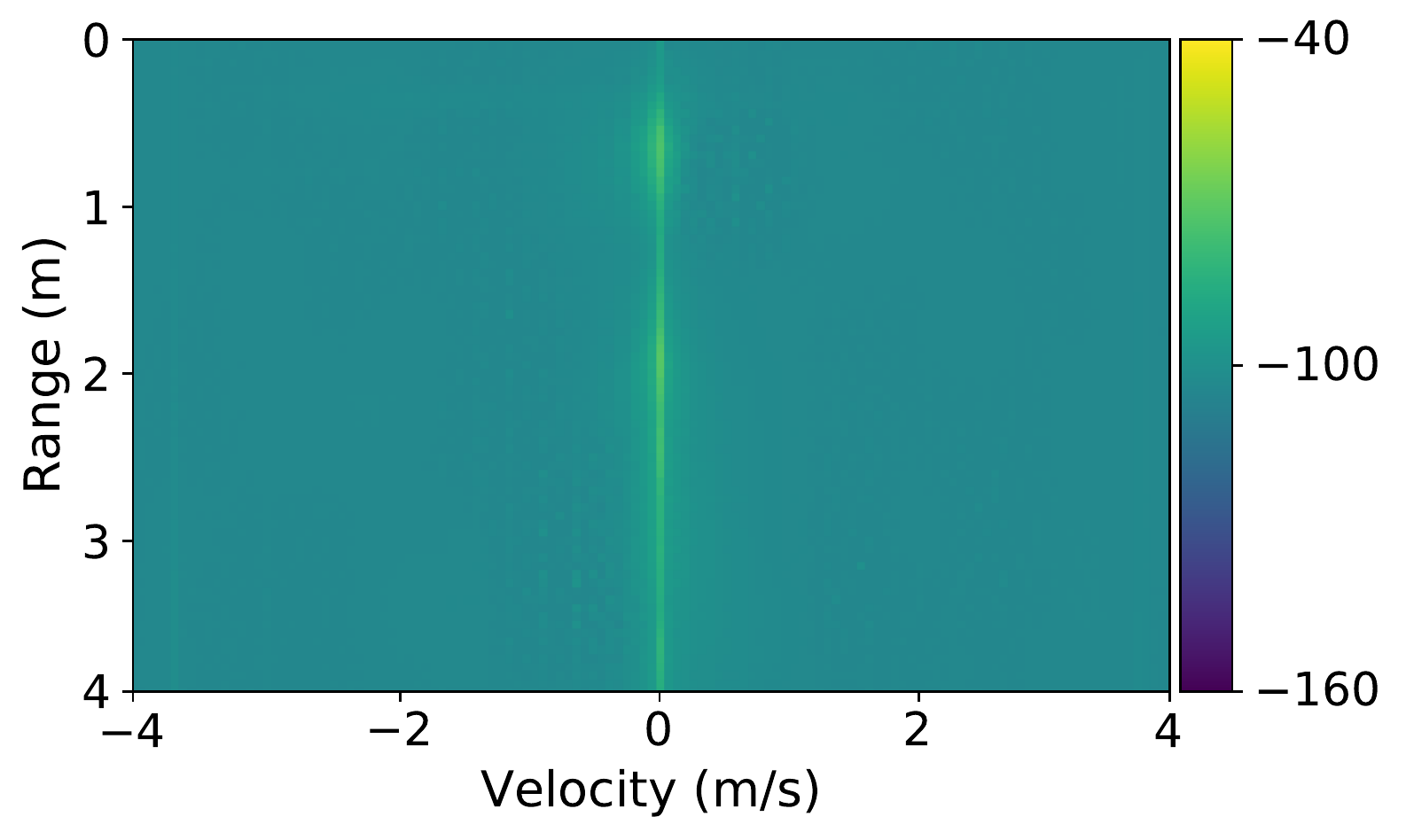}}}
\\\vspace{0.5em}
\rotatebox[origin=l]{90}{\phantom{----}Frame: 38}\hspace{0.4em}
\subfloat{{\includegraphics[width=4.2cm, height=2.6cm]{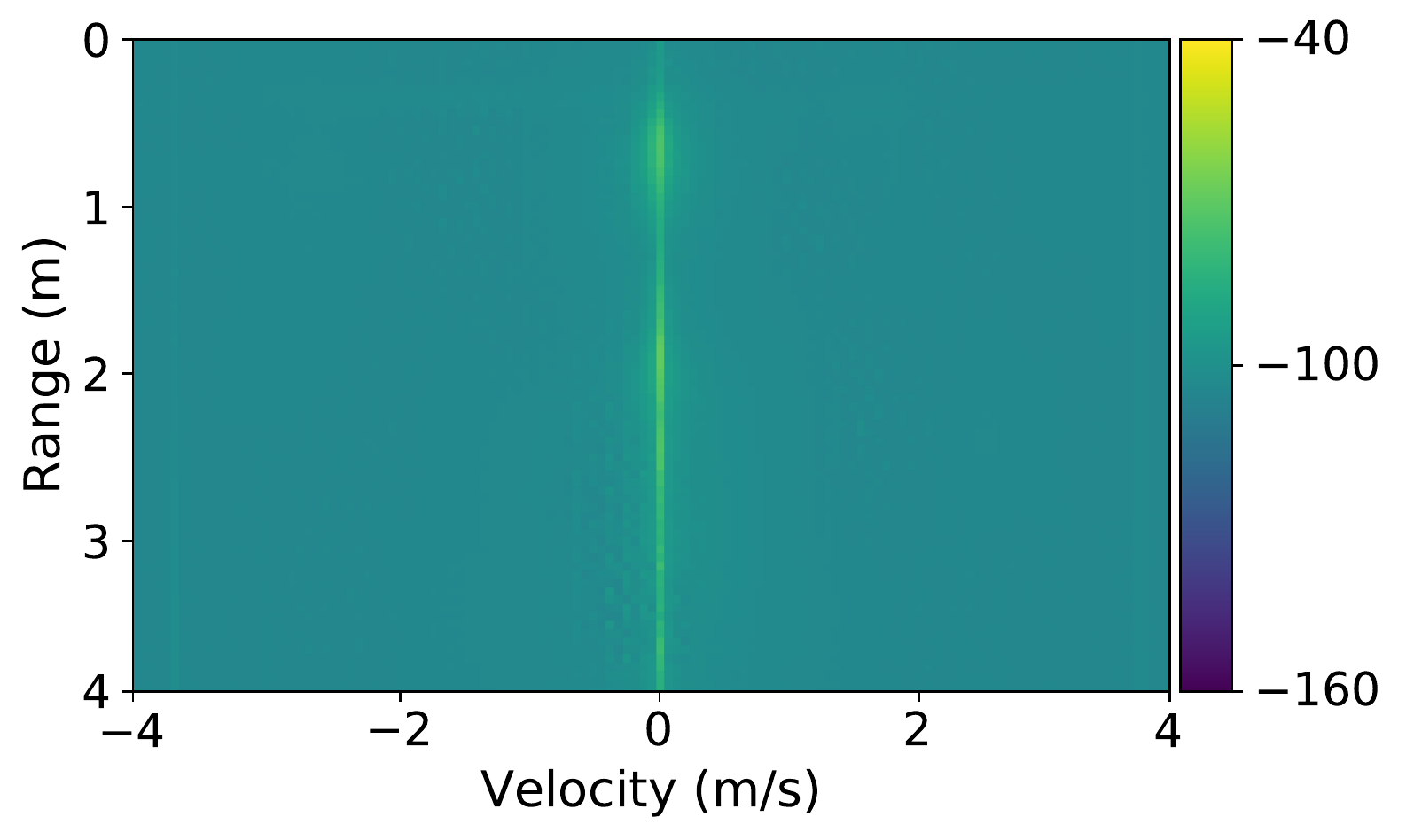}}}
\hspace{0.3em}
\rotatebox[origin=l]{90}{\phantom{----}Frame: 43}\hspace{0.4em}
\subfloat{{\includegraphics[width=4.2cm, height=2.6cm]{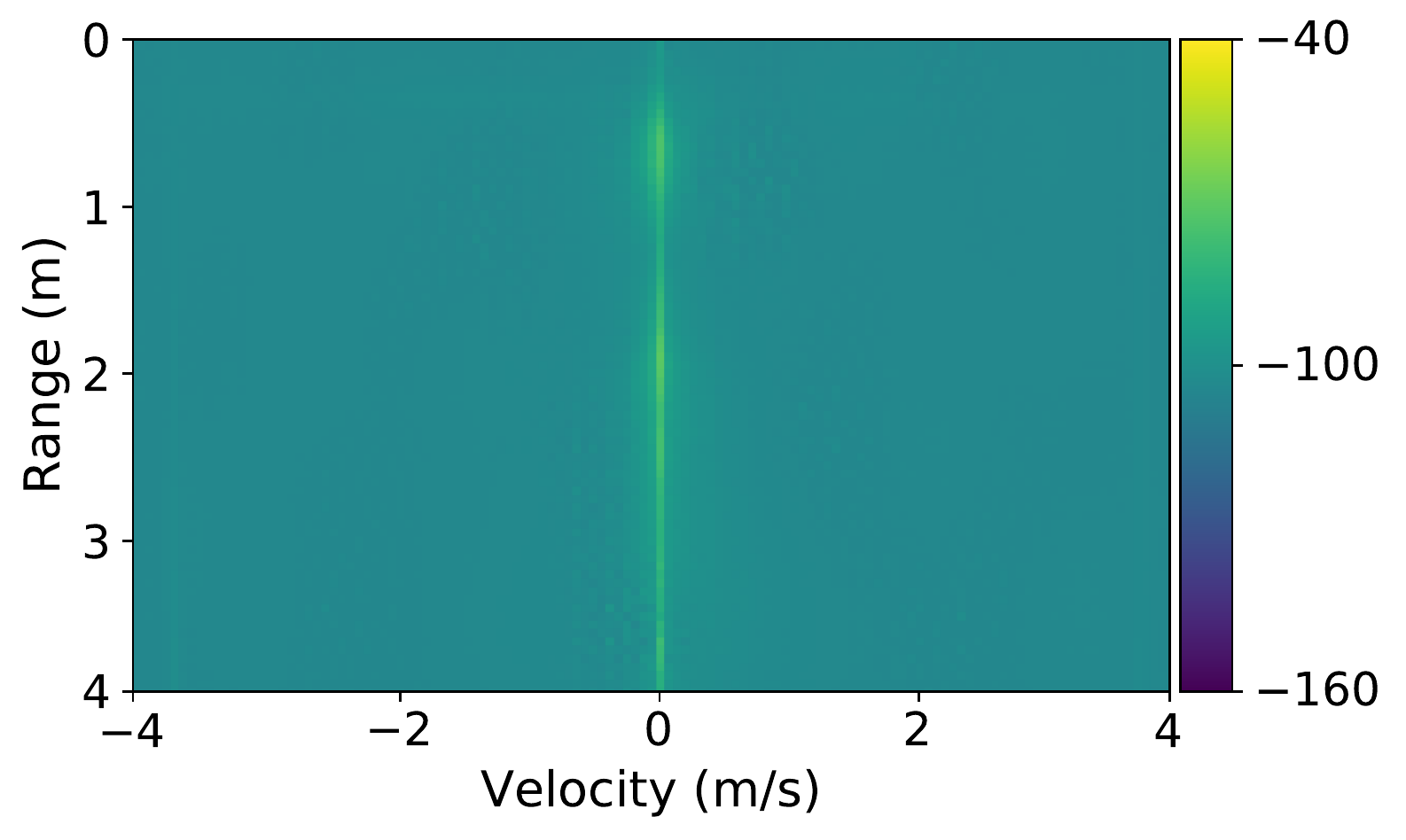}}}
\hspace{0.3em}
\rotatebox[origin=l]{90}{\phantom{----}Frame: 48}\hspace{0.4em}
\subfloat{{\includegraphics[width=4.2cm, height=2.6cm]{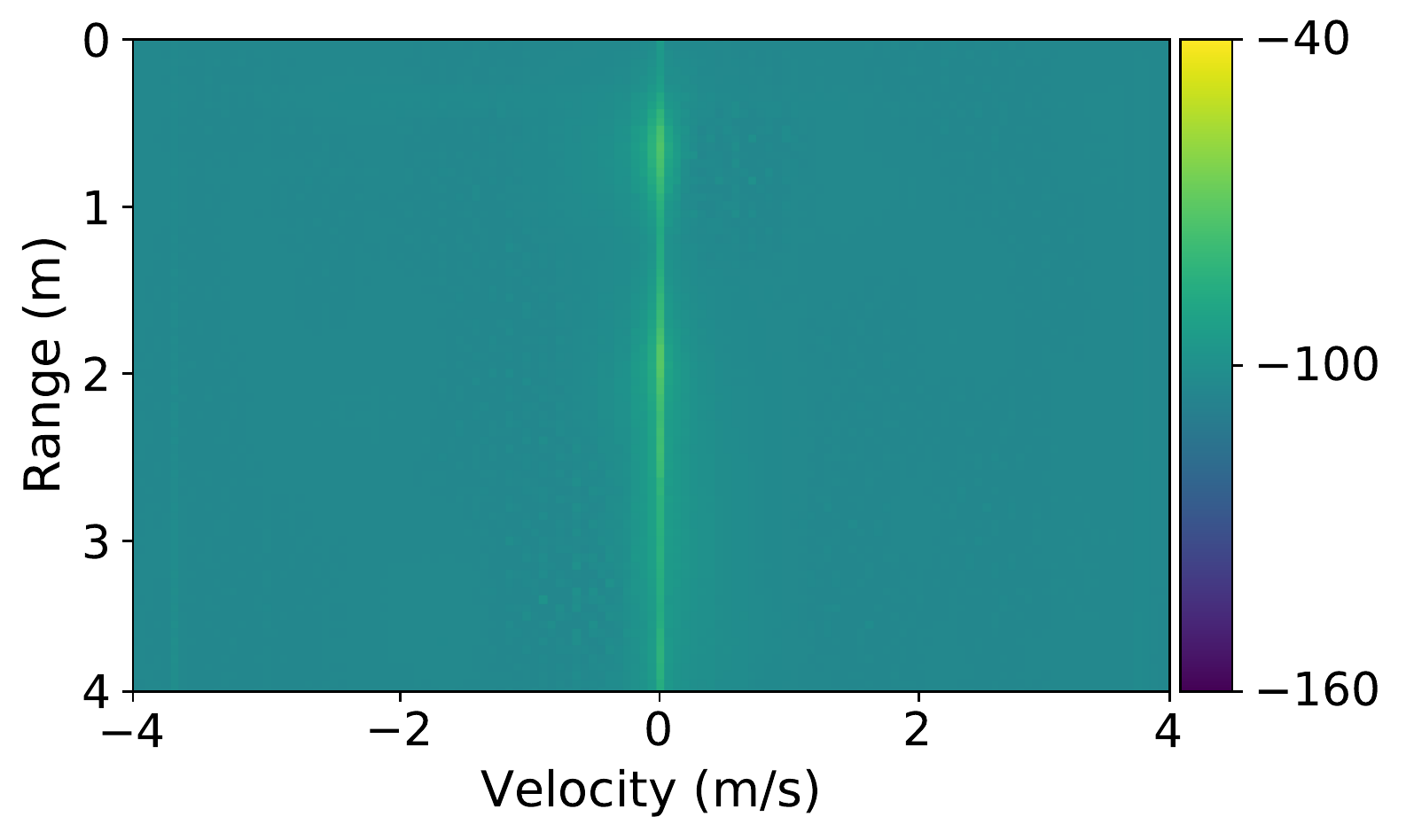}}}
\\\vspace{0.5em}
\rotatebox[origin=l]{90}{\phantom{----}Frame: 39}\hspace{0.4em}
\subfloat{{\includegraphics[width=4.2cm, height=2.6cm]{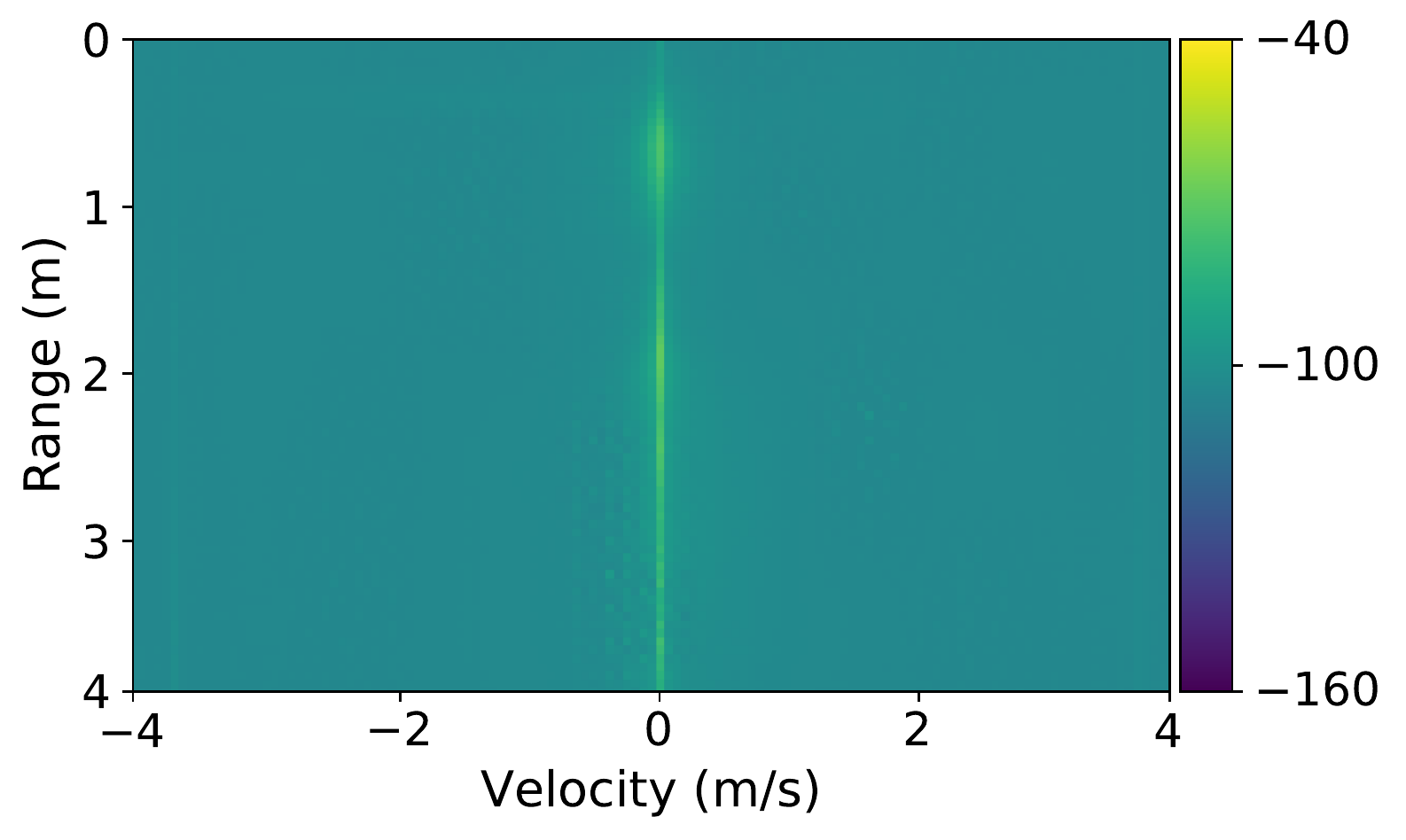}}}
\hspace{0.3em}
\rotatebox[origin=l]{90}{\phantom{----}Frame: 44}\hspace{0.4em}
\subfloat{{\includegraphics[width=4.2cm, height=2.6cm]{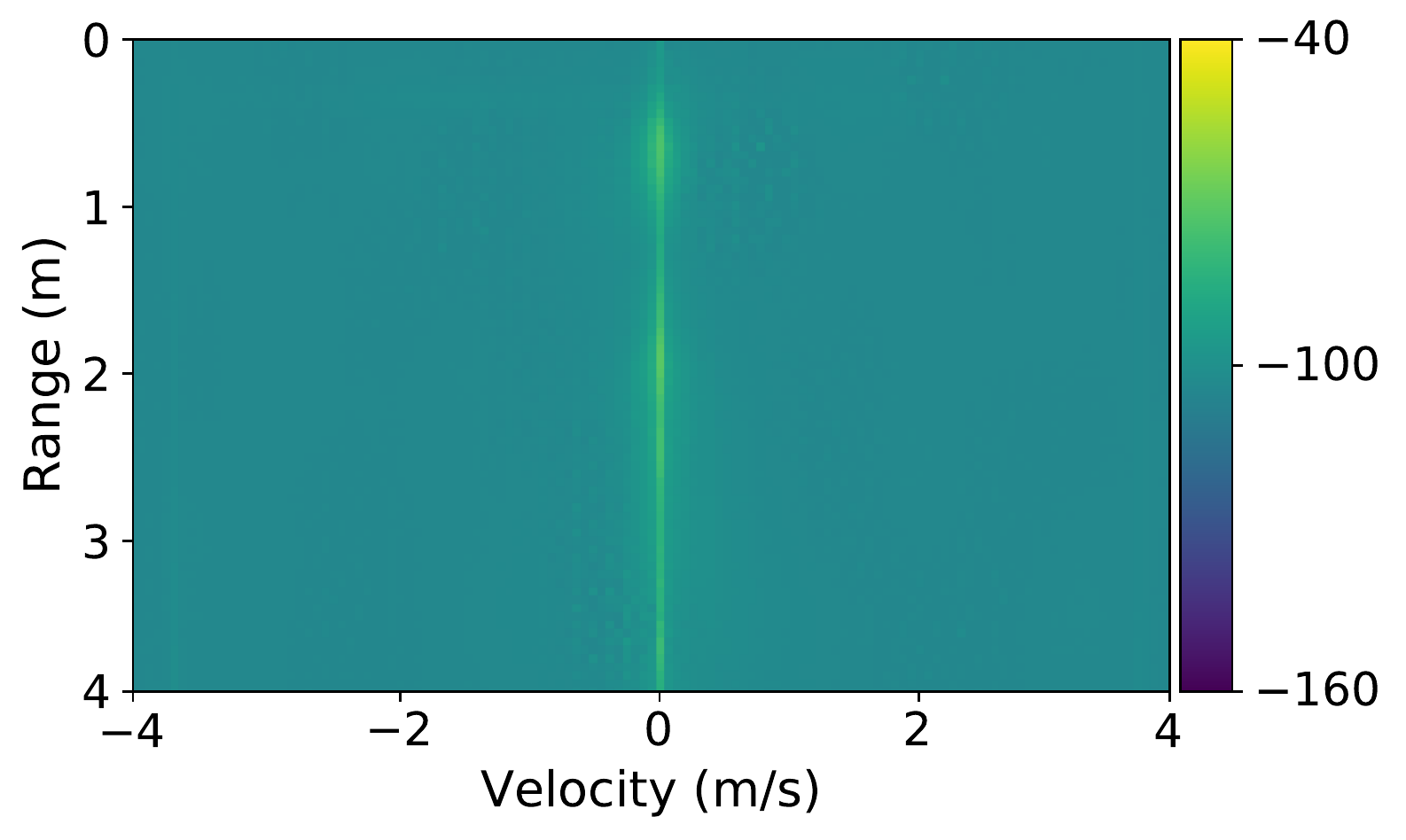}}}
\hspace{0.3em}
\rotatebox[origin=l]{90}{\phantom{----}Frame: 49}\hspace{0.4em}
\subfloat{{\includegraphics[width=4.2cm, height=2.6cm]{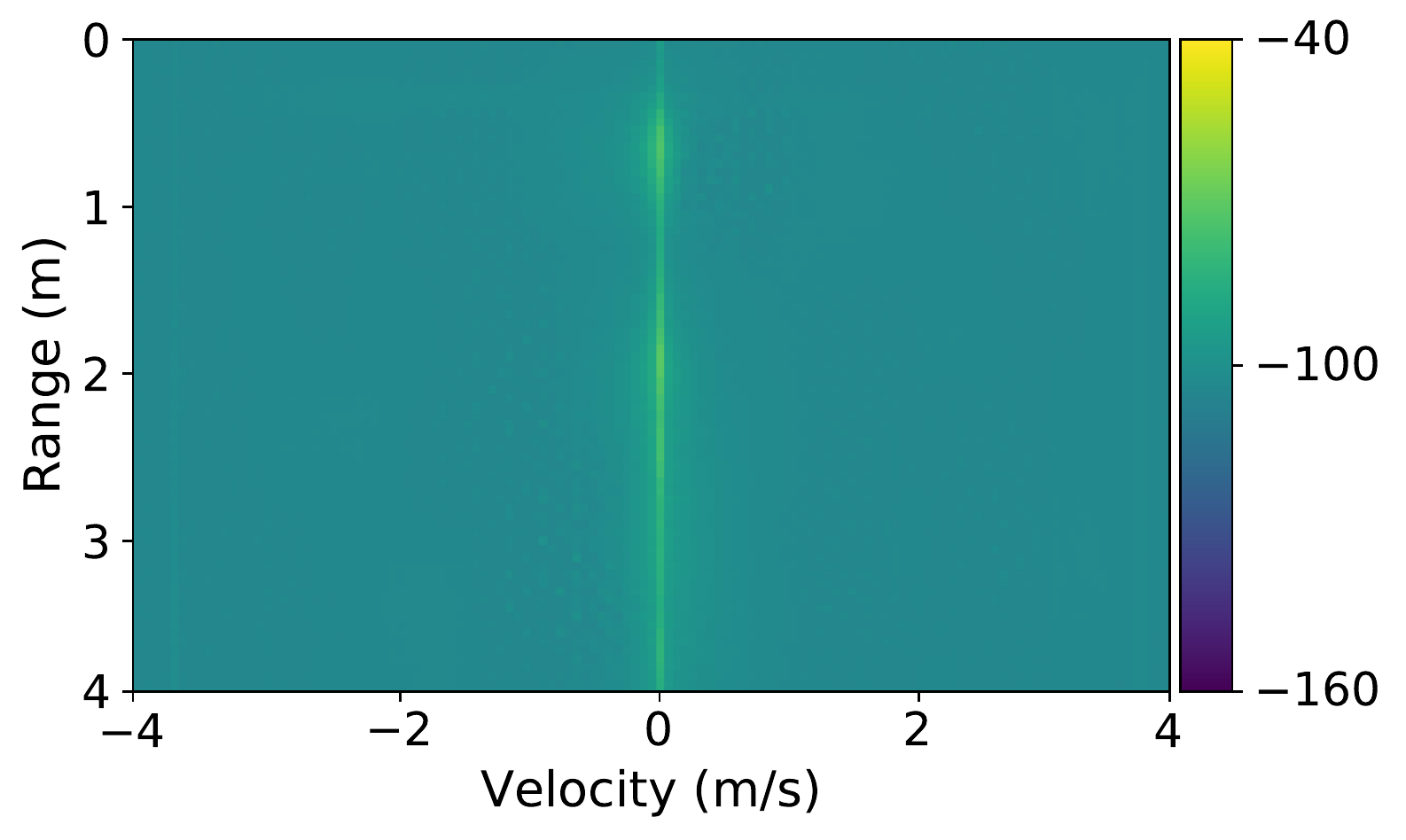}}}
\\\vspace{0.5em}
\rotatebox[origin=l]{90}{\phantom{----}Frame: 40}\hspace{0.4em}
\subfloat{{\includegraphics[width=4.2cm, height=2.6cm]{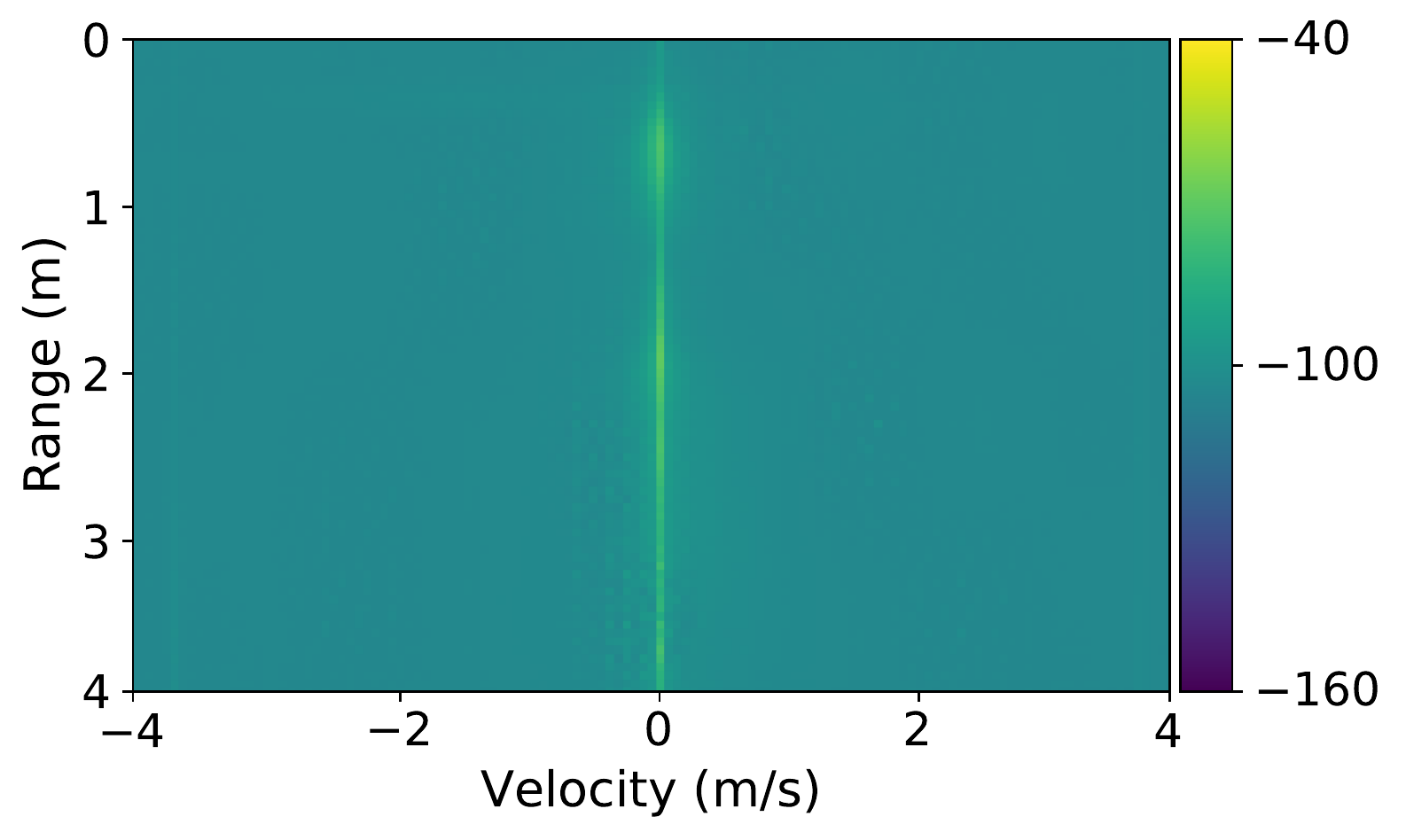}}}
\hspace{0.3em}
\rotatebox[origin=l]{90}{\phantom{----}Frame: 45}\hspace{0.4em}
\subfloat{{\includegraphics[width=4.2cm, height=2.6cm]{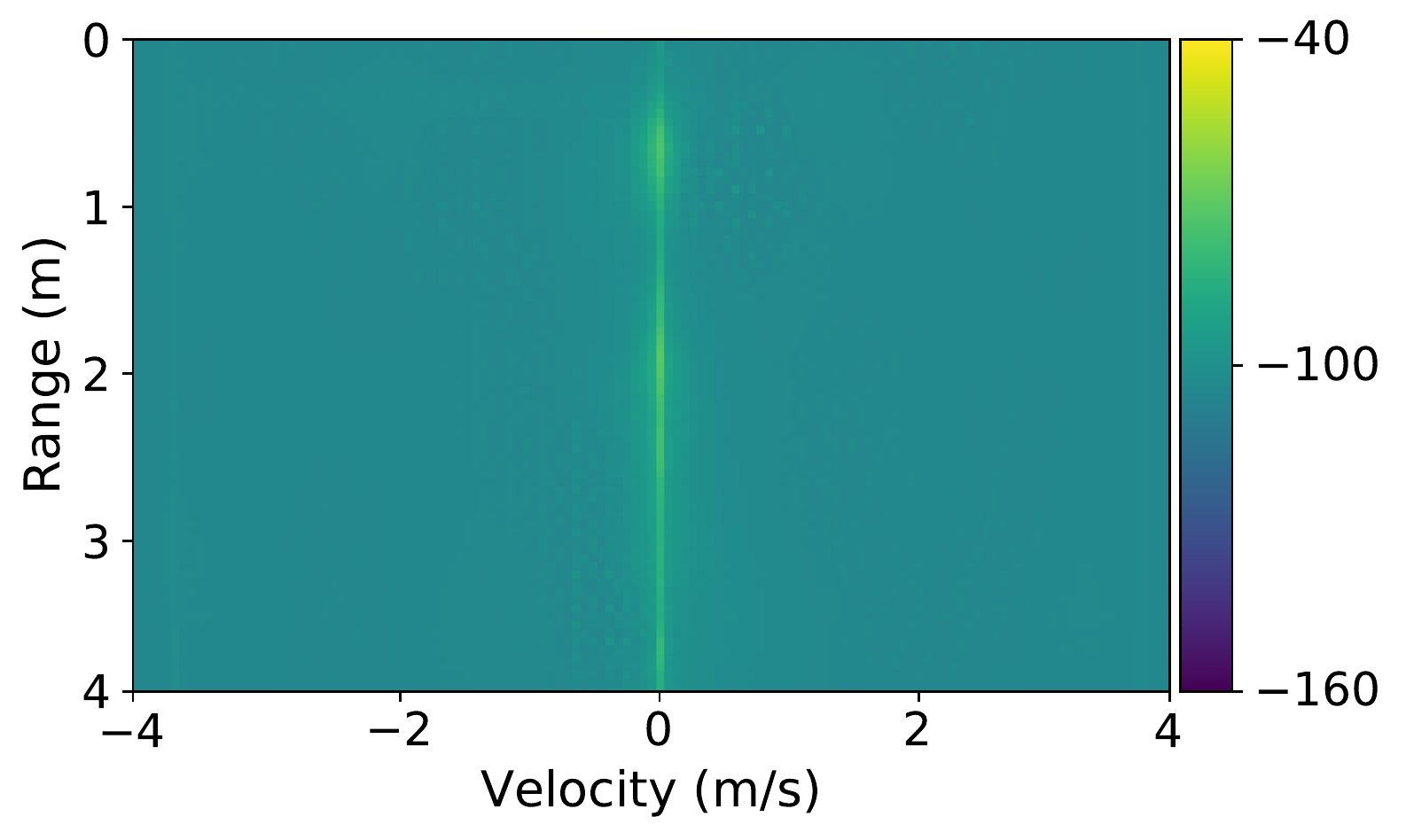}}}
\hspace{0.3em}
\rotatebox[origin=l]{90}{\phantom{----}Frame: 50}\hspace{0.4em}
\subfloat{{\includegraphics[width=4.2cm, height=2.6cm]{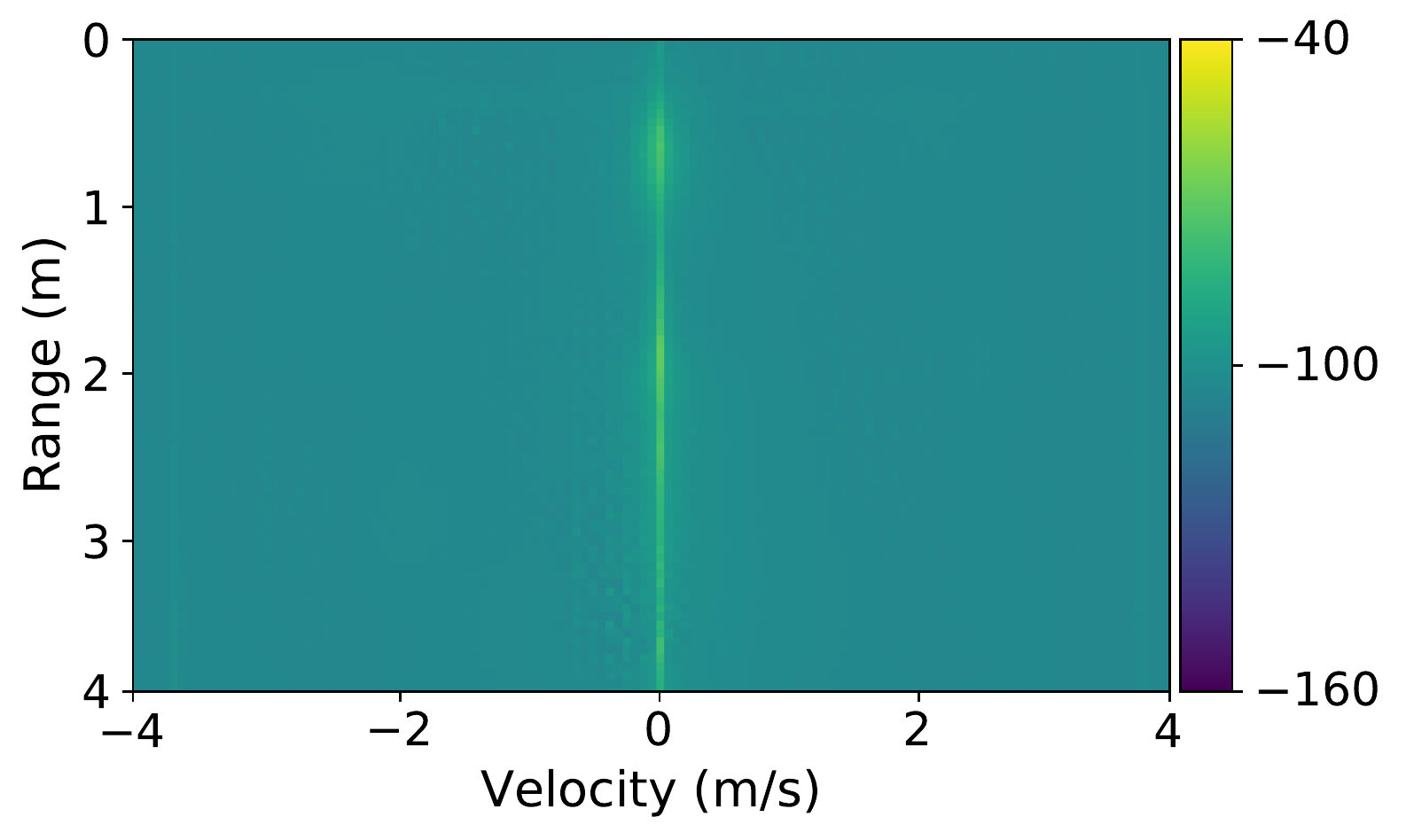}}}
\\\vspace{0.5em}
\caption{A sequence of adversarial padding frames (from frame $36$ to $50$), as generated by the method proposed in the main text. The total amount of adversarial perturbation, as calculated for these frames, corresponds to an $L_2$ distance of $4.75$.}
\label{fig:adversarial_padding}
\centering
\end{figure*}

Our experiments show that it is indeed possible to exploit the structure of data sets that contain a temporal dimension with special attacks similar to the above-described padding attack. In this case, we demonstrated the possibility of changing a model prediction by only perturbing the frames where the activity does not take place. Moreover, the adversarial padding generated by our padding attack only needs to be computed once and can then be used multiple times, thus allowing it to be incorporated in scenarios where the attacker has limited time for performing a malicious attack. In the supplementary materials, we provide a detailed illustration of adversarial padding in Fig.~\ref{fig:adversarial_padding}, showing how remarkably hard it is to spot adversarial padding using the bare eye.

In the next section, we discuss a number of interesting observations related to model interpretability, as made during our analysis of adversarial attacks on radar-based CNNs.

\section{Relation of Adversarial Attacks to Interpretability}
\label{Relation of Adversarial Attacks to Interpretability}

A major criticism regarding DNNs is their lack of interpretability; it is often challenging (if not impossible) to understand the reasoning behind the decisions made by a neural network-based model. In order to overcome this issue and to increase the trustworthiness of DNNs, several techniques have been proposed. These can broadly be divided into the following two groups: (1) perturbation-based forward propagation methods~\citep{DeepLift,ZeilerFergus_VisUnderstand} and (2) back-propagation-based approaches~\citep{grad_cam,simonyan,class_activation_map}. The main goal of these techniques is to highlight those parts of the input that are \textit{important} for the prediction made by a neural network. When the input consists of a natural image, this analysis is often done subjectively, unless the evaluated data comes with, for example, weakly-supervised localization labels, which can then be used for evaluating the correctness of the selected interpretability technique. In our case, different from natural images, the input consists of a sequence of RD frames that are significantly harder to interpret by humans. However, different from prediction using a single image, radar data also bring useful features, such as allowing for a frame-by-frame analysis.

\begin{figure}[t]
\centering
\begin{tikzpicture}[thick,scale=0.8, every node/.style={scale=0.8}]
\centering
\foreach \xtimes in {0,1,2,3}
{
\def\repetition{0.6}
\def\xpos{\xtimes*\repetition}
\def\ypos{0}
\def\incamt{1}
\draw[draw=black, opacity=0.5,dashed] (\xpos, \ypos) -- (\xpos+\incamt, \ypos+\incamt) -- (\xpos+\incamt, \ypos-\incamt) -- (\xpos, \ypos-2*\incamt) -- (\xpos, \ypos);
}
\node[] at (3.2, -0.5)  {\Large$\cdots$};   
\node[] at (3.2, 1.6)  {Radar data};
\foreach \xtimes in {6,7,8}
{
\def\repetition{0.6}
\def\xpos{\xtimes*\repetition}
\def\ypos{0}
\def\incamt{1}
\draw[draw=black, opacity=0.5,dashed] (\xpos, \ypos) -- (\xpos+\incamt, \ypos+\incamt) -- (\xpos+\incamt, \ypos-\incamt) -- (\xpos, \ypos-2*\incamt) -- (\xpos, \ypos);
}
\def\incamt{1.2}
\def\xposA{-0.25}
\def\yposA{0.1}
\draw[draw=black, opacity=1] (\xposA, \yposA) -- (\xposA+\incamt, \yposA+\incamt) -- (\xposA+\incamt, \yposA-\incamt) -- (\xposA, \yposA-2*\incamt) -- (\xposA, \yposA);
\def\xposB{5}
\def\yposB{0.1}
\draw[draw=black, opacity=1] (\xposB, \yposB) -- (\xposB+\incamt, \yposB+\incamt) -- (\xposB+\incamt, \yposB-\incamt) -- (\xposB, \yposB-2*\incamt) -- (\xposB, \yposB);

\draw[draw=black, opacity=1] (\xposB, \yposB-\incamt*2) -- (\xposA, \yposB-\incamt*2);
\draw[draw=black, opacity=1] (\xposB, \yposB) -- (\xposA, \yposB);
\draw[draw=black, opacity=1] (\xposB+\incamt, \yposB+\incamt) -- (\xposA+\incamt, \yposB+\incamt);
\draw[draw=black, opacity=0.26] (\xposB+\incamt, \yposB-\incamt) -- (\xposA+\incamt, \yposB-\incamt);

\draw[-{Latex[width=1.5mm]}, color=red, opacity=0.7] (4, -1.6) arc (0:-40:2cm);
\draw[{Latex[width=1.5mm]}-, color=blue] (4, -1.6) arc (0:40:-2cm);

\def\xpos{1.55}
\def\ypos{-2.75}
\node[inner sep=0pt] (vid2) at (\xpos + 0.6, \ypos-0.6)
    {\includegraphics[width=.15\textwidth]{radar_ims/10_org_im.pdf}};
  
\def\xpos{5.4}
\def\ypos{-2.75}
\node[inner sep=0pt] (vid2) at (\xpos + 0.6, \ypos-0.6)
    {\includegraphics[width=.15\textwidth]{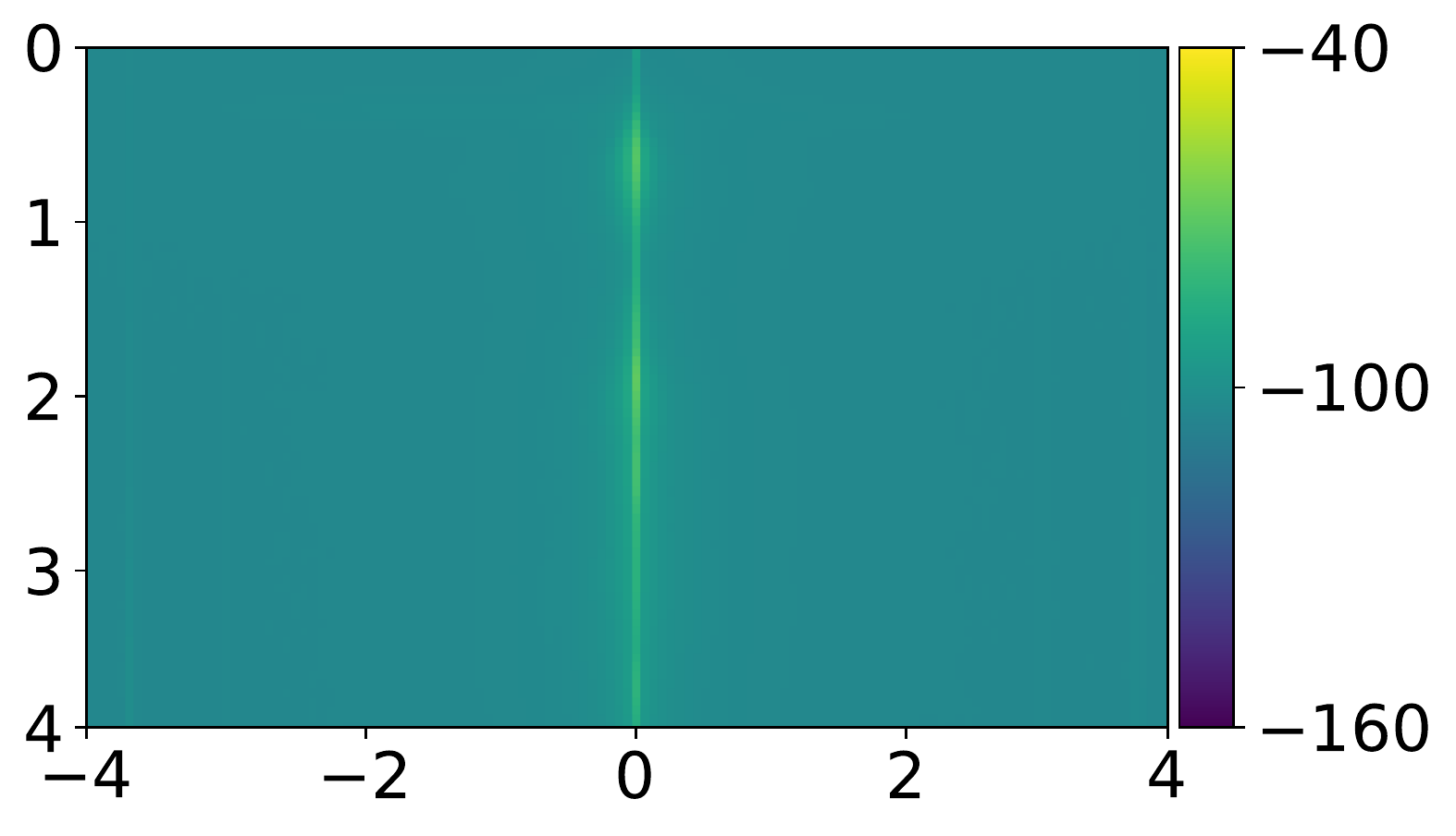}};

\def\xpos{7.4}
\def\ypos{0.6}
\def\height{2.2}
\draw[draw=black] (\xpos, \ypos) -- (\xpos+1, \ypos) -- (\xpos+1, \ypos-\height) -- (\xpos, \ypos-\height) -- (\xpos, \ypos);
\node[align=center,rotate=90] at (\xpos+0.5, \ypos-\height/2) {\small  CNN Model};
\draw[-{Latex[width=1.5mm]}, line width=0.5mm] (\xpos-1, -0.5) -- (\xpos-0.2, -0.5);
\draw[-{Latex[width=1.5mm]}, line width=0.5mm] (\xpos+1.3, -0.5) -- (\xpos+2, -0.5);

\node[align=center] at (\xpos+2.5, \ypos-\height/2) {\large$\left[\begin{array}{@{}c@{}}
    y_{0} \\
    y_{1} \\
    \vdots \\
    y_{5} 
    \end{array}\right]$};
\end{tikzpicture}
\caption{Visual representation of the frame replacement operation as explained in Section~\ref{Relation of Adversarial Attacks to Interpretability}.}
\label{fig:frame-replacement-vis}

\vspace{1em}
\centering
\includegraphics[width=8.5cm]{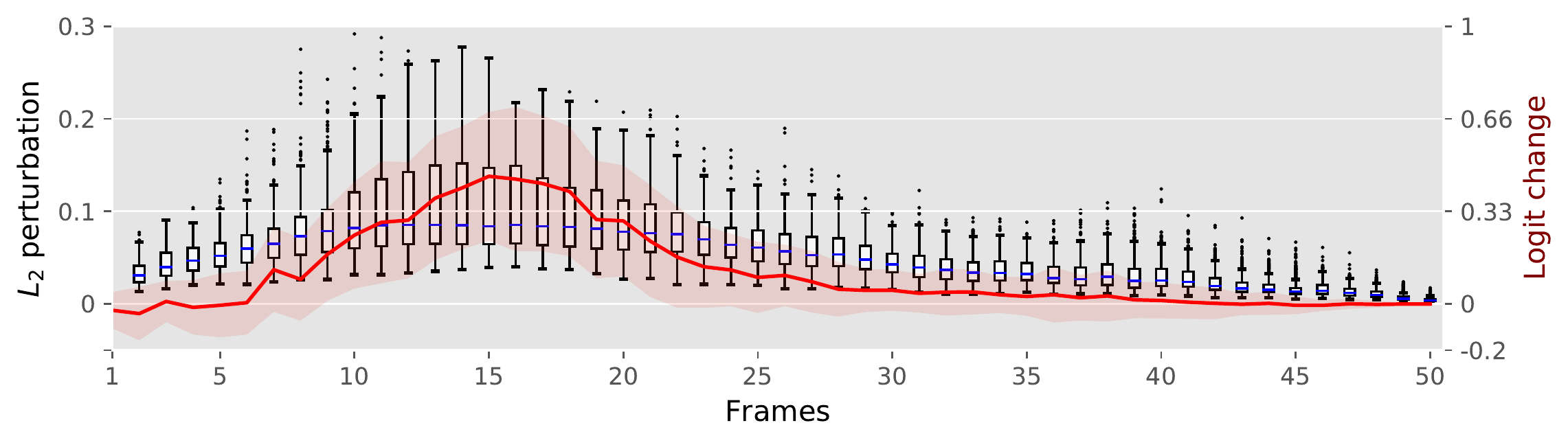}
\caption{A boxplot representation of added perturbation, as generated by CW, displayed for individual frames of adversarial examples that transfer from ${A}_{S_+}$ to ${A}_{S_-}$. The amount of added perturbation is plotted against the median frame importance, as calculated by the experiment detailed in Section~\ref{Relation of Adversarial Attacks to Interpretability}.}
\label{fig:frame-perturb-boxplot}

\vspace{1em}
\centering
\includegraphics[width=8.5cm]{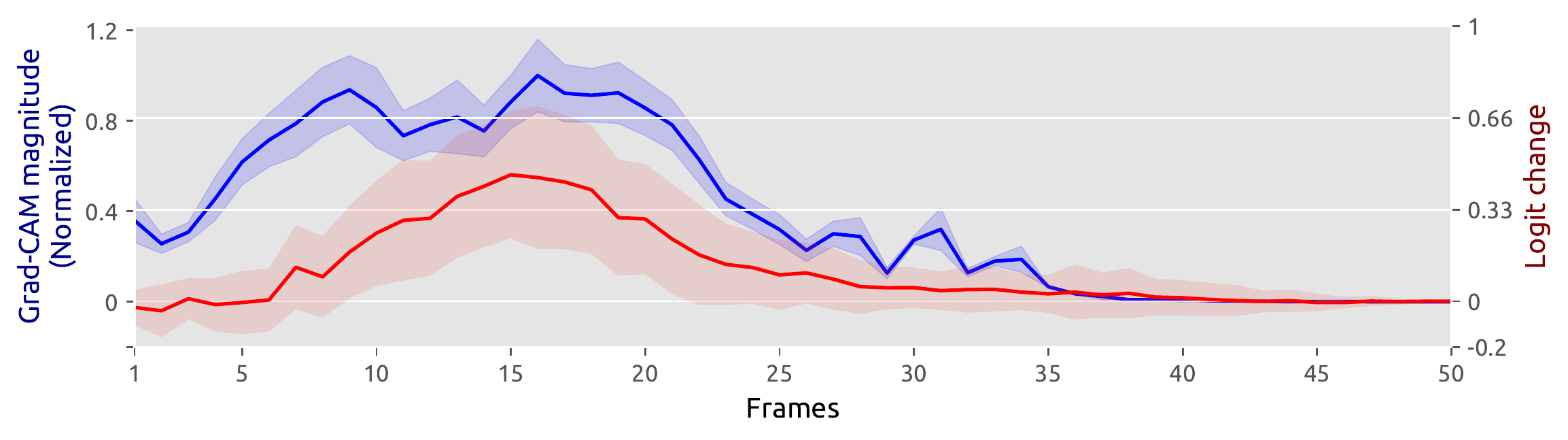}
\caption{Mean Grad-CAM magnitude (normalized) is plotted against the mean frame importance, as calculated by the experiment detailed in Section~\ref{Relation of Adversarial Attacks to Interpretability}.}
\label{fig:frame-gradcam-line}
\end{figure}

A first peculiar observation we made during the experiments presented in Section~\ref{Threat Model Evaluation} is that CW focuses on only introducing perturbation in certain frames, rather than spreading out the perturbation equally. In~Fig.~\ref{fig:frame-perturb-boxplot}, we present the amount of perturbation added by CW to each frame in the form of boxplots. Note that padding frames at the end receive considerably less perturbation than the frames containing the action. We hypothesize that frames that are the recipient of stronger perturbations are \textit{important} frames, making it possible to distinguish actions from one another.

\begin{figure}[t]
\centering
\subfloat[Aggregate]{\includegraphics[width=8.5cm]{other_images/Agg_frame_perturbation.pdf}}\\\vspace{-0.5em}
\subfloat[Activity (0)]{\includegraphics[width=8.5cm]{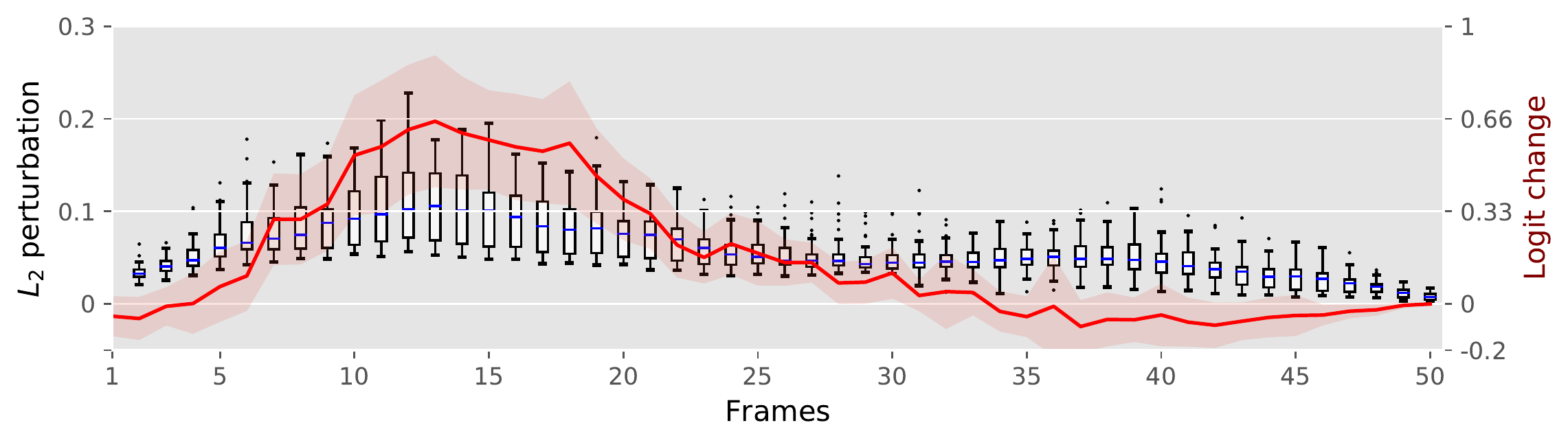}}\\\vspace{-0.5em}
\subfloat[Activity (1)]{\includegraphics[width=8.5cm]{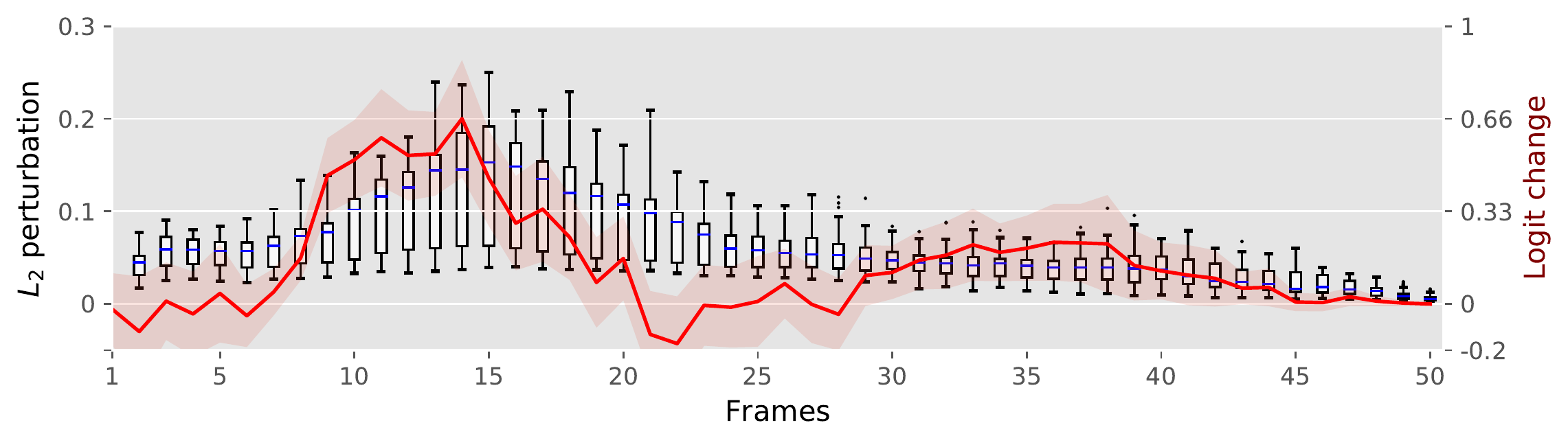}}\\\vspace{-0.5em}
\subfloat[Activity (2)]{\includegraphics[width=8.5cm]{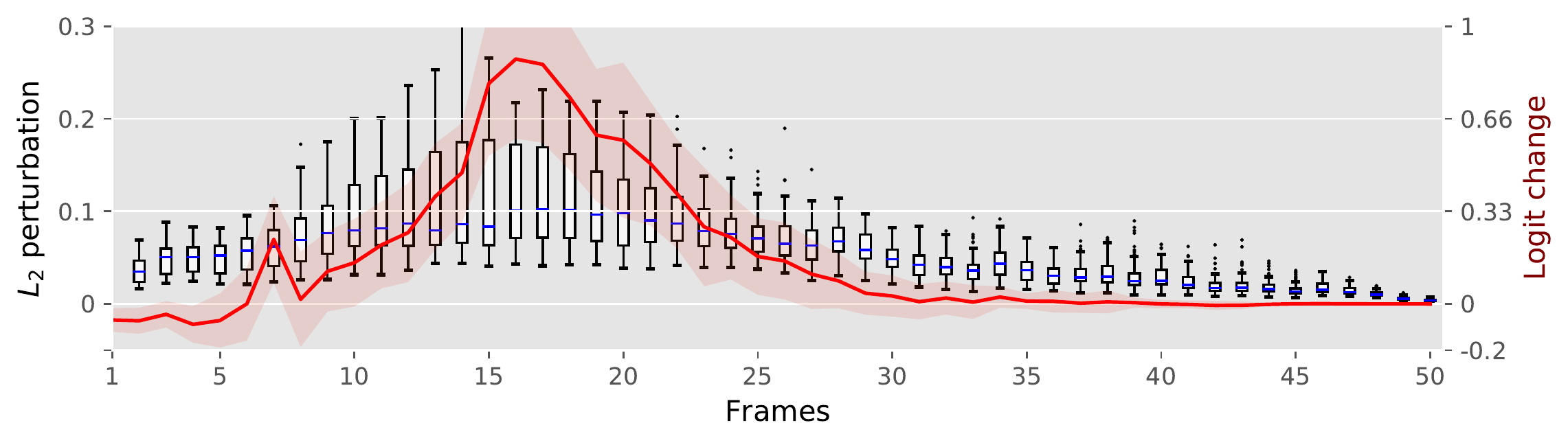}}\\\vspace{-0.5em}
\subfloat[Activity (3)]{\includegraphics[width=8.5cm]{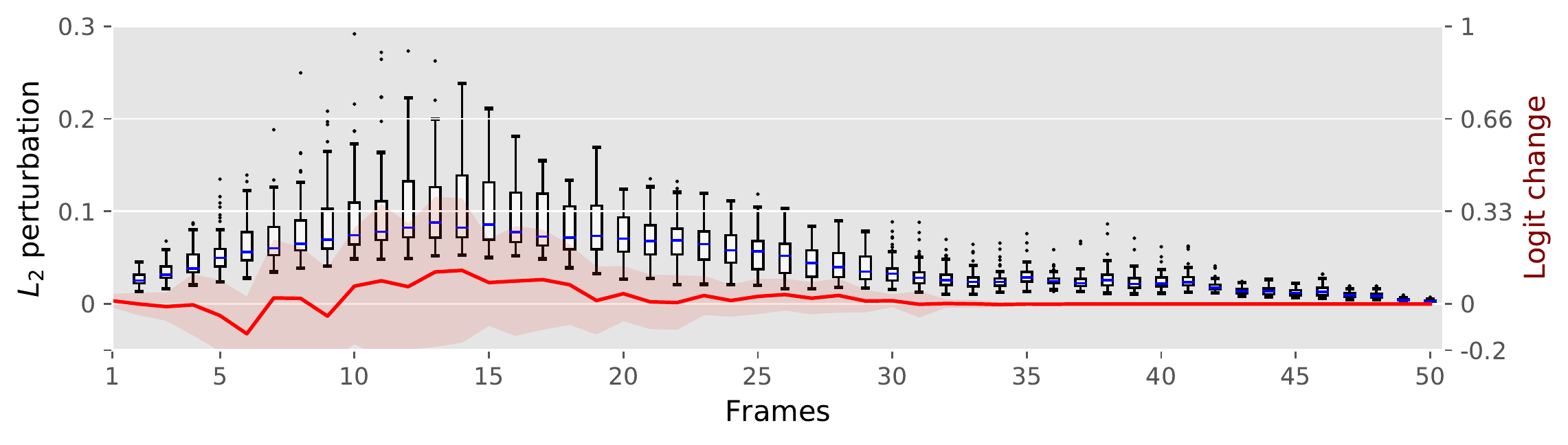}}\\\vspace{-0.5em}
\subfloat[Activity (4)]{\includegraphics[width=8.5cm]{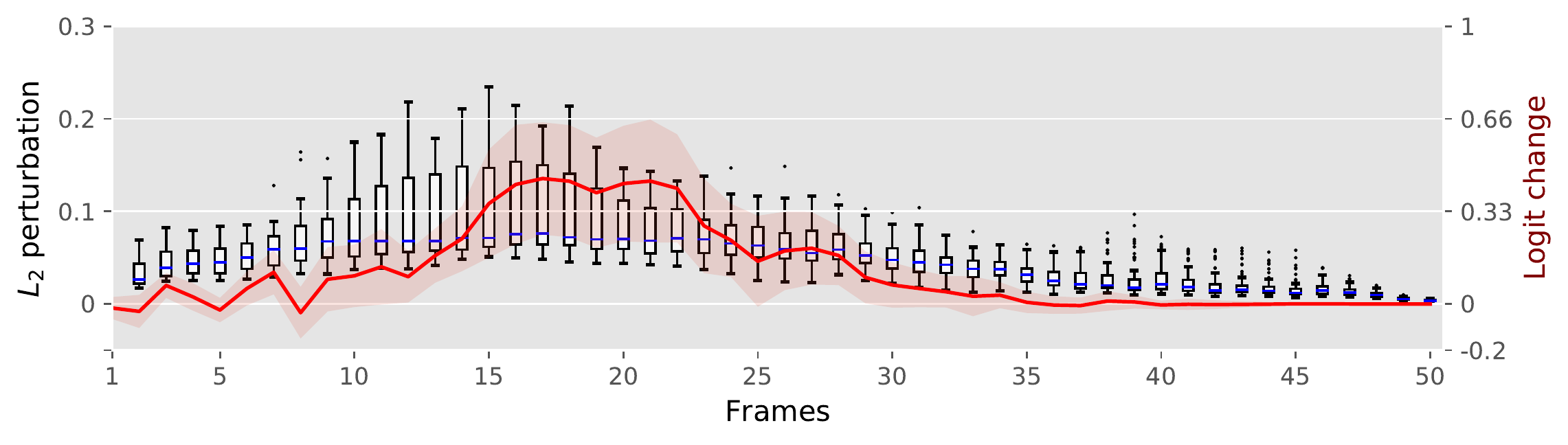}}\\\vspace{-0.5em}
\subfloat[Activity (5)]{\includegraphics[width=8.5cm]{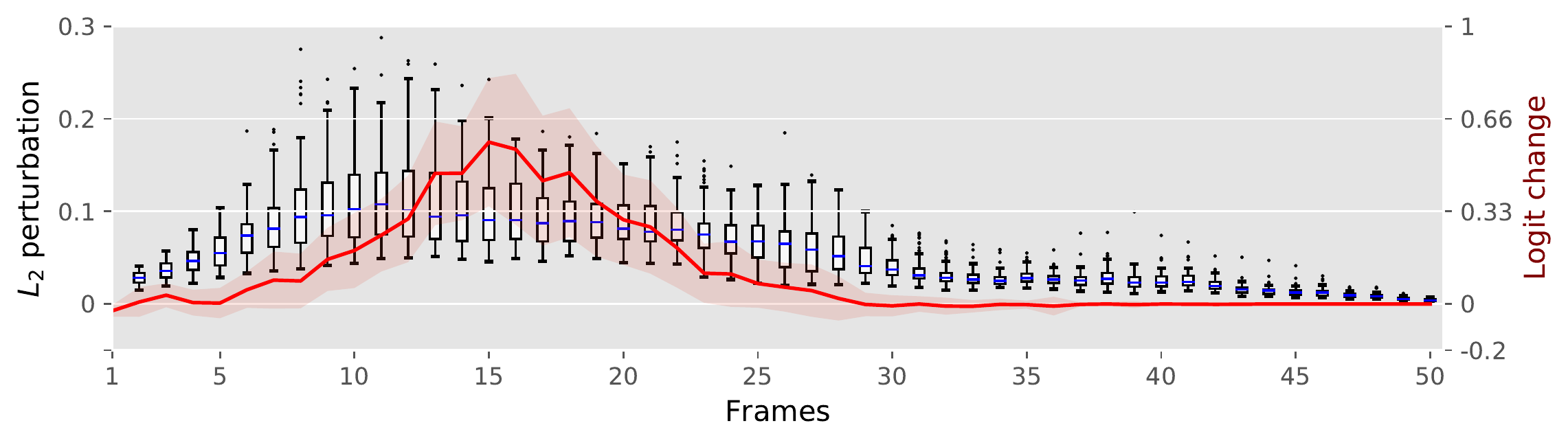}}
\caption{A boxplot representation of added perturbation, as generated by CW, displayed for individual frames that belong to adversarial examples that transfer from ${A}_{S_+}$ to ${A}_{S_-}$. The amount of added perturbation is plotted against the median frame importance, as calculated by the experiment detailed in Section~\ref{Relation of Adversarial Attacks to Interpretability} of the main text. The sub-figure labeled with (a) shows this graph for all of the adversarial examples in an aggregated manner, whereas the following sub-figures are class-specific. The line graph represents the median, whereas the shaded area represents the interquartile range.}
\label{fig:frame-perturb-boxplot-ext}
\vspace{-1em}
\end{figure}

In order to confirm this hypothesis, we perform an exhaustive experiment on measuring the importance of a frame. As illustrated in Fig.~\ref{fig:frame-replacement-vis}, we replace individual frames, one at a time, by the median frame we described in Section~\ref{Framework}, subsequently performing a forward pass. Since this median frame is used throughout the training procedure to pad the data, it is not an out-of-distribution sample, thus not favoring one class over another. By doing so, for each data point, we measure the change in the prediction logit for the correct class $50$ times (for each frame individually) and plot the median difference in Fig.~\ref{fig:frame-perturb-boxplot}, showing the relation between the perturbation amount per frame and the logit change when those frames are replaced. Specifically, the red line represents the median logit change and the shaded area represents the interquartile range. As can be observed, the frames favored by adversarial attacks in terms of added perturbation are also the ones that contribute more to the prediction, confirming our hypothesis. In the supplementary materials, an extended version of this experiment, conducted on each class individually, can be found in Fig.~\ref{fig:frame-perturb-boxplot-ext}.

\begin{figure*}[t!]
\centering
\rotatebox[origin=l]{90}{\phantom{--}Frame: 1}\hspace{0.2em}
\subfloat{{\frame{\includegraphics[width=3.5cm, height=2cm]{radar_ims/001_mask.png}}}}
\hspace{0.4em}
\subfloat{{{\includegraphics[width=3.5cm, height=2cm]{radar_ims/1_org_im.pdf}}}}
\hspace{0.2em}
\subfloat{{{\includegraphics[width=3.5cm, height=2cm]{radar_ims/1_cam_pos.pdf}}}}
\\\vspace{0.5em}
\rotatebox[origin=l]{90}{\phantom{--}Frame: 5}\hspace{0.2em}
\subfloat{{\frame{\includegraphics[width=3.5cm, height=2cm]{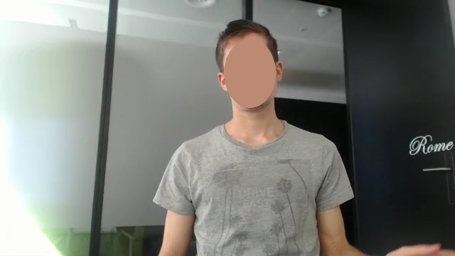}}}}
\hspace{0.4em}
\subfloat{{{\includegraphics[width=3.5cm, height=2cm]{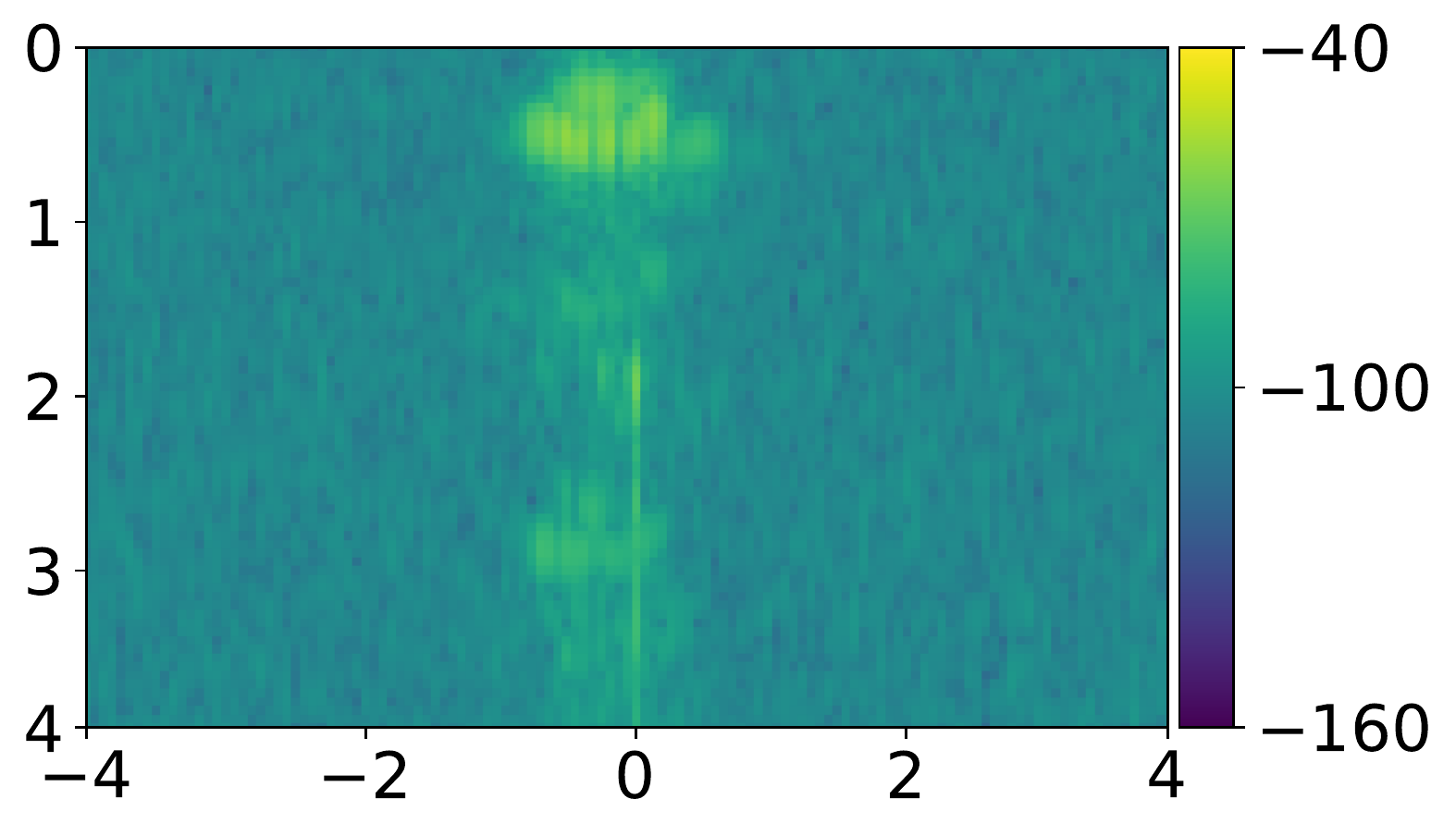}}}}
\hspace{0.2em}
\subfloat{{{\includegraphics[width=3.5cm, height=2cm]{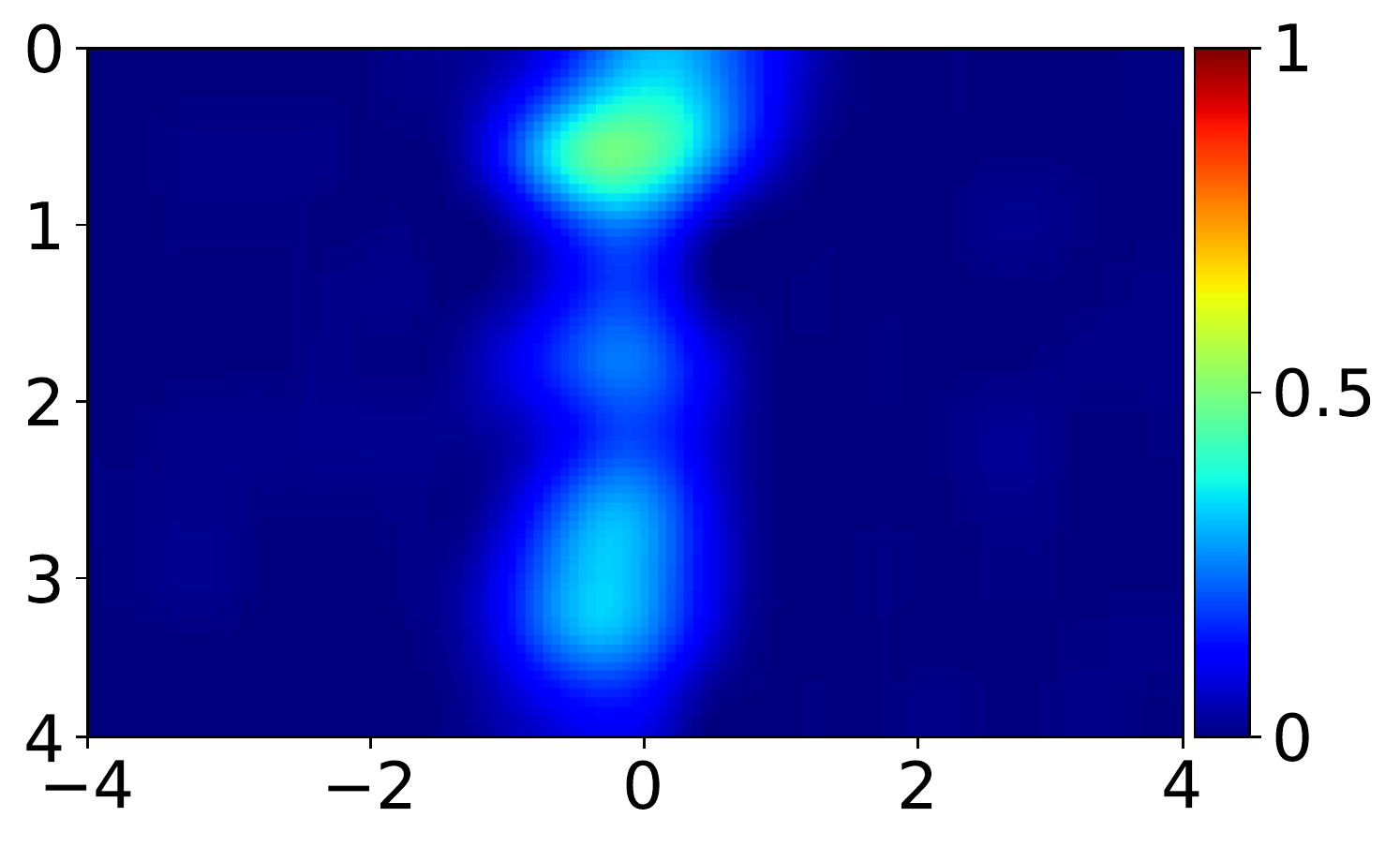}}}}
\\\vspace{0.5em}
\rotatebox[origin=l]{90}{\phantom{--}Frame: 7}\hspace{0.2em}
\subfloat{{\frame{\includegraphics[width=3.5cm, height=2cm]{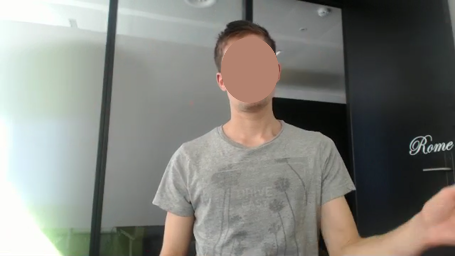}}}}
\hspace{0.4em}
\subfloat{{{\includegraphics[width=3.5cm, height=2cm]{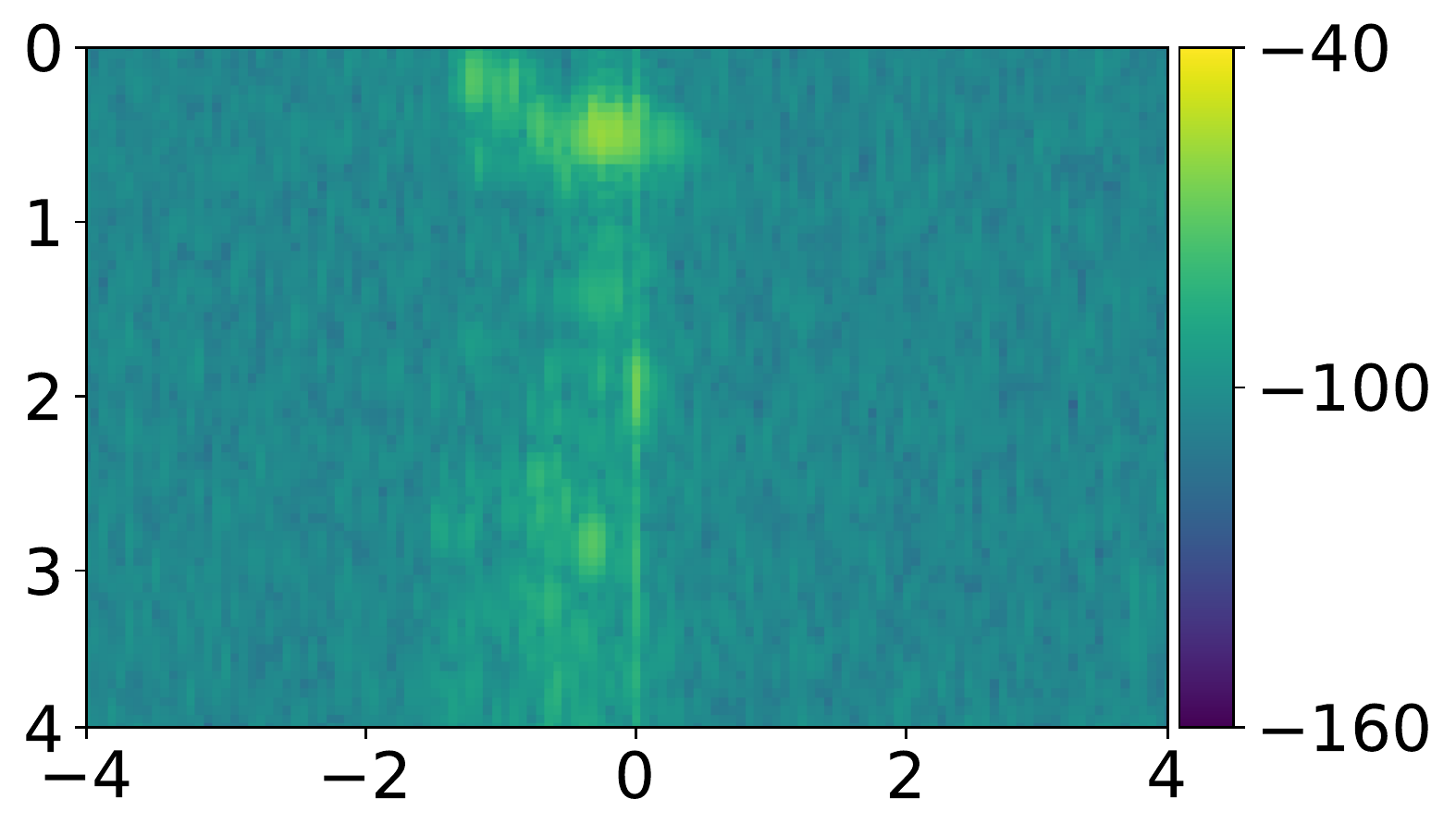}}}}
\hspace{0.2em}
\subfloat{{{\includegraphics[width=3.5cm, height=2cm]{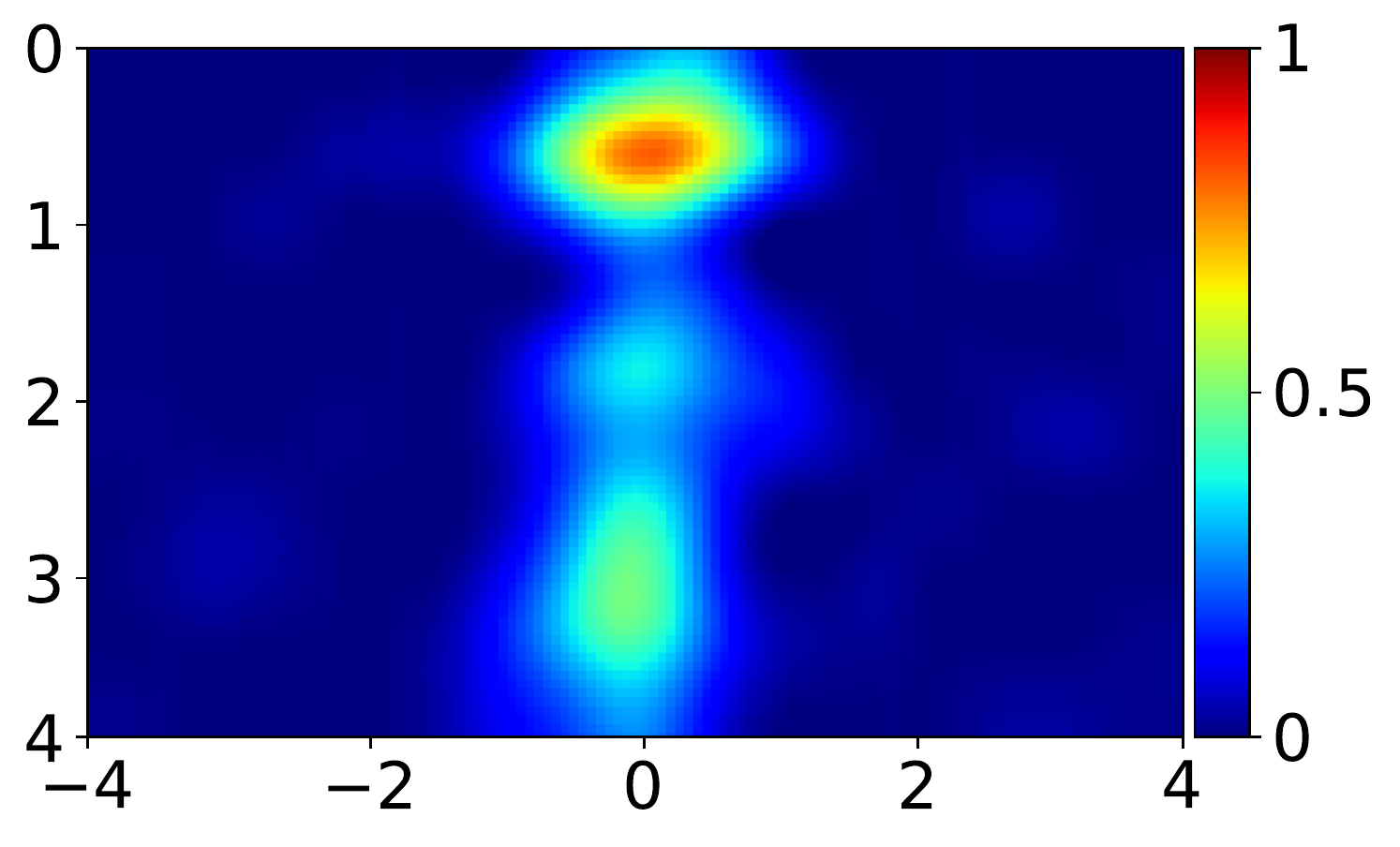}}}}
\\\vspace{0.5em}
\rotatebox[origin=l]{90}{\phantom{-}Frame: 10}\hspace{0.2em}
\subfloat{{\frame{\includegraphics[width=3.5cm, height=2cm]{radar_ims/010_mask.png}}}}
\hspace{0.4em}
\subfloat{{{\includegraphics[width=3.5cm, height=2cm]{radar_ims/10_org_im.pdf}}}}
\hspace{0.2em}
\subfloat{{{\includegraphics[width=3.5cm, height=2cm]{radar_ims/10_cam_pos.pdf}}}}
\\\vspace{0.5em}
\rotatebox[origin=l]{90}{\phantom{-}Frame: 11}\hspace{0.2em}
\subfloat{{\frame{\includegraphics[width=3.5cm, height=2cm]{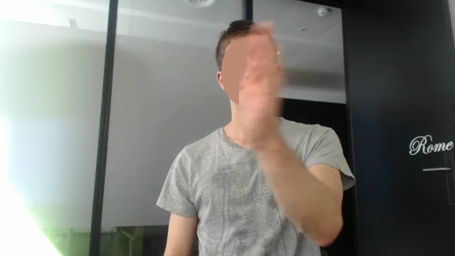}}}}
\hspace{0.4em}
\subfloat{{{\includegraphics[width=3.5cm, height=2cm]{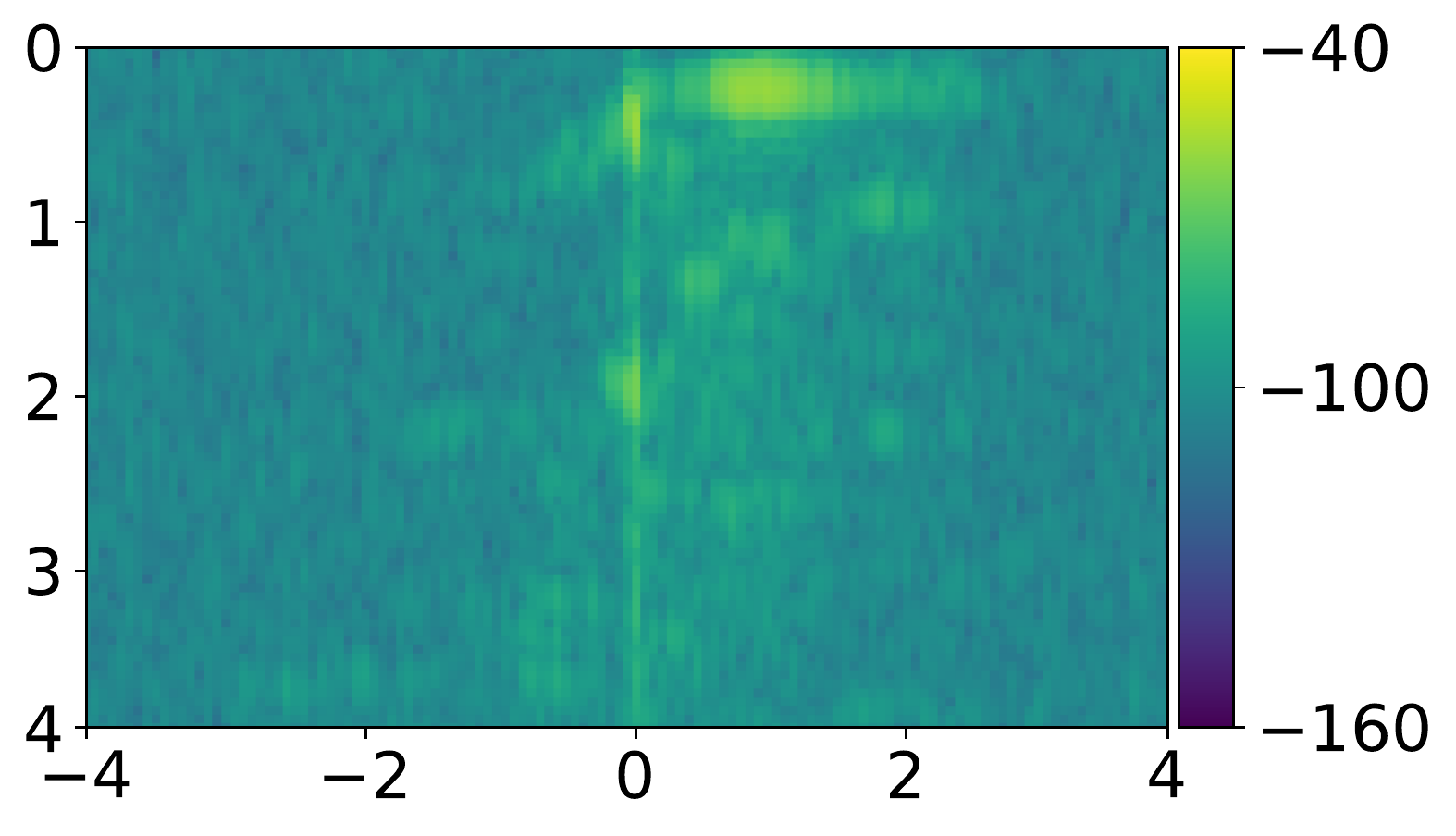}}}}
\hspace{0.2em}
\subfloat{{{\includegraphics[width=3.5cm, height=2cm]{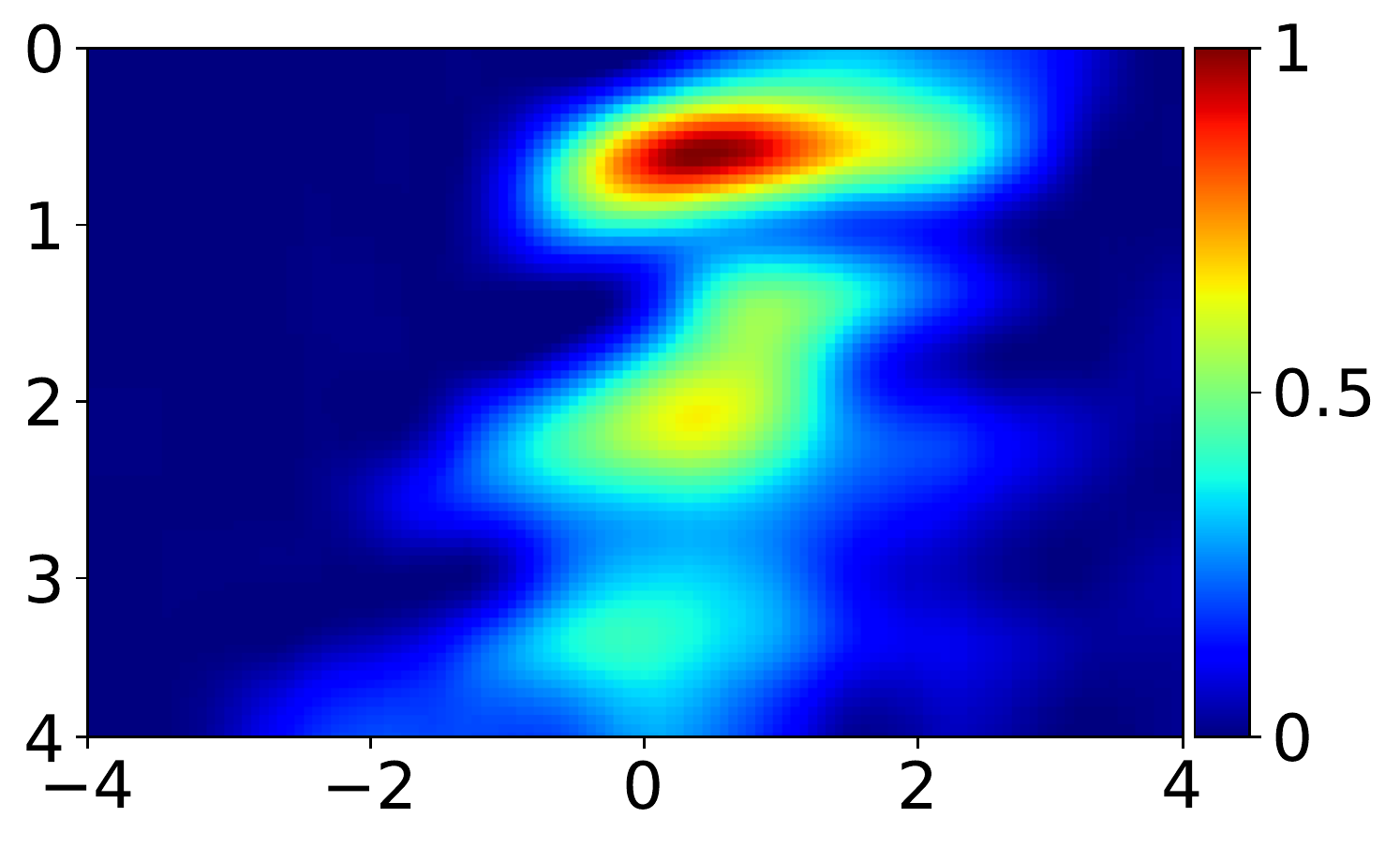}}}}
\\\vspace{0.5em}
\rotatebox[origin=l]{90}{\phantom{-}Frame: 13}\hspace{0.2em}
\subfloat{{\frame{\includegraphics[width=3.5cm, height=2cm]{radar_ims/013_mask.png}}}}
\hspace{0.4em}
\subfloat{{{\includegraphics[width=3.5cm, height=2cm]{radar_ims/13_org_im.pdf}}}}
\hspace{0.2em}
\subfloat{{{\includegraphics[width=3.5cm, height=2cm]{radar_ims/13_cam_pos.pdf}}}}
\\\vspace{0.5em}
\rotatebox[origin=l]{90}{\phantom{-}Frame: 16}\hspace{0.2em}
\subfloat{{\frame{\includegraphics[width=3.5cm, height=2cm]{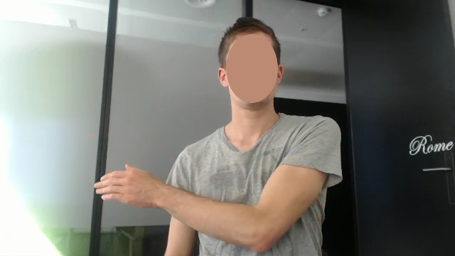}}}}
\hspace{0.4em}
\subfloat{{{\includegraphics[width=3.5cm, height=2cm]{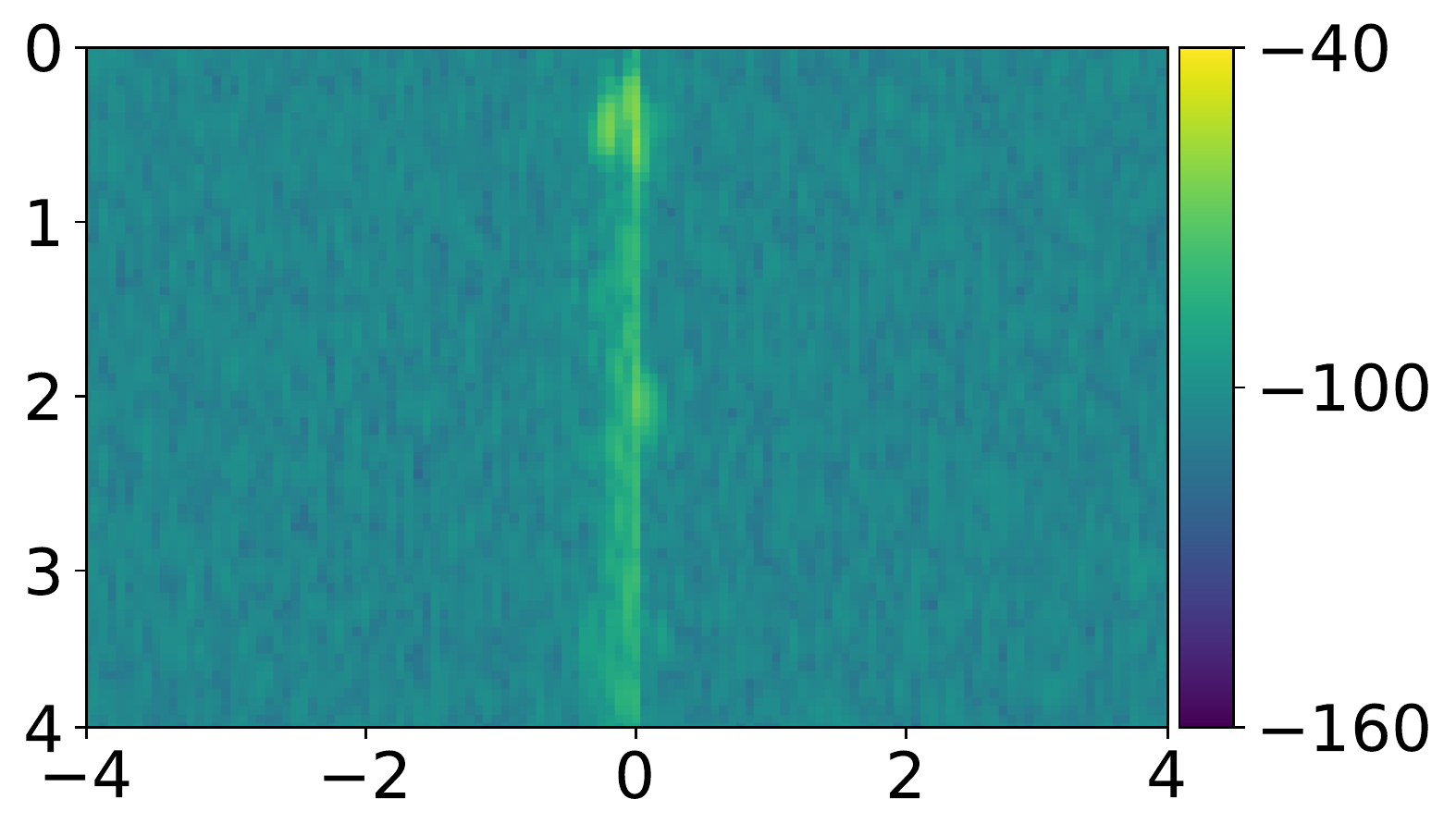}}}}
\hspace{0.2em}
\subfloat{{{\includegraphics[width=3.5cm, height=2cm]{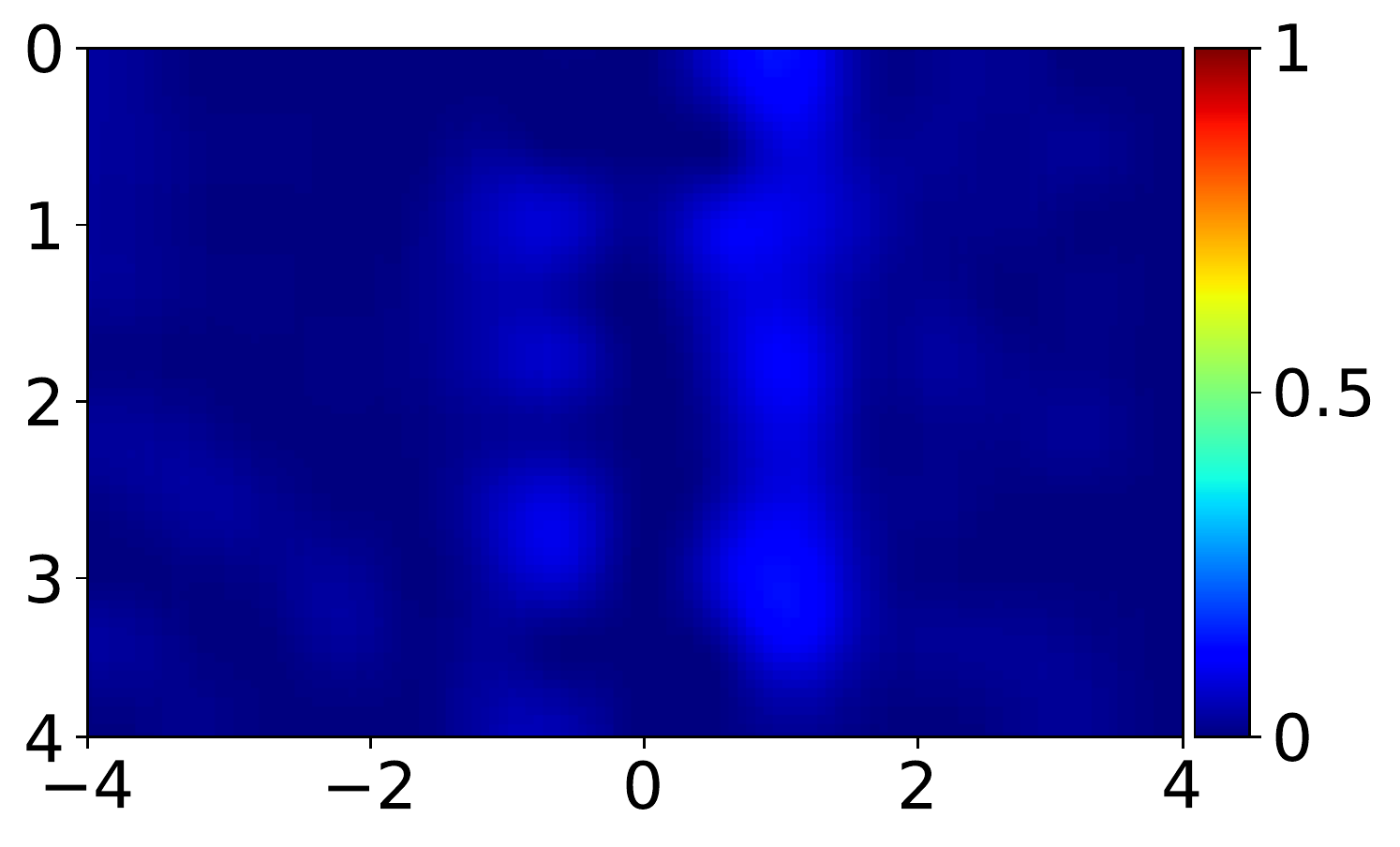}}}}
\\\vspace{0.5em}
\rotatebox[origin=l]{90}{\phantom{-}Frame: 18}\hspace{0.2em}
\subfloat{{\frame{\includegraphics[width=3.5cm, height=2cm]{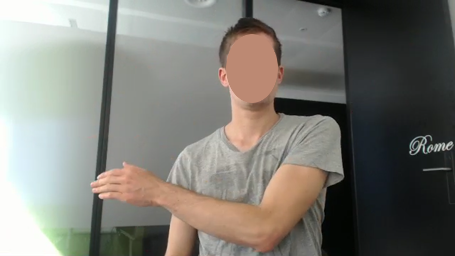}}}}
\hspace{0.4em}
\subfloat{{{\includegraphics[width=3.5cm, height=2cm]{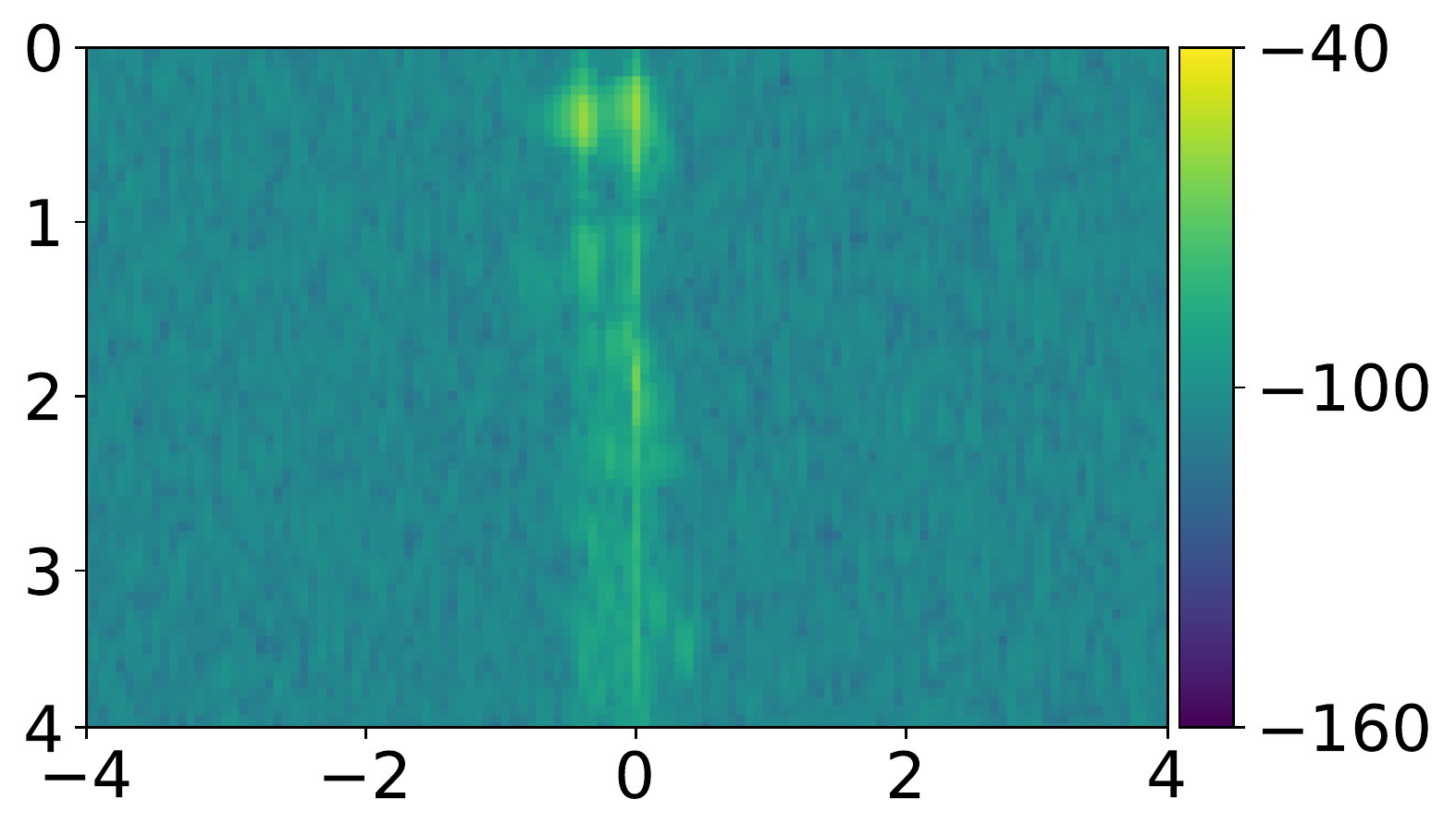}}}}
\hspace{0.2em}
\subfloat{{{\includegraphics[width=3.5cm, height=2cm]{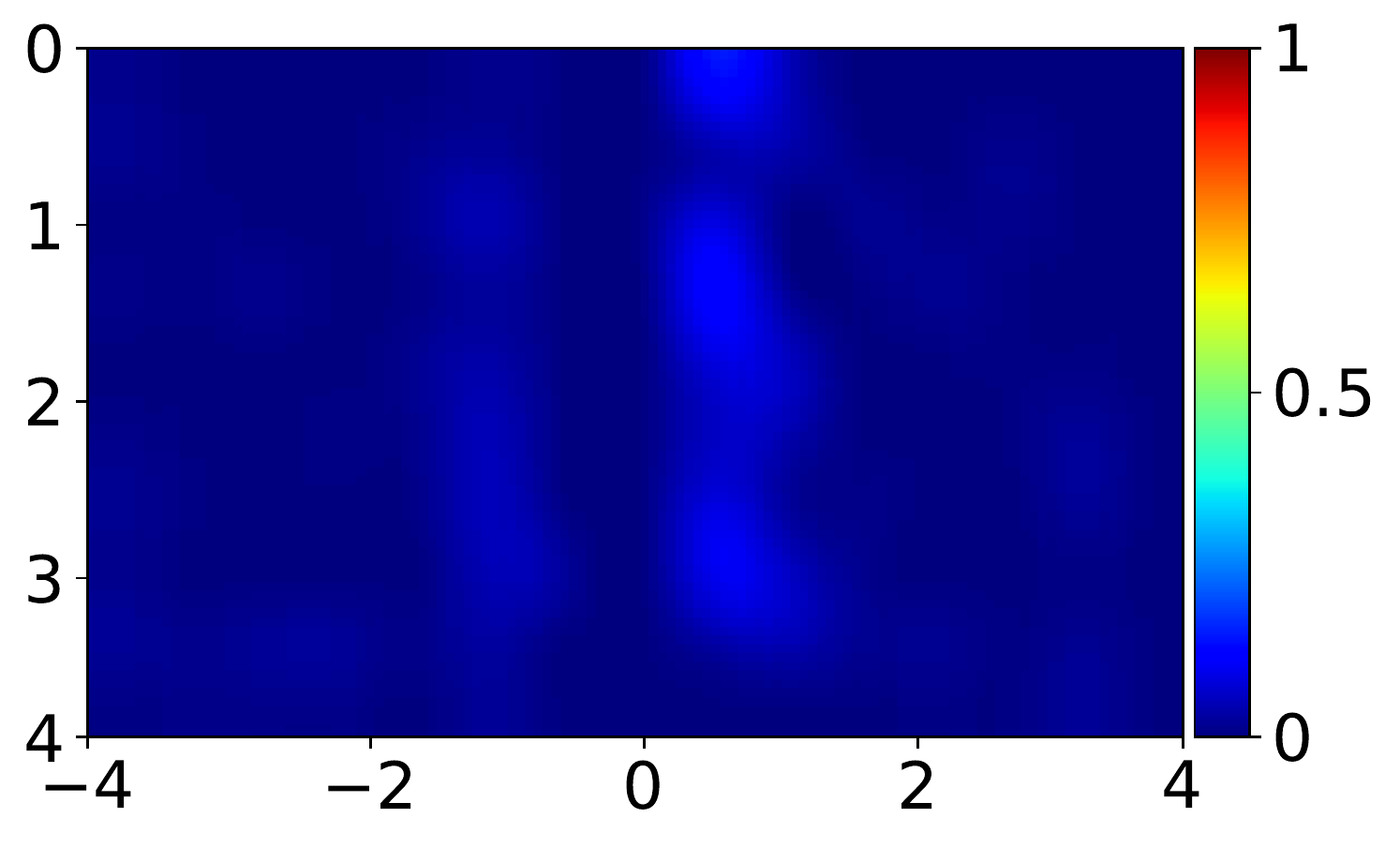}}}}
\\\vspace{0.5em}
\rotatebox[origin=l]{90}{\phantom{-}Frame: 19}\hspace{0.2em}
\subfloat{{\frame{\includegraphics[width=3.5cm, height=2cm]{example-image-a}}}}
\hspace{0.4em}
\subfloat{{{\includegraphics[width=3.5cm, height=2cm]{radar_ims/19_org_im.pdf}}}}
\hspace{0.2em}
\subfloat{{{\includegraphics[width=3.5cm, height=2cm]{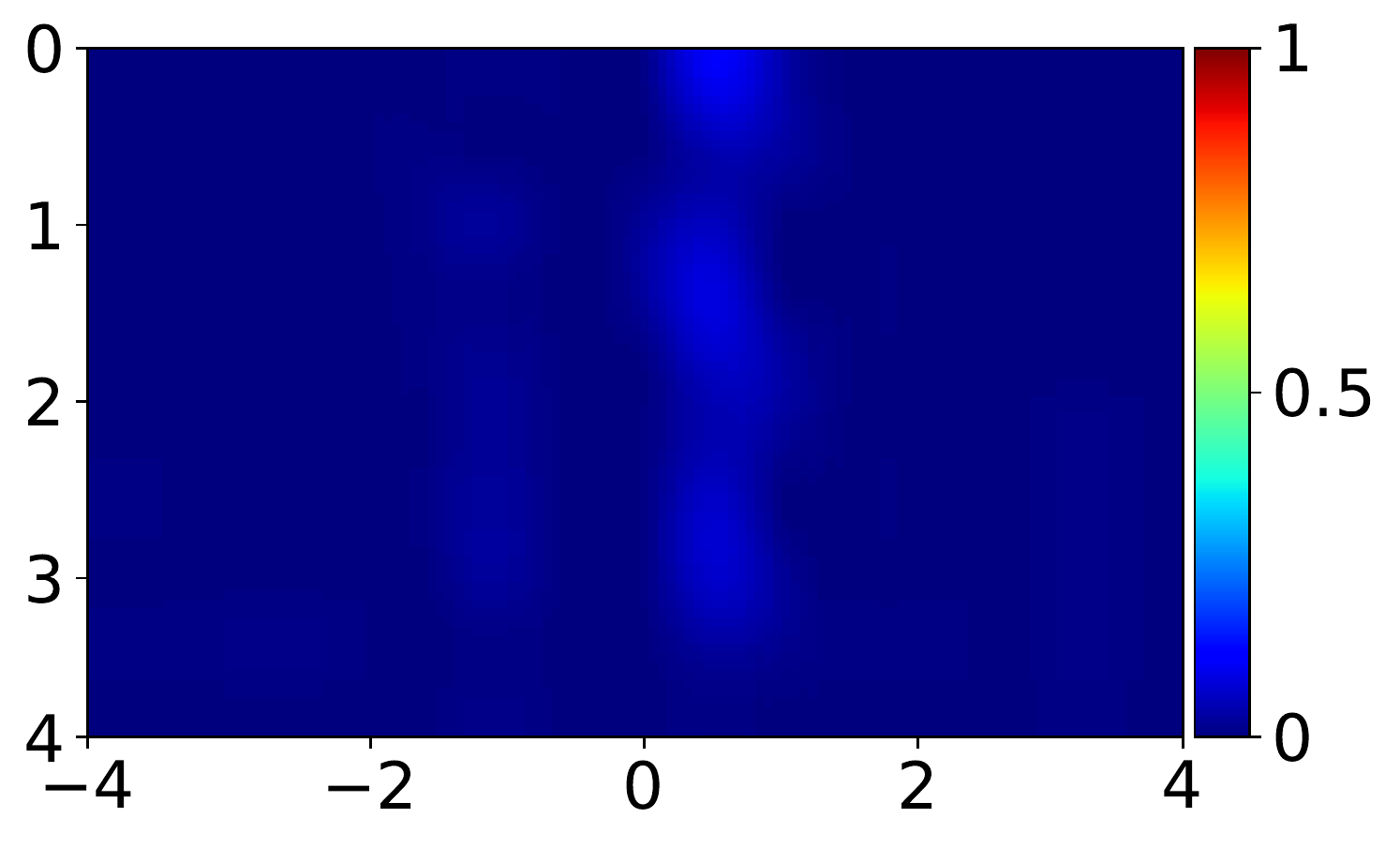}}}}

\caption{Visualization of a number of consecutive (1) RD frames, (2) video frames, and (3) Grad-CAM heatmaps for the action \textit{swipe left}. The $X$-axis and $Y$-axis of the RD frames and the Grad-CAM images, which are omitted for visual clarity, correspond to range and velocity, respectively.}
\label{fig:detailed_swipe_left}
\centering
\end{figure*}

Following this experiment, we investigate the applicability of CNN interpretability techniques to radar data. Among different interpretability techniques, Grad-CAM~\citep{grad_cam} stands out, thanks to its superior weakly-supervised localization results obtained on ImageNet. Another reason for selecting this method is that its approach is based on backpropagation, meaning that the input is not perturbed. We especially want to avoid methods based on input perturbation because, unlike natural images, small changes in RD frames may lead to large changes in terms of correctness of the data (i.e., being a valid data point). In our setting, Grad-CAM is defined as follows:
\begin{align}
    \text{Grad\textendash CAM} = \sum_k \Big( \, ReLU \big( \sum_i \sum_j \nabla_x \mathbf{L}^{p}_{i,j} \big) \mathbf{L}^ {p}_{k} \Big) \,, \label{eq:gradcam1} 
\end{align}
where $\mathbf{L}^p$ denotes the output of the forward pass after the $p$-th layer (i.e., discriminative features) and $\nabla_x \mathbf{L}^p$ denotes the gradient obtained with a backward pass from the same layer with respect to the input (i.e., weighted gradient). Different from adversarial attacks, as well as vanilla and guided backpropagation, Grad-CAM does not use the gradients of the first layer, thus arguably allowing for a more robust explanatory approach. Because the input is not a single image but a sequence of frames, Grad-CAM produces class activation maps for each frame individually. An example set of video frames, their corresponding radar frames, and the obtained Grad-CAM heatmaps can be found in Fig.~\ref{fig:gracam_swipe_left}. In the supplementary materials, an extended version of the same activity sequence is provided in Fig.~\ref{fig:detailed_swipe_left}. These qualitative results show that the heatmaps usually highlight (1) those frames where the \textit{most important} part of the activity occurs and (2) those locations where the radar activity is the largest.

\begin{figure}[t]
\centering
\subfloat[Aggregate]{\includegraphics[width=8.5cm]{other_images/Agg_gradcam_frame.pdf}}\\\vspace{-0.5em}
\subfloat[Activity (0)]{\includegraphics[width=8.5cm]{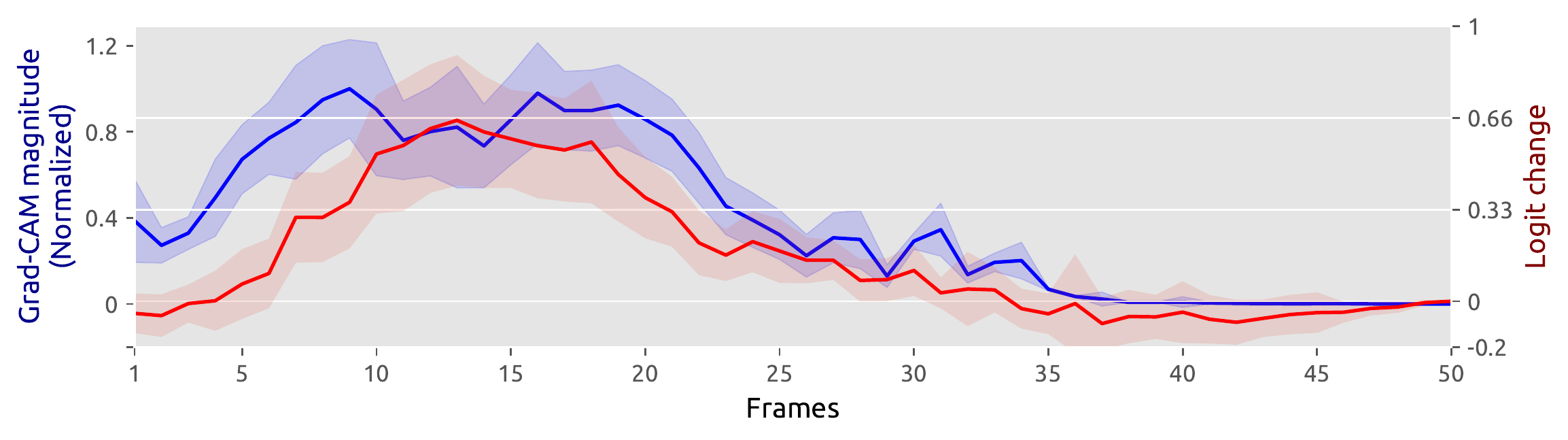}}\\\vspace{-0.5em}
\subfloat[Activity (1)]{\includegraphics[width=8.5cm]{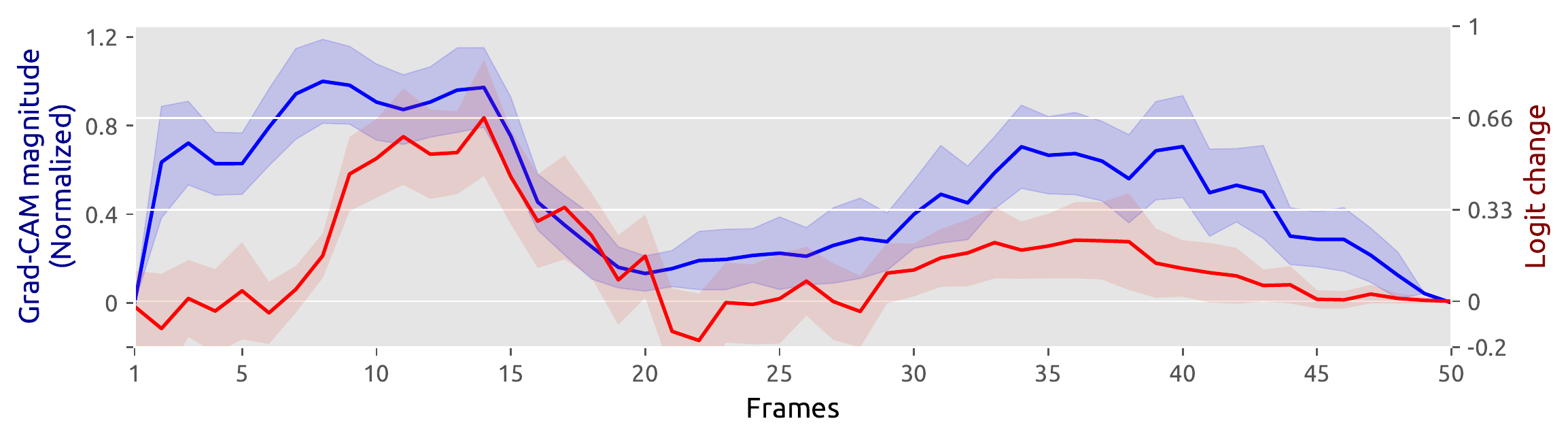}}\\\vspace{-0.5em}
\subfloat[Activity (2)]{\includegraphics[width=8.5cm]{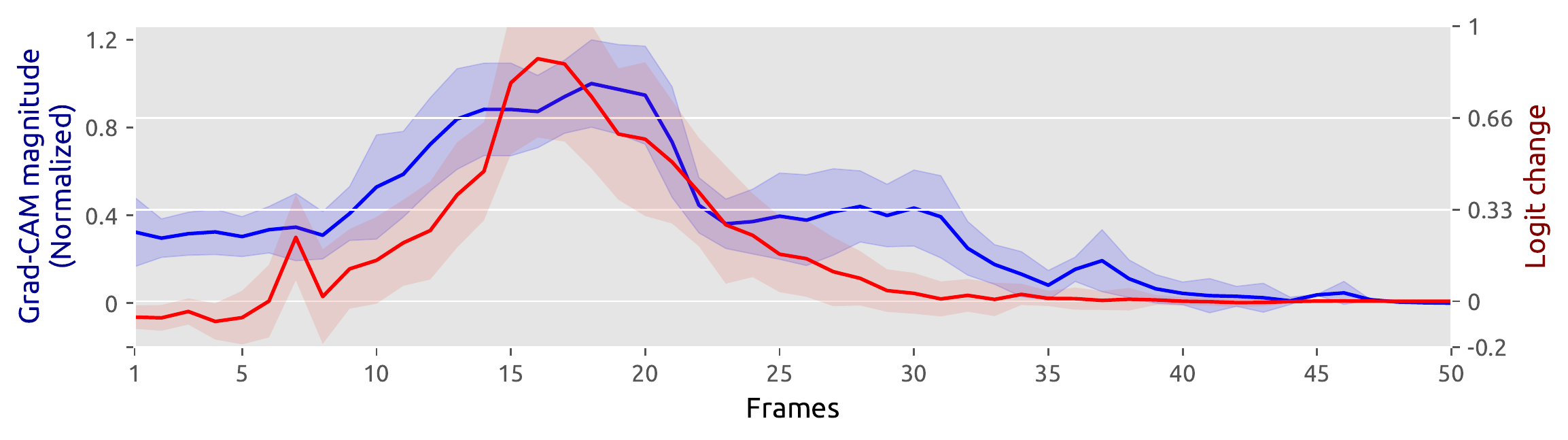}}\\\vspace{-0.5em}
\subfloat[Activity (3)]{\includegraphics[width=8.5cm]{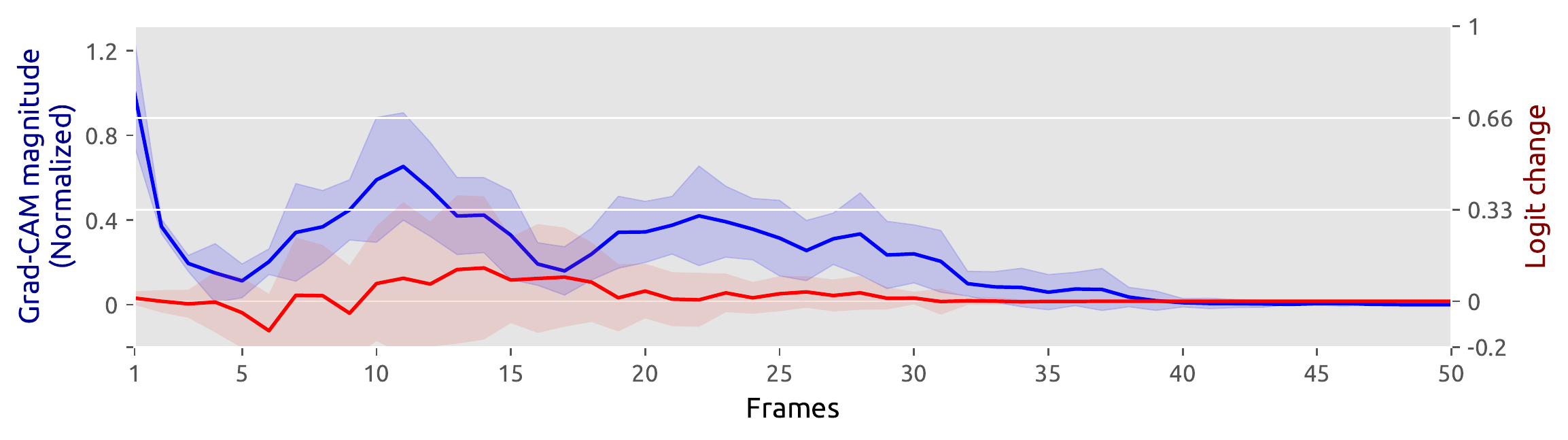}}\\\vspace{-0.5em}
\subfloat[Activity (4)]{\includegraphics[width=8.5cm]{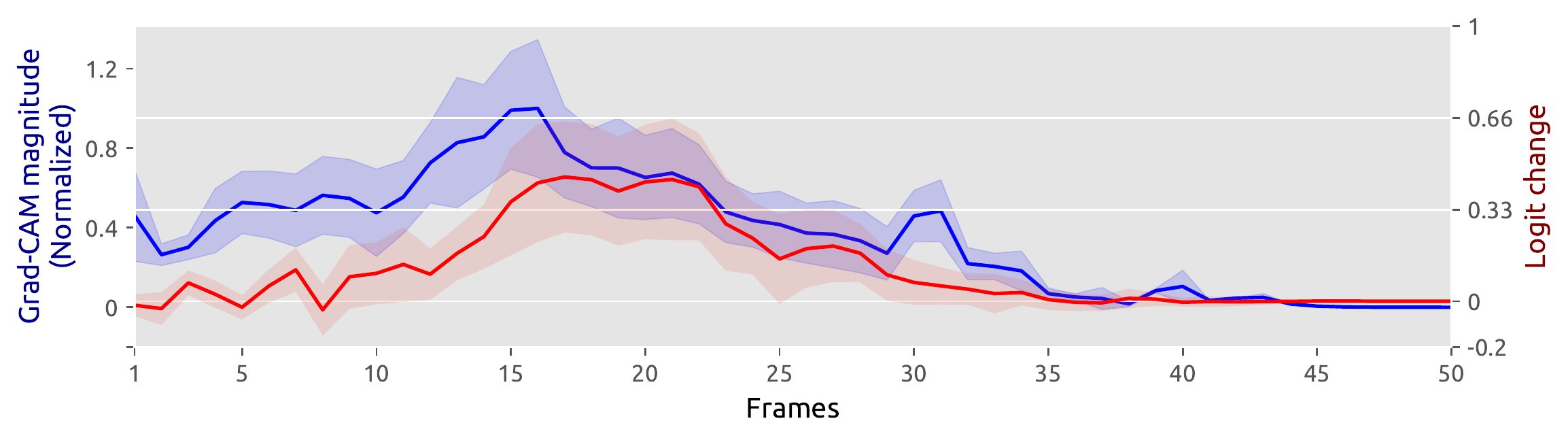}}\\\vspace{-0.5em}
\subfloat[Activity (5)]{\includegraphics[width=8.5cm]{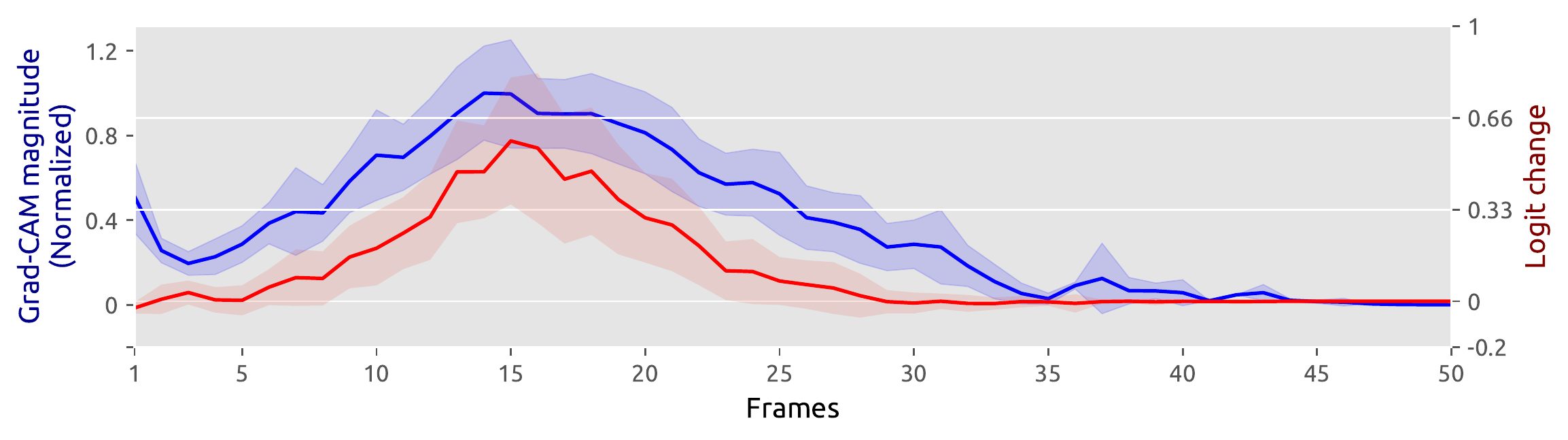}}
\caption{(blue) A line graph representation of Grad-CAM magnitudes (normalized between $0$ and $1$), displayed for individual frames that belong to genuine data points of adversarial examples that transfer from ${A}_{S_+}$ to ${A}_{S_-}$. This graph is plotted against the median frame importance (red), as calculated by the experiment detailed in Section~\ref{Relation of Adversarial Attacks to Interpretability} of the main text. The sub-figure labeled with (a) shows this graph for all of the adversarial examples in an aggregated manner, whereas the following sub-figures are class-specific. The line graph represents the median, whereas the shaded area represents the interquartile range.}
\label{fig:frame-gradcam-ext}
\vspace{-1em}
\end{figure}

Apart from the qualitative results provided in Fig.~\ref{fig:gracam_swipe_left}, which are heavily criticized in~\citet{lipton2016mythos} and~\citet{interpretation_fragile}, we now aim at performing a quantitative evaluation of the correctness of the produced Grad-CAM activation frames. We calculate the median magnitude of the produced Grad-CAM frames, which are normalized between $0$ and $1$, and compare it to the previously presented frame importance data in Fig.~\ref{fig:frame-gradcam-line}, where the blue line represents the median Grad-CAM magnitude and the red line the median frame importance. The bands around the lines correspond to the respective interquartile ranges. In the supplementary materials, the same type of illustration on a per-class basis is provided in Fig.~\ref{fig:frame-gradcam-ext}.

Given Fig.~\ref{fig:frame-gradcam-line}, we again observe a correlation between the importance of frames and their corresponding Grad-CAM activations. Both experiments, as presented in Fig.~\ref{fig:frame-perturb-boxplot} and Fig.~\ref{fig:frame-gradcam-line}, show strong correlation between their respective data. In particular, the higher the magnitude of the positive Grad-CAM heatmap, the larger the change in the prediction will be when replacing the underlying frame with the padding frame. Consequently, our experiments confirm that the output of Grad-CAM can indeed be used to assess the relative importance of each radar frame for the prediction made. Indeed, the frames that contribute the most to a prediction are also the ones that are naturally perturbed more than the others during an adversarial optimization, pointing to a strong connection between adversarial optimization and model interpretability.

\section{Conclusions and Future Work}
\label{Conclusion and Future Work}

In this study, we evaluated multiple scenarios in which adversarial attacks are performed on CNNs trained with a sequence of range-Doppler images obtained from a low-power FMCW radar sensor, with the goal of performing gesture recognition. Our analysis showed that these models are vulnerable not only to commonly used attacks, but also to unique attacks that take advantage of how the data set is crafted. In order to demonstrate a unique attack that leverages knowledge about the data set, we proposed a padding attack that creates a padding sequence that changes the predictions made by CNNs. 

An often mentioned drawback of CNNs is their lack of interpretability. By taking advantage of the data selected for this study, we were able to show the connection between the perturbation exercised by adversarial attacks and the importance of individual frames. Moreover, we were also able to demonstrate that it is possible to identify \textit{important} frames using Grad-CAM, thus showing (1) the relation between adversarial optimization and interpretability, and (2) a quantitative method to evaluate interpretability techniques.

In future research, we aim to analyze multiple shortcomings of radar sensors against so-called real-world adversarial examples~\citep{IFGS,sun2018survey}, as well as black-box attacks that do not use surrogate models~\citep{cheng2019improvinge_black_black_box,gragnaniello2019perceptual_black_black_box,tu2019autozoom_black_black_box}. In the case of activity detection, real-world adversarial examples may occur when radar sensors are employed in environments exhibiting poor recording conditions, such as environments that contain reflective materials (e.g., metal objects), or similarly, when the subject itself carries any reflective material. Moreover, it would be of interest to investigate the influence of multiple moving subjects in the same recording environment on adversariality.

\section*{Acknowledgements}
We would like to thank the anonymous reviewers for their valuable and insightful comments. We believe their comments significantly improved the quality of this manuscript.

The research activities described in this paper were funded by Ghent University Global Campus, Ghent University, imec, Flanders Innovation \& Entrepreneurship (VLAIO), the Fund for Scientific Research-Flanders (FWO-Flanders), and the EU.

\bibliographystyle{model2-names}
\bibliography{Preprint}

\begin{thebibliography}{57}
\expandafter\ifx\csname natexlab\endcsname\relax\def\natexlab#1{#1}\fi
\providecommand{\url}[1]{\texttt{#1}}
\providecommand{\href}[2]{#2}
\providecommand{\path}[1]{#1}
\providecommand{\DOIprefix}{doi:}
\providecommand{\ArXivprefix}{arXiv:}
\providecommand{\URLprefix}{URL: }
\providecommand{\Pubmedprefix}{pmid:}
\providecommand{\doi}[1]{\href{http://dx.doi.org/#1}{\path{#1}}}
\providecommand{\Pubmed}[1]{\href{pmid:#1}{\path{#1}}}
\providecommand{\bibinfo}[2]{#2}
\ifx\xfnm\relax \def\xfnm[#1]{\unskip,\space#1}\fi
\bibitem[{Athalye et~al.(2018)Athalye, Carlini and
  Wagner}]{athalye2018_obfuscated}
\bibinfo{author}{Athalye, A.}, \bibinfo{author}{Carlini, N.},
  \bibinfo{author}{Wagner, D.}, \bibinfo{year}{2018}.
\newblock \bibinfo{title}{{Obfuscated Gradients Give A False Sense Of Security:
  Circumventing Defenses To Adversarial Examples}}.
\newblock \bibinfo{journal}{CoRR} \bibinfo{volume}{abs/1802.00420}.
\bibitem[{Biggio et~al.(2013)Biggio, Corona, Maiorca, Nelson, {\v{S}}rndi{\'c},
  Laskov, Giacinto and Roli}]{biggio2013evasion}
\bibinfo{author}{Biggio, B.}, \bibinfo{author}{Corona, I.},
  \bibinfo{author}{Maiorca, D.}, \bibinfo{author}{Nelson, B.},
  \bibinfo{author}{{\v{S}}rndi{\'c}, N.}, \bibinfo{author}{Laskov, P.},
  \bibinfo{author}{Giacinto, G.}, \bibinfo{author}{Roli, F.},
  \bibinfo{year}{2013}.
\newblock \bibinfo{title}{{Evasion Attacks Against Machine Learning At Test
  Time}}, in: \bibinfo{booktitle}{Joint European conference on machine learning
  and knowledge discovery in databases}, \bibinfo{organization}{Springer}. pp.
  \bibinfo{pages}{387--402}.
\bibitem[{Brown et~al.(2017)Brown, Man{\'e}, Roy, Abadi and
  Gilmer}]{brown2017adversarial_patch}
\bibinfo{author}{Brown, T.B.}, \bibinfo{author}{Man{\'e}, D.},
  \bibinfo{author}{Roy, A.}, \bibinfo{author}{Abadi, M.},
  \bibinfo{author}{Gilmer, J.}, \bibinfo{year}{2017}.
\newblock \bibinfo{title}{{Adversarial Patch}}.
\newblock \bibinfo{journal}{CoRR} \bibinfo{volume}{abs/1712.09665}.
\bibitem[{Carlini and Wagner(2016)}]{CW_Attack}
\bibinfo{author}{Carlini, N.}, \bibinfo{author}{Wagner, D.A.},
  \bibinfo{year}{2016}.
\newblock \bibinfo{title}{{Towards Evaluating The Robustness of Neural
  Networks}}.
\newblock \bibinfo{journal}{CoRR} \bibinfo{volume}{abs/1608.04644}.
\bibitem[{Carlini and Wagner(2017)}]{DBLP:journalsCarliniW17}
\bibinfo{author}{Carlini, N.}, \bibinfo{author}{Wagner, D.A.},
  \bibinfo{year}{2017}.
\newblock \bibinfo{title}{{Adversarial Examples Are Not Easily Detected:
  Bypassing Ten Detection Methods}}.
\newblock \bibinfo{journal}{CoRR} \bibinfo{volume}{abs/1705.07263}.
\bibitem[{Chen et~al.(2014)Chen, Tahmoush and Miceli}]{2014Chen}
\bibinfo{author}{Chen, V.C.}, \bibinfo{author}{Tahmoush, D.},
  \bibinfo{author}{Miceli, W.J.}, \bibinfo{year}{2014}.
\newblock \bibinfo{title}{{Radar Micro-{{Doppler}} Signatures: Processing And
  Applications}}.
\newblock Radar, Sonar \& Navigation, \bibinfo{publisher}{Institution of
  Engineering and Technology}.
\bibitem[{Cheng et~al.(2019)Cheng, Dong, Pang, Su and
  Zhu}]{cheng2019improvinge_black_black_box}
\bibinfo{author}{Cheng, S.}, \bibinfo{author}{Dong, Y.}, \bibinfo{author}{Pang,
  T.}, \bibinfo{author}{Su, H.}, \bibinfo{author}{Zhu, J.},
  \bibinfo{year}{2019}.
\newblock \bibinfo{title}{{Improving Black-box Adversarial Attacks with a
  Transfer-based Prior}}, in: \bibinfo{booktitle}{Advances in Neural
  Information Processing Systems}, pp. \bibinfo{pages}{10932--10942}.
\bibitem[{Etmann et~al.(2019)Etmann, Lunz, Maass and
  Sch{\"o}nlieb}]{etmann2019connection}
\bibinfo{author}{Etmann, C.}, \bibinfo{author}{Lunz, S.},
  \bibinfo{author}{Maass, P.}, \bibinfo{author}{Sch{\"o}nlieb, C.B.},
  \bibinfo{year}{2019}.
\newblock \bibinfo{title}{{On The Connection Between Adversarial Robustness And
  Saliency Map Interpretability}}.
\newblock \bibinfo{journal}{arXiv preprint arXiv:1905.04172} .
\bibitem[{Ghorbani et~al.(2017)Ghorbani, Abid and Zou}]{interpretation_fragile}
\bibinfo{author}{Ghorbani, A.}, \bibinfo{author}{Abid, A.},
  \bibinfo{author}{Zou, J.}, \bibinfo{year}{2017}.
\newblock \bibinfo{title}{{Interpretation Of Neural Networks Is Fragile}}.
\newblock \bibinfo{journal}{CoRR} \bibinfo{volume}{abs/1710.10547}.
\bibitem[{Goodfellow et~al.(2018)Goodfellow, McDaniel and
  Papernot}]{Goodfellow:2018:MML:3234519.3134599}
\bibinfo{author}{Goodfellow, I.}, \bibinfo{author}{McDaniel, P.},
  \bibinfo{author}{Papernot, N.}, \bibinfo{year}{2018}.
\newblock \bibinfo{title}{{Making Machine Learning Robust Against Adversarial
  Inputs}}.
\newblock \bibinfo{journal}{Communications of the ACM} \bibinfo{volume}{61},
  \bibinfo{pages}{56--66}.
\bibitem[{Goodfellow et~al.(2014)Goodfellow, Shlens and
  Szegedy}]{Goodfellow-expharnessing}
\bibinfo{author}{Goodfellow, I.}, \bibinfo{author}{Shlens, J.},
  \bibinfo{author}{Szegedy, C.}, \bibinfo{year}{2014}.
\newblock \bibinfo{title}{{Explaining and Harnessing Adversarial Examples}}.
\newblock \bibinfo{journal}{CoRR} \bibinfo{volume}{abs/1412.6572}.
\bibitem[{Gragnaniello et~al.(2019)Gragnaniello, Marra, Poggi and
  Verdoliva}]{gragnaniello2019perceptual_black_black_box}
\bibinfo{author}{Gragnaniello, D.}, \bibinfo{author}{Marra, F.},
  \bibinfo{author}{Poggi, G.}, \bibinfo{author}{Verdoliva, L.},
  \bibinfo{year}{2019}.
\newblock \bibinfo{title}{Perceptual quality-preserving black-box attack
  against deep learning image classifiers}.
\newblock \bibinfo{journal}{CoRR} \bibinfo{volume}{abs/1902.07776}.
\bibitem[{Graves et~al.(2013)Graves, Mohamed and Hinton}]{graves2013speech}
\bibinfo{author}{Graves, A.}, \bibinfo{author}{Mohamed, A.r.},
  \bibinfo{author}{Hinton, G.}, \bibinfo{year}{2013}.
\newblock \bibinfo{title}{{Speech Recognition With Deep Recurrent Neural
  Networks}}, in: \bibinfo{booktitle}{2013 IEEE international conference on
  acoustics, speech and signal processing}, \bibinfo{organization}{IEEE}. pp.
  \bibinfo{pages}{6645--6649}.
\bibitem[{Guo et~al.(2019)Guo, Gardner, You, Wilson and
  Weinberger}]{guo2019simple_black_black_box}
\bibinfo{author}{Guo, C.}, \bibinfo{author}{Gardner, J.R.},
  \bibinfo{author}{You, Y.}, \bibinfo{author}{Wilson, A.G.},
  \bibinfo{author}{Weinberger, K.Q.}, \bibinfo{year}{2019}.
\newblock \bibinfo{title}{{Simple Black-box Adversarial Attacks}}.
\newblock \bibinfo{journal}{CoRR} \bibinfo{volume}{abs/1905.07121}.
\bibitem[{Hara et~al.(2018)Hara, Kataoka and Satoh}]{resnext_video}
\bibinfo{author}{Hara, K.}, \bibinfo{author}{Kataoka, H.},
  \bibinfo{author}{Satoh, Y.}, \bibinfo{year}{2018}.
\newblock \bibinfo{title}{{Can Spatiotemporal 3D CNNs Retrace the History of 2D
  CNNs and ImageNet?}}, in: \bibinfo{booktitle}{Proceedings of the IEEE
  conference on Computer Vision and Pattern Recognition}, pp.
  \bibinfo{pages}{6546--6555}.
\bibitem[{He et~al.(2016)He, Zhang, Ren and Sun}]{resnet}
\bibinfo{author}{He, K.}, \bibinfo{author}{Zhang, X.}, \bibinfo{author}{Ren,
  S.}, \bibinfo{author}{Sun, J.}, \bibinfo{year}{2016}.
\newblock \bibinfo{title}{{Deep Residual Learning For Image Recognition}}, in:
  \bibinfo{booktitle}{Proceedings of the IEEE conference on computer vision and
  pattern recognition}, pp. \bibinfo{pages}{770--778}.
\bibitem[{Hochreiter and Schmidhuber(1997)}]{LSTM}
\bibinfo{author}{Hochreiter, S.}, \bibinfo{author}{Schmidhuber, J.},
  \bibinfo{year}{1997}.
\newblock \bibinfo{title}{{Long short-term memory}}.
\newblock \bibinfo{journal}{Neural computation} \bibinfo{volume}{9},
  \bibinfo{pages}{1735--1780}.
\bibitem[{Ilyas et~al.(2018)Ilyas, Engstrom, Athalye and
  Lin}]{ilyas2018black_black_black_box}
\bibinfo{author}{Ilyas, A.}, \bibinfo{author}{Engstrom, L.},
  \bibinfo{author}{Athalye, A.}, \bibinfo{author}{Lin, J.},
  \bibinfo{year}{2018}.
\newblock \bibinfo{title}{{Black-box Adversarial Attacks with Limited Queries
  and Information}}.
\newblock \bibinfo{journal}{CoRR} \bibinfo{volume}{abs/1804.08598}.
\bibitem[{Jalalvand et~al.(2019)Jalalvand, Vandersmissen, De~Neve and
  Mannens}]{Reservoir_Radar2019}
\bibinfo{author}{Jalalvand, A.}, \bibinfo{author}{Vandersmissen, B.},
  \bibinfo{author}{De~Neve, W.}, \bibinfo{author}{Mannens, E.},
  \bibinfo{year}{2019}.
\newblock \bibinfo{title}{{Radar Signal Processing For Human Identification By
  Means Of Reservoir Computing Networks}}, in: \bibinfo{booktitle}{IEEE Radar
  Conference}, pp. \bibinfo{pages}{1--6}.
\bibitem[{Jokanovic et~al.(2016)Jokanovic, Amin and
  Ahmad}]{jokanovic2016radar_cite_data_is_small}
\bibinfo{author}{Jokanovic, B.}, \bibinfo{author}{Amin, M.},
  \bibinfo{author}{Ahmad, F.}, \bibinfo{year}{2016}.
\newblock \bibinfo{title}{{Radar Fall Motion Detection Using Deep Learning}},
  in: \bibinfo{booktitle}{2016 IEEE radar conference (RadarConf)},
  \bibinfo{organization}{IEEE}. pp. \bibinfo{pages}{1--6}.
\bibitem[{Kim and Moon(2015)}]{kim2015humanr_cite_data_is_small}
\bibinfo{author}{Kim, Y.}, \bibinfo{author}{Moon, T.}, \bibinfo{year}{2015}.
\newblock \bibinfo{title}{{Human Detection and Activity Classification Based on
  Micro-Doppler Signatures Using Deep Convolutional Neural Networks}}.
\newblock \bibinfo{journal}{IEEE geoscience and remote sensing letters}
  \bibinfo{volume}{13}, \bibinfo{pages}{8--12}.
\bibitem[{Kim and Toomajian(2016)}]{md_hand_gesture}
\bibinfo{author}{Kim, Y.}, \bibinfo{author}{Toomajian, B.},
  \bibinfo{year}{2016}.
\newblock \bibinfo{title}{{Hand Gesture Recognition Using Micro-Doppler
  Signatures With Convolutional Neural Network}}.
\newblock \bibinfo{journal}{IEEE Access} \bibinfo{volume}{4},
  \bibinfo{pages}{7125--7130}.
\bibitem[{Kingma and Ba(2014)}]{adaptive_momentum}
\bibinfo{author}{Kingma, D.P.}, \bibinfo{author}{Ba, J.}, \bibinfo{year}{2014}.
\newblock \bibinfo{title}{{Adam: {{A}} Method For Stochastic Optimization}}.
\newblock \bibinfo{journal}{CoRR} \bibinfo{volume}{abs/1412.6980}.
\bibitem[{Krizhevsky et~al.(2012)Krizhevsky, Sutskever and Hinton}]{Alexnet}
\bibinfo{author}{Krizhevsky, A.}, \bibinfo{author}{Sutskever, I.},
  \bibinfo{author}{Hinton, G.E.}, \bibinfo{year}{2012}.
\newblock \bibinfo{title}{{ImageNet classification with deep convolutional
  neural networks}}, in: \bibinfo{booktitle}{Advances in neural information
  processing systems}, pp. \bibinfo{pages}{1097--1105}.
\bibitem[{Kurakin et~al.(2016)Kurakin, Goodfellow and Bengio}]{IFGS}
\bibinfo{author}{Kurakin, A.}, \bibinfo{author}{Goodfellow, I.},
  \bibinfo{author}{Bengio, S.}, \bibinfo{year}{2016}.
\newblock \bibinfo{title}{{Adversarial Examples In The Physical World}}.
\newblock \bibinfo{journal}{CoRR} \bibinfo{volume}{abs/1607.02533}.
\bibitem[{LeCun et~al.(1998)LeCun, Bottou, Bengio and
  Haffner}]{lecun1998gradient}
\bibinfo{author}{LeCun, Y.}, \bibinfo{author}{Bottou, L.},
  \bibinfo{author}{Bengio, Y.}, \bibinfo{author}{Haffner, P.},
  \bibinfo{year}{1998}.
\newblock \bibinfo{title}{{Gradient-Based Learning Applied To Document
  Recognition}}.
\newblock \bibinfo{journal}{Proceedings of the IEEE} \bibinfo{volume}{86},
  \bibinfo{pages}{2278--2324}.
\bibitem[{Lipton(2016)}]{lipton2016mythos}
\bibinfo{author}{Lipton, Z.C.}, \bibinfo{year}{2016}.
\newblock \bibinfo{title}{{The Mythos Of Model Interpretability}}.
\newblock \bibinfo{journal}{arXiv preprint arXiv:1606.03490} .
\bibitem[{Madry et~al.(2017)Madry, Makelov, Schmidt, Tsipras and
  Vladu}]{PGD_attack}
\bibinfo{author}{Madry, A.}, \bibinfo{author}{Makelov, A.},
  \bibinfo{author}{Schmidt, L.}, \bibinfo{author}{Tsipras, D.},
  \bibinfo{author}{Vladu, A.}, \bibinfo{year}{2017}.
\newblock \bibinfo{title}{{Towards Deep Learning Models Resistant To
  Adversarial Attacks}}.
\newblock \bibinfo{journal}{CoRR} \bibinfo{volume}{abs/1706.06083}.
\bibitem[{Moosavi-Dezfooli et~al.(2017)Moosavi-Dezfooli, Fawzi, Fawzi and
  Frossard}]{moosavi2017universal}
\bibinfo{author}{Moosavi-Dezfooli, S.M.}, \bibinfo{author}{Fawzi, A.},
  \bibinfo{author}{Fawzi, O.}, \bibinfo{author}{Frossard, P.},
  \bibinfo{year}{2017}.
\newblock \bibinfo{title}{{Universal Adversarial Perturbations}}, in:
  \bibinfo{booktitle}{Proceedings of the IEEE conference on computer vision and
  pattern recognition}, pp. \bibinfo{pages}{1765--1773}.
\bibitem[{Nair et~al.(2018)Nair, Thomas and Jayagopi}]{residual_lstm_2}
\bibinfo{author}{Nair, N.}, \bibinfo{author}{Thomas, C.},
  \bibinfo{author}{Jayagopi, D.B.}, \bibinfo{year}{2018}.
\newblock \bibinfo{title}{{Human Activity Recognition Using Temporal
  Convolutional Network}}, in: \bibinfo{booktitle}{Proceedings of the 5th
  international Workshop on Sensor-based Activity Recognition and Interaction},
  pp. \bibinfo{pages}{1--8}.
\bibitem[{Ozbulak et~al.(2020)Ozbulak, Gasparyan, De~Neve and
  Van~Messem}]{ozbulak2020perturbation}
\bibinfo{author}{Ozbulak, U.}, \bibinfo{author}{Gasparyan, M.},
  \bibinfo{author}{De~Neve, W.}, \bibinfo{author}{Van~Messem, A.},
  \bibinfo{year}{2020}.
\newblock \bibinfo{title}{Perturbation analysis of gradient-based adversarial
  attacks}.
\newblock \bibinfo{journal}{Pattern Recognition Letters} .
\bibitem[{Papernot et~al.(2015)Papernot, McDaniel, Jha, Fredrikson, Celik and
  Swami}]{JSMA}
\bibinfo{author}{Papernot, N.}, \bibinfo{author}{McDaniel, P.D.},
  \bibinfo{author}{Jha, S.}, \bibinfo{author}{Fredrikson, M.},
  \bibinfo{author}{Celik, Z.B.}, \bibinfo{author}{Swami, A.},
  \bibinfo{year}{2015}.
\newblock \bibinfo{title}{{The Limitations Of Deep Learning In Adversarial
  Settings}}.
\newblock \bibinfo{journal}{CoRR} \bibinfo{volume}{abs/1511.07528}.
\bibitem[{Rajpoot and Jensen(2015)}]{rajpoot2015video}
\bibinfo{author}{Rajpoot, Q.M.}, \bibinfo{author}{Jensen, C.D.},
  \bibinfo{year}{2015}.
\newblock \bibinfo{title}{{Video Surveillance: Privacy Issues And Legal
  Compliance}}, in: \bibinfo{booktitle}{Promoting Social Change and Democracy
  through Information Technology}. \bibinfo{publisher}{IGI global}, pp.
  \bibinfo{pages}{69--92}.
\bibitem[{Ross and Doshi-Velez(2018)}]{ross2018improving}
\bibinfo{author}{Ross, A.S.}, \bibinfo{author}{Doshi-Velez, F.},
  \bibinfo{year}{2018}.
\newblock \bibinfo{title}{{Improving The Adversarial Robustness And
  Interpretability Of Deep Neural Networks By Regularizing Their Input
  Gradients}}, in: \bibinfo{booktitle}{Thirty-second AAAI conference on
  artificial intelligence}.
\bibitem[{Russakovsky et~al.(2015)Russakovsky, Deng, Su, Krause, Satheesh, Ma,
  Huang, Karpathy, Khosla, Bernstein, Berg and Fei-Fei}]{ILSVRC15:rus}
\bibinfo{author}{Russakovsky, O.}, \bibinfo{author}{Deng, J.},
  \bibinfo{author}{Su, H.}, \bibinfo{author}{Krause, J.},
  \bibinfo{author}{Satheesh, S.}, \bibinfo{author}{Ma, S.},
  \bibinfo{author}{Huang, Z.}, \bibinfo{author}{Karpathy, A.},
  \bibinfo{author}{Khosla, A.}, \bibinfo{author}{Bernstein, M.},
  \bibinfo{author}{Berg, A.C.}, \bibinfo{author}{Fei-Fei, L.},
  \bibinfo{year}{2015}.
\newblock \bibinfo{title}{{ImageNet Large Scale Visual Recognition Challenge}}.
\newblock \bibinfo{journal}{International Journal of Computer Vision}
  \bibinfo{volume}{115}, \bibinfo{pages}{211--252}.
\bibitem[{Selvaraju et~al.(2016)Selvaraju, Das, Vedantam, Cogswell, Parikh and
  Batra}]{grad_cam}
\bibinfo{author}{Selvaraju, R.R.}, \bibinfo{author}{Das, A.},
  \bibinfo{author}{Vedantam, R.}, \bibinfo{author}{Cogswell, M.},
  \bibinfo{author}{Parikh, D.}, \bibinfo{author}{Batra, D.},
  \bibinfo{year}{2016}.
\newblock \bibinfo{title}{{Grad-Cam: Why Did You Say That? Visual Explanations
  From Deep Networks Via Gradient-Based Localization}}.
\newblock \bibinfo{journal}{CVPR 2016} .
\bibitem[{Seyfio{\u{g}}lu et~al.(2018)Seyfio{\u{g}}lu, {\"O}zbayo{\u{g}}lu and
  G{\"u}rb{\"u}z}]{seyfiouglu2018deep_cite_data_is_small}
\bibinfo{author}{Seyfio{\u{g}}lu, M.S.}, \bibinfo{author}{{\"O}zbayo{\u{g}}lu,
  A.M.}, \bibinfo{author}{G{\"u}rb{\"u}z, S.Z.}, \bibinfo{year}{2018}.
\newblock \bibinfo{title}{{Deep Convolutional Autoencoder for Radar-Based
  Classification of Similar Aided and Unaided Human Activities}}.
\newblock \bibinfo{journal}{IEEE Transactions on Aerospace and Electronic
  Systems} \bibinfo{volume}{54}, \bibinfo{pages}{1709--1723}.
\bibitem[{Shrikumar et~al.(2017)Shrikumar, Greenside and Kundaje}]{DeepLift}
\bibinfo{author}{Shrikumar, A.}, \bibinfo{author}{Greenside, P.},
  \bibinfo{author}{Kundaje, A.}, \bibinfo{year}{2017}.
\newblock \bibinfo{title}{{Learning Important Features Through Propagating
  Activation Differences}}.
\newblock \bibinfo{journal}{CoRR} \bibinfo{volume}{abs/1704.02685}.
\bibitem[{Simonyan et~al.(2014)Simonyan, Vedaldi and Zisserman}]{simonyan}
\bibinfo{author}{Simonyan, K.}, \bibinfo{author}{Vedaldi, A.},
  \bibinfo{author}{Zisserman, A.}, \bibinfo{year}{2014}.
\newblock \bibinfo{title}{{Deep Inside Convolutional Networks: Visualising
  Image Classification Models And Saliency Maps}}, in:
  \bibinfo{booktitle}{Workshop, Proceedings of 2th International Conference on
  Learning Representations (ICLR)}.
\bibitem[{Simonyan and Zisserman(2014)}]{VGG}
\bibinfo{author}{Simonyan, K.}, \bibinfo{author}{Zisserman, A.},
  \bibinfo{year}{2014}.
\newblock \bibinfo{title}{{Very Deep Convolutional Networks For Large-Scale
  Image Recognition}}.
\newblock \bibinfo{journal}{CoRR} \bibinfo{volume}{abs/1409.1556}.
\bibitem[{Staples(2019)}]{media-privacy}
\bibinfo{author}{Staples, P.}, \bibinfo{year}{2019}.
\newblock \bibinfo{title}{{Thinking About Buying A Smart Home Device? Heres
  What You Need To Know About Security}}.
\newblock \bibinfo{howpublished}{\url{https://www.forbes.com}}.
\newblock \bibinfo{note}{Accessed: 2019-07-26}.
\bibitem[{Sun et~al.(2018)Sun, Tan and Zhou}]{sun2018survey}
\bibinfo{author}{Sun, L.}, \bibinfo{author}{Tan, M.}, \bibinfo{author}{Zhou,
  Z.}, \bibinfo{year}{2018}.
\newblock \bibinfo{title}{{A Survey of Practical Adversarial Example Attacks}}.
\newblock \bibinfo{journal}{Cybersecurity} \bibinfo{volume}{1},
  \bibinfo{pages}{9}.
\bibitem[{Szegedy et~al.(2016)Szegedy, Vanhoucke, Ioffe, Shlens and
  Wojna}]{inceptionv3}
\bibinfo{author}{Szegedy, C.}, \bibinfo{author}{Vanhoucke, V.},
  \bibinfo{author}{Ioffe, S.}, \bibinfo{author}{Shlens, J.},
  \bibinfo{author}{Wojna, Z.}, \bibinfo{year}{2016}.
\newblock \bibinfo{title}{{Rethinking The Inception Architecture For Computer
  Vision}}, in: \bibinfo{booktitle}{Proceedings of the IEEE conference on
  computer vision and pattern recognition}, pp. \bibinfo{pages}{2818--2826}.
\bibitem[{Szegedy et~al.(2013)Szegedy, Zaremba, Sutskever, Bruna, Erhan,
  Goodfellow and Fergus}]{LBFGS}
\bibinfo{author}{Szegedy, C.}, \bibinfo{author}{Zaremba, W.},
  \bibinfo{author}{Sutskever, I.}, \bibinfo{author}{Bruna, J.},
  \bibinfo{author}{Erhan, D.}, \bibinfo{author}{Goodfellow, I.},
  \bibinfo{author}{Fergus, R.}, \bibinfo{year}{2013}.
\newblock \bibinfo{title}{{Intriguing Properties Of Neural Networks}}.
\newblock \bibinfo{journal}{CoRR} \bibinfo{volume}{abs/1312.6199}.
\bibitem[{Tao et~al.(2018)Tao, Ma, Liu and Zhang}]{tao2018attacks}
\bibinfo{author}{Tao, G.}, \bibinfo{author}{Ma, S.}, \bibinfo{author}{Liu, Y.},
  \bibinfo{author}{Zhang, X.}, \bibinfo{year}{2018}.
\newblock \bibinfo{title}{{Attacks Meet Interpretability: Attribute-Steered
  Detection Of Adversarial Samples}}, in: \bibinfo{booktitle}{Advances in
  Neural Information Processing Systems}, pp. \bibinfo{pages}{7717--7728}.
\bibitem[{Tu et~al.(2019)Tu, Ting, Chen, Liu, Zhang, Yi, Hsieh and
  Cheng}]{tu2019autozoom_black_black_box}
\bibinfo{author}{Tu, C.C.}, \bibinfo{author}{Ting, P.}, \bibinfo{author}{Chen,
  P.Y.}, \bibinfo{author}{Liu, S.}, \bibinfo{author}{Zhang, H.},
  \bibinfo{author}{Yi, J.}, \bibinfo{author}{Hsieh, C.J.},
  \bibinfo{author}{Cheng, S.M.}, \bibinfo{year}{2019}.
\newblock \bibinfo{title}{Autozoom: Autoencoder-based zeroth order optimization
  method for attacking black-box neural networks}, in:
  \bibinfo{booktitle}{Proceedings of the AAAI Conference on Artificial
  Intelligence}, pp. \bibinfo{pages}{742--749}.
\bibitem[{Vandersmissen et~al.(2018)Vandersmissen, Knudde, Jalalvand, Couckuyt,
  Bourdoux, De~Neve and Dhaene}]{vandersmissen2018indoor}
\bibinfo{author}{Vandersmissen, B.}, \bibinfo{author}{Knudde, N.},
  \bibinfo{author}{Jalalvand, A.}, \bibinfo{author}{Couckuyt, I.},
  \bibinfo{author}{Bourdoux, A.}, \bibinfo{author}{De~Neve, W.},
  \bibinfo{author}{Dhaene, T.}, \bibinfo{year}{2018}.
\newblock \bibinfo{title}{{Indoor Person Identification Using A Low-Power Fmcw
  Radar}}.
\newblock \bibinfo{journal}{IEEE Transactions on Geoscience and Remote Sensing}
  \bibinfo{volume}{56}, \bibinfo{pages}{3941--3952}.
\bibitem[{Vandersmissen et~al.(2019)Vandersmissen, Knudde, Jalalvand, Couckuyt,
  Dhaene and De~Neve}]{vandersmissen2019indoor}
\bibinfo{author}{Vandersmissen, B.}, \bibinfo{author}{Knudde, N.},
  \bibinfo{author}{Jalalvand, A.}, \bibinfo{author}{Couckuyt, I.},
  \bibinfo{author}{Dhaene, T.}, \bibinfo{author}{De~Neve, W.},
  \bibinfo{year}{2019}.
\newblock \bibinfo{title}{{Indoor Human Activity Recognition Using
  High-Dimensional Sensors And Deep Neural Networks}}.
\newblock \bibinfo{journal}{Neural Computing and Applications} ,
  \bibinfo{pages}{1--15}.
\bibitem[{Wang et~al.(2016)Wang, Song, Lien, Poupyrev and
  Hilliges}]{gesture_rec_radar}
\bibinfo{author}{Wang, S.}, \bibinfo{author}{Song, J.}, \bibinfo{author}{Lien,
  J.}, \bibinfo{author}{Poupyrev, I.}, \bibinfo{author}{Hilliges, O.},
  \bibinfo{year}{2016}.
\newblock \bibinfo{title}{{Interacting With Soli: Exploring Fine-Grained
  Dynamic Gesture Recognition In The Radio-Frequency Spectrum}}, in:
  \bibinfo{booktitle}{Proceedings of the 29th Annual Symposium on User
  Interface Software and Technology}, \bibinfo{organization}{ACM}. pp.
  \bibinfo{pages}{851--860}.
\bibitem[{Wu et~al.(2020)Wu, Wang, Xia, Bailey and
  Ma}]{skip_connections_easier}
\bibinfo{author}{Wu, D.}, \bibinfo{author}{Wang, Y.}, \bibinfo{author}{Xia,
  S.T.}, \bibinfo{author}{Bailey, J.}, \bibinfo{author}{Ma, X.},
  \bibinfo{year}{2020}.
\newblock \bibinfo{title}{{Skip Connections Matter: On the Transferability of
  Adversarial Examples Generated with ResNets}}, in:
  \bibinfo{booktitle}{International Conference on Learning Representations}.
\bibitem[{Xie et~al.(2017)Xie, Girshick, Doll{\'a}r, Tu and He}]{resnext}
\bibinfo{author}{Xie, S.}, \bibinfo{author}{Girshick, R.},
  \bibinfo{author}{Doll{\'a}r, P.}, \bibinfo{author}{Tu, Z.},
  \bibinfo{author}{He, K.}, \bibinfo{year}{2017}.
\newblock \bibinfo{title}{{Aggregated Residual Transformations for Deep Neural
  Networks}}, in: \bibinfo{booktitle}{Proceedings of the IEEE conference on
  computer vision and pattern recognition}, pp. \bibinfo{pages}{1492--1500}.
\bibitem[{Yao and Qian(2018)}]{residual_lstm_3}
\bibinfo{author}{Yao, L.}, \bibinfo{author}{Qian, Y.}, \bibinfo{year}{2018}.
\newblock \bibinfo{title}{{DT-3DResNet-LSTM: An Architecture for Temporal
  Activity Recognition in Videos}}, in: \bibinfo{booktitle}{Pacific Rim
  Conference on Multimedia}, \bibinfo{organization}{Springer}. pp.
  \bibinfo{pages}{622--632}.
\bibitem[{Zeiler and Fergus(2014)}]{ZeilerFergus_VisUnderstand}
\bibinfo{author}{Zeiler, M.D.}, \bibinfo{author}{Fergus, R.},
  \bibinfo{year}{2014}.
\newblock \bibinfo{title}{{Visualizing And Understanding Convolutional
  Networks}}, in: \bibinfo{booktitle}{European conference on computer vision},
  \bibinfo{organization}{Springer}. pp. \bibinfo{pages}{818--833}.
\bibitem[{Zhang et~al.(2019)Zhang, Zhang, Zhong, Lei, Yang, Du and
  Chen}]{video_camera_is_strong}
\bibinfo{author}{Zhang, H.B.}, \bibinfo{author}{Zhang, Y.X.},
  \bibinfo{author}{Zhong, B.}, \bibinfo{author}{Lei, Q.},
  \bibinfo{author}{Yang, L.}, \bibinfo{author}{Du, J.X.},
  \bibinfo{author}{Chen, D.S.}, \bibinfo{year}{2019}.
\newblock \bibinfo{title}{{A Comprehensive Survey Of Vision-Based Human Action
  Recognition Methods}}.
\newblock \bibinfo{journal}{Sensors} \bibinfo{volume}{19},
  \bibinfo{pages}{1005}.
\bibitem[{Zhao et~al.(2018a)Zhao, Li, Abu~Alsheikh, Tian, Zhao, Torralba and
  Katabi}]{through_wall_sensing}
\bibinfo{author}{Zhao, M.}, \bibinfo{author}{Li, T.},
  \bibinfo{author}{Abu~Alsheikh, M.}, \bibinfo{author}{Tian, Y.},
  \bibinfo{author}{Zhao, H.}, \bibinfo{author}{Torralba, A.},
  \bibinfo{author}{Katabi, D.}, \bibinfo{year}{2018}a.
\newblock \bibinfo{title}{{Through-Wall Human Pose Estimation Using Radio
  Signals}}, in: \bibinfo{booktitle}{Proceedings of the IEEE Conference on
  Computer Vision and Pattern Recognition}, pp. \bibinfo{pages}{7356--7365}.
\bibitem[{Zhao et~al.(2018b)Zhao, Yang, Chevalier, Xu and
  Zhang}]{residual_lstm_1}
\bibinfo{author}{Zhao, Y.}, \bibinfo{author}{Yang, R.},
  \bibinfo{author}{Chevalier, G.}, \bibinfo{author}{Xu, X.},
  \bibinfo{author}{Zhang, Z.}, \bibinfo{year}{2018}b.
\newblock \bibinfo{title}{{Deep Residual Bidir-LSTM for Human Activity
  Recognition Using Wearable Sensors}}.
\newblock \bibinfo{journal}{Mathematical Problems in Engineering}
  \bibinfo{volume}{2018}.
\bibitem[{Zhou et~al.(2016)Zhou, Khosla, Lapedriza, Oliva and
  Torralba}]{class_activation_map}
\bibinfo{author}{Zhou, B.}, \bibinfo{author}{Khosla, A.},
  \bibinfo{author}{Lapedriza, A.}, \bibinfo{author}{Oliva, A.},
  \bibinfo{author}{Torralba, A.}, \bibinfo{year}{2016}.
\newblock \bibinfo{title}{{Learning Deep Features For Discriminative
  Localization}}, in: \bibinfo{booktitle}{Proceedings of the IEEE conference on
  computer vision and pattern recognition}, pp. \bibinfo{pages}{2921--2929}.

\end{thebibliography}

\clearpage

\onecolumn
\begin{center}
    \LARGE 
Supplementary materials for:
\\
Investigating the significance of adversarial attacks and their relation to interpretability for radar-based human activity recognition systems
\end{center}

\vspace{2em}

\section*{Detailed Architectural Descriptions}
\begin{figure}[h]
\centering
\begin{tikzpicture}
\foreach \y in {-1,0,1,2,3,4,5,6,7,8,9,10,11,12,13,14}
{
\draw (0.5, \y-\y*0.2) -- (0.5, \y-\y*0.2 - 0.3 );
}
\foreach \y in {13, 10, 7, 4, 1,-1}
{
\draw[draw=black] (-1,\y-\y*0.2) -- (-1,\y-\y*0.2+0.5) -- (2,\y-\y*0.2+0.5) -- (2,\y-\y*0.2) -- (-1,\y-\y*0.2);
}
\foreach \y in {14,12,11,9,8,6,5,3,2,0,-2}
{
\draw[draw=gray] (-0.75,\y-\y*0.2) -- (-0.75,\y-\y*0.2+0.5) -- (1.75,\y-\y*0.2+0.5) -- (1.75,\y-\y*0.2) -- (-0.75,\y-\y*0.2);
}
\foreach \y in {13, 10, 7, 4}
{
\node[align=left] at (0.5, \y-\y*0.2+0.225) {\scriptsize $(3\times3\times3)$ Conv};
}
\foreach \y in {1, -1}
{
\node[align=left] at (0.5, \y-\y*0.2+0.225) {\scriptsize Linear};
}
\foreach \y in {12, 9, 6, 3,0}
{
\node[align=left] at (0.5, \y-\y*0.2+0.225) {\scriptsize ELU+Dropout};
}
\foreach \y in {11, 8, 5,2}
{
\node[align=left] at (0.5, \y-\y*0.2+0.225) {\scriptsize Pooling};
}

\node[align=left] at (0.5, 14-14*0.2+0.225) {\scriptsize Input};
\node[align=left] at (0.5, -2- -2*0.2+0.225) {\scriptsize Output};
\node[align=left] at (3.25, 11.45) {\scriptsize $(1\times50\times80\times126)$};
\node[align=left] at (3.25, 10.65) {\scriptsize $(8\times50\times80\times126)$};
\node[align=left] at (3.25, 9.05) {\scriptsize $(8\times16\times40\times63)$};
\node[align=left] at (3.25, 8.25) {\scriptsize $(16\times16\times40\times63)$};
\node[align=left] at (3.25, 6.65) {\scriptsize $(16\times8\times20\times31)$};
\node[align=left] at (3.25, 5.85) {\scriptsize $(32\times8\times20\times31)$};
\node[align=left] at (3.25, 4.25) {\scriptsize $(32\times4\times10\times15)$};
\node[align=left] at (3.25, 3.45) {\scriptsize $(64\times4\times10\times15)$};
\node[align=left] at (3.25, 1.85) {\scriptsize $(64\times2\times10\times15)$};
\node[align=left] at (3.25, 1.05) {\scriptsize $(128\times 1)$};
\node[align=left] at (3.25, -0.55) {\scriptsize $(6 \times 1)$};

\node[align=center] at (0.5, -3.8- -3*0.2+0.225) {Architecture: $\mathcal{A}$\\ Trainable parameters:$\sim6.4 \times 10^{5}$\\ Size: $\sim2.5$MB };
\end{tikzpicture}
\caption{Detailed description for architecture $\mathcal{A}$.}
\label{fig:arch1}
\end{figure}

\begin{figure}[h]
\centering
\begin{tikzpicture}
\foreach \y in {0,1,2,3,4,5,6,7,8,9,10,11}
{
\draw (-4, \y-\y*0.2) -- (-4, \y-\y*0.2 - 0.3 );
}
\foreach \y in {10, 6, 5, 4, 3, 0}
{
\draw[draw=black] (-5.5,\y-\y*0.2) -- (-5.5,\y-\y*0.2+0.5) -- (-2.5,\y-\y*0.2+0.5) -- (-2.5,\y-\y*0.2) -- (-5.5,\y-\y*0.2);
}
\foreach \y in{11,9,8,7,2,1,-1}
{
\draw[draw=gray] (-5.25,\y-\y*0.2) -- (-5.25,\y-\y*0.2+0.5) -- (-2.75,\y-\y*0.2+0.5) -- (-2.75,\y-\y*0.2) -- (-5.25,\y-\y*0.2);
}
\foreach \y in {10}
{
\node[align=left] at (-4, \y-\y*0.2+0.225) {\scriptsize $(3\times3\times3)$ Conv};}
\foreach \y in {6,5,4,3}
{
\node[align=left] at (-4, \y-\y*0.2+0.225) {\scriptsize ResNext Bottleneck};}
\foreach \y in {0}
{
\node[align=left] at (-4, \y-\y*0.2+0.225) {\scriptsize Linear};
}
\foreach \y in {9}
{
\node[align=left] at (-4, \y-\y*0.2+0.225) {\scriptsize Batch Normalization};
}
\foreach \y in {8}
{
\node[align=left] at (-4, \y-\y*0.2+0.225) {\scriptsize ReLU};
}
\foreach \y in {2}
{
\node[align=left] at (-4, \y-\y*0.2+0.225) {\scriptsize ReLU+Dropout};
}
\foreach \y in {7, 1}
{
\node[align=left] at (-4, \y-\y*0.2+0.225) {\scriptsize Pooling};
}
\node[align=left] at (-4, 11-11*0.2+0.225) {\scriptsize Input};
\node[align=left] at (-4, -1- -1*0.2+0.225) {\scriptsize Output};

\node[align=left] at (-6.75, 9.05) {\scriptsize $(1\times50\times80\times126)$};
\node[align=left] at (-6.75, 8.25) {\scriptsize $(64\times25\times40\times63)$};
\node[align=left] at (-6.75, 5.85) {\scriptsize $(64\times13\times20\times32)$};
\node[align=left] at (-6.75, 5.05) {\scriptsize $(256\times13\times20\times32)$};
\node[align=left] at (-6.75, 4.25) {\scriptsize $(512\times7\times10\times16)$};
\node[align=left] at (-6.75, 3.45) {\scriptsize $(1024\times4\times5\times8)$};
\node[align=left] at (-6.75, 2.65) {\scriptsize $(2048\times2\times3\times4)$};
\node[align=left] at (-6.75, 1.05) {\scriptsize $(2048\times1)$};
\node[align=left] at (-6.75, 0.25) {\scriptsize $(6\times1)$};

\node[align=center] at (-4, -3- -3*0.2+0.225) {Architecture: $\mathcal{R}$\\ Trainable parameters:$\sim8.1 \times 10^{6}$\\ Size: $\sim 32.9$MB};
\end{tikzpicture}
\caption{Detailed description for architecture $\mathcal{R}$.}
\label{fig:arch2}
\end{figure}

\begin{figure}[h!]
\centering
\begin{tikzpicture}
\foreach \y in {-1,0,1,2,3,4,5,6,7,8,9,10,11}
{
\draw (-4, \y-\y*0.2) -- (-4, \y-\y*0.2 - 0.3 );
}
\foreach \y in {10, 6, 5, 4, 3, 0, -1}
{
\draw[draw=black] (-5.5,\y-\y*0.2) -- (-5.5,\y-\y*0.2+0.5) -- (-2.5,\y-\y*0.2+0.5) -- (-2.5,\y-\y*0.2) -- (-5.5,\y-\y*0.2);
}
\foreach \y in{11,9,8,7,2,1,-2}
{
\draw[draw=gray] (-5.25,\y-\y*0.2) -- (-5.25,\y-\y*0.2+0.5) -- (-2.75,\y-\y*0.2+0.5) -- (-2.75,\y-\y*0.2) -- (-5.25,\y-\y*0.2);
}
\foreach \y in {10}
{
\node[align=left] at (-4, \y-\y*0.2+0.225) {\scriptsize $(3\times3\times3)$ Conv};}
\foreach \y in {6,5,4,3}
{
\node[align=left] at (-4, \y-\y*0.2+0.225) {\scriptsize ResNext Bottleneck};}
\foreach \y in {-1}
{
\node[align=left] at (-4, \y-\y*0.2+0.225) {\scriptsize Linear};
}
\foreach \y in {9}
{
\node[align=left] at (-4, \y-\y*0.2+0.225) {\scriptsize Batch Normalization};
}
\foreach \y in {8}
{
\node[align=left] at (-4, \y-\y*0.2+0.225) {\scriptsize ReLU};
}
\foreach \y in {2}
{
\node[align=left] at (-4, \y-\y*0.2+0.225) {\scriptsize ReLU+Dropout};
}
\foreach \y in {7, 1}
{
\node[align=left] at (-4, \y-\y*0.2+0.225) {\scriptsize Pooling};
}
\node[align=left] at (-4, 11-11*0.2+0.225) {\scriptsize Input};
\node[align=left] at (-4, -0- -0*0.2+0.225) {\scriptsize LSTM};
\node[align=left] at (-4, -2- -2*0.2+0.225) {\scriptsize Output};

\node[align=left] at (-1.2, 9.05) {\scriptsize $(1\times50\times80\times126)$};
\node[align=left] at (-1.2, 8.25) {\scriptsize $(50\times64\times40\times63)$};
\node[align=left] at (-1.2, 5.85) {\scriptsize $(50\times64\times20\times32)$};
\node[align=left] at (-1.2, 5.05) {\scriptsize $(50\times64\times20\times32)$};
\node[align=left] at (-1.2, 4.25) {\scriptsize $(50\times256\times20\times32)$};
\node[align=left] at (-1.2, 3.45) {\scriptsize $(50\times512\times10\times16)$};
\node[align=left] at (-1.2, 2.65) {\scriptsize $(50\times1024\times5\times8)$};
\node[align=left] at (-1.2, 1.05) {\scriptsize $(50\times2048\times1\times1)$};
\node[align=left] at (-1.2, 0.25) {\scriptsize $(1024\times1)$};
\node[align=left] at (-1.2, -0.55) {\scriptsize $(6\times1)$};

\node[align=center] at (-4, -4- -4*0.2+0.225) {Architecture: $\mathcal{L}$\\ Trainable parameters:$\sim17.8 \times 10^{6}$\\ Size: $\sim 214$MB};
\end{tikzpicture}
\caption{Detailed description for architecture $\mathcal{L}$.}
\label{fig:arch3}
\end{figure}

\begin{table*}[!h]
\vspace{1em}
\centering
\caption{Validation and testing accuracy of the architectures $\mathcal{A}$, $\mathcal{R}$, and $\mathcal{L}$ for each class, as obtained for the respective evaluation splits described in Section~\ref{Framework} of the main text.}\label{tab:architecture_split_results}
\begin{tabularx}{10cm}{YYYYYYY}
 \cmidrule[1pt]{1-7} 
  &   \multicolumn{2}{c}{$\mathcal{A}$} &  \multicolumn{2}{c}{$\mathcal{R}$} &  \multicolumn{2}{c}{$\mathcal{L}$} \\
 \cmidrule{2-7}  
Class &  Validation & Test & Validation & Test & Validation & Test\\
    \cmidrule{1-7}  
(0)  & $57\%$ & $67\%$  & $60\%$ & $67\%$ & $75\%$ & $79\%$ \\
(1)  & $88\%$ & $86\%$  & $80\%$ & $81\%$ & $58\%$ & $72\%$ \\
(2)  & $83\%$ & $79\%$  & $77\%$ & $78\%$ & $85\%$ & $77\%$ \\
(3)  & $76\%$ & $79\%$  & $76\%$ & $71\%$ & $62\%$ & $55\%$ \\
(4)  & $55\%$ & $68\%$  & $54\%$ & $69\%$ & $55\%$ & $52\%$ \\
(5)  & $70\%$ & $69\%$  & $59\%$ & $63\%$ & $60\%$ & $65\%$ \\
    \cmidrule{1-7}
Total & $72\%$ & $74\%$ & $67\%$ & $72\%$ & $66\%$ & $67\%$\\
\cmidrule[1pt]{1-7} 
\end{tabularx}
\end{table*}

Fig.~\ref{fig:arch1} shows a detailed description of the neural network layers of $\mathcal{A}$, our 3D-CNN model, as well as the evolving size of the input as it is processed with a forward pass. This architecture is introduced in~\cite{vandersmissen2019indoor} as a lightweight model, only taking up a space of about $2.5$MB and coming with approximately $6.4 \times 10^{5}$ trainable parameters. The main purpose of this model is to be deployed in household environments. Thanks to its lightweight nature, a prediction can be performed in an efficient way, while reducing the cost of the required hardware.

On the data set splits explained in Section~\ref{Framework} of the main text, we train $\mathcal{A}$ for 100 epochs, using Adaptive Momentum~\citep{adaptive_momentum} with a learning rate of 0.001.

Fig.~\ref{fig:arch2} shows a detailed description of the second model used in our study ($\mathcal{R}$). This model is a variant of the ResNeXt model~\citep{resnext}, so to be able to handle data with a temporal dimension (e.g., video and radar data), and where this variant has been presented in~\citet{resnext_video}. The architecture shown in~Fig.~\ref{fig:arch2}, which contains four ResNeXt Bottleneck layers (more detailed information about such layers can be found in~\citet{resnext}), is significantly more complex compared to the model presented in Fig.~\ref{fig:arch1}. Indeed, this model contains approximately $8.1 \times 10^{6}$ trainable parameters (roughly $12$ times more than $\mathcal{A}$). Moreover, this model is also larger in terms of size, taking up a space of about $32.9$MB. To add, a single prediction made by the ResNeXt model takes about $9$ times longer than a single prediction made by the 3D-CNN model, thus possibly introducing significant time delays when it is deployed on similar hardware. The aforementioned considerations make it challenging for the ResNeXt model to be deployed in household environments with cheap hardware, not only because of its size, but also because of the time required to make a prediction. Nevertheless, we selected this model in order to be able to make a comparison, in terms of adversarial vulnerability, between a simple model that can be easily deployed and a larger model that is more capable.

On the data set splits explained in Section~\ref{Framework} of the main text, we train $\mathcal{R}$ for 50 epochs, using Adaptive Momentum~\citep{adaptive_momentum} with a learning rate of $1\mathrm{e}{-5}$. 

Fig.~\ref{fig:arch3} shows a detailed description of the third and the last model $\mathcal{L}$ used in our study. The architecture of $\mathcal{L}$ is similar to that of $\mathcal{R}$. However, $\mathcal{L}$ comes with a simple but crucial difference: it is a CNN-LSTM architecture that uses convolutions as feature extractors and that leverages an LSTM layer to discover the underlying relations along the temporal dimension. As such, the convolution operations are only performed on individual frames and not along the temporal dimension. This allows the employed LSTM to discover temporal relations and make judgements based on the information stored over an extended period of time. 
The down side of this model is its greater complexity in terms of storage and inference time. Due to the addition of an LSTM layer, the size of the model significantly increases compared to $\mathcal{R}$, containing $\sim17.8 \times 10^{6}$ trainable parameters and taking a space of about $214$MB. As a result, both a forward pass (prediction) and a backward pass (training) take significantly longer than in the case of $\mathcal{A}$ and $\mathcal{R}$. These properties pose an important challenge when employing such models using low-cost and low-power equipment in smart homes.  

The best performing model for this architecture is the one we trained on the data set splits explained in Section~\ref{Framework} of the main text, for 30 epochs, hereby using Adaptive Momentum~\citep{adaptive_momentum} with a learning rate of 0.00001.

Per-class accuracy of all three models observed for unseen data (i.e., validation and testing) is provided in Table~\ref{tab:architecture_split_results}. As can be seen, even though all of the employed models are vastly different in terms of architecture as well as trainable parameters, they achieve roughly similar results for the task at hand, making them suitable for a study on adversarial research.

\begin{figure*}[t!]
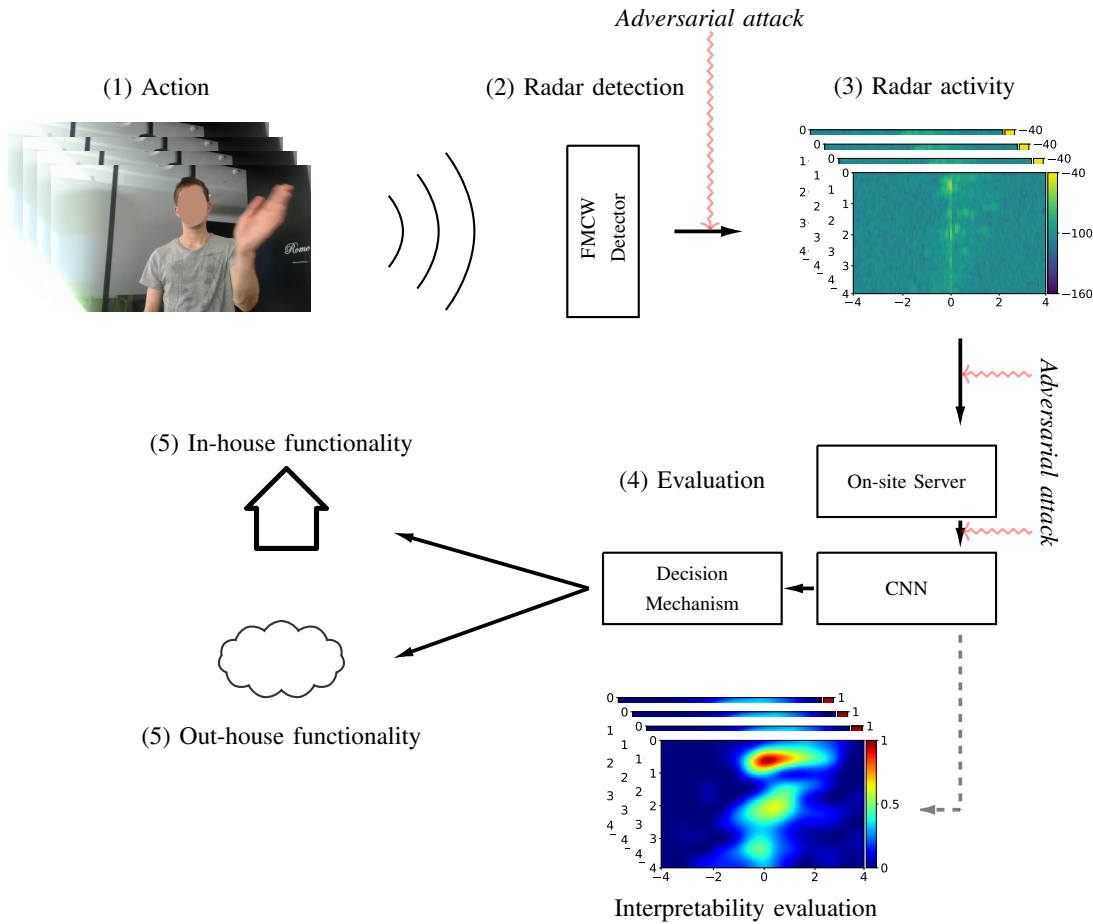

\centering
\begin{tikzpicture}[thick,scale=0.95, every node/.style={scale=0.95}]
\centering
\def\y{2.5} 
\def\x{-2.5} 
\node[inner sep=0pt] (vid1) at (\x, \y)  
    {\includegraphics[width=.2\textwidth]{radar_ims/010_mask.png}};
\node[inner sep=0pt] (vid2) at (\x + 0.2, \y-0.2)  
    {\includegraphics[width=.2\textwidth]{radar_ims/010_mask.png}};
\node[inner sep=0pt] (vid2) at (\x + 0.4, \y-0.4)  
    {\includegraphics[width=.2\textwidth]{radar_ims/010_mask.png}};
\node[inner sep=0pt] (vid2) at (\x + 0.6, \y-0.6)  
    {\includegraphics[width=.2\textwidth]{radar_ims/010_mask.png}};
\node[] at (-2.25, 4)  {(1) Action};
\def\y{2.5} 
\def\x{8.5} 
\node[inner sep=0pt] (vid1) at (\x, \y)  
    {\includegraphics[width=.2\textwidth]{radar_ims/10_org_im.pdf}};
\node[inner sep=0pt] (vid2) at (\x + 0.2, \y-0.2)  
    {\includegraphics[width=.2\textwidth]{radar_ims/10_org_im.pdf}};
\node[inner sep=0pt] (vid2) at (\x + 0.4, \y-0.4)  
    {\includegraphics[width=.2\textwidth]{radar_ims/10_org_im.pdf}};
\node[inner sep=0pt] (vid2) at (\x + 0.6, \y-0.6)  
    {\includegraphics[width=.2\textwidth]{radar_ims/10_org_im.pdf}};

\draw[draw=black] (3.5, 3.2) -- (4.5, 3.2) -- (4.5, 0.8) -- (3.5, 0.8) -- (3.5, 3.2);
\node[align=center,rotate=90] at (4, 2.05) {\footnotesize FMCW\\\footnotesize Detector};
\node[] at (3.75, 4)  {(2) Radar detection};    

\node[] at (8.5, 4)  {(3) Radar activity};
\draw[-{Latex[width=1.5mm]}, line width=0.5mm] (5, 2) -- (6, 2);
\draw [black, xshift=0cm] plot [smooth, tension=1] coordinates { (1, 2.5) (1.2, 2) (1, 1.5)};
\draw [black, xshift=0cm] plot [smooth, tension=1] coordinates { (1.4, 2.8) (1.7, 2) (1.4, 1.2)};
\draw [black, xshift=0cm] plot [smooth, tension=1] coordinates { (1.8, 3.1) (2.2, 2) (1.8, 0.9)};
\def\y{-4}  

\draw[-{Latex[width=1.3mm]}, line width=0.5mm] (9, 0.5) -- (9, -0.8);
\def\loc_ser{-1} 
\draw[draw=black] (7, \loc_ser) -- (9.5, \loc_ser) -- (9.5,  \loc_ser-1) -- (7,  \loc_ser-1) -- (7,  \loc_ser);
\node[align=center] at (8.25,  \loc_ser-0.5)  {\footnotesize On-site Server};
\node[align=center] at (5.25, -1.5)  {(4) Evaluation};
\draw[-{Latex[width=1.3mm]}, line width=0.5mm] (9, -2.05) -- (9, -2.5);
\def\loc_ser{-2.5} 
\draw[draw=black] (7, \loc_ser) -- (9.5, \loc_ser) -- (9.5,  \loc_ser-1) -- (7,  \loc_ser-1) -- (7,  \loc_ser);
\node[align=center] at (8.25,  \loc_ser-0.5)  {\footnotesize CNN};
\draw[-{Latex[width=1.3mm]}, line width=0.5mm] (6.95, -3) -- (6.45, -3);
\def\ml_y{-2.5} 
\def\mlx{4} 
\draw[draw=black] (\mlx, \ml_y) -- (\mlx+2.5, \ml_y) -- (\mlx+2.5, \ml_y-1) -- (\mlx, \ml_y-1) -- (\mlx, \ml_y);
\node[align=center] at (\mlx+1.25,  \ml_y-0.5)  {\footnotesize Decision\\\footnotesize Mechanism};

\node [cloud, draw,cloud puffs=11,cloud puff arc=120, aspect=2, inner ysep=1em, opacity=0.8, scale=0.85] at (-0.5, -4) {};
\node[align=center, opacity=0.5] at (-0.5, -4)  {};
\node[align=center] at (-0.5, -5.1)  {(5) Out-house functionality};
\node[align=center] at (-0.5, -1)  {(5) In-house functionality};
\node[align=center] at (-0.5, -1.95)  {\fontsize{40}{38} \selectfont {\Hut}};
\draw[-{Latex[width=1.3mm]}, line width=0.5mm] (3.8, -3) -- (1, -2.2);
\draw[-{Latex[width=1.3mm,]}, line width=0.5mm] (3.8, -3) -- (1, -4);

\def\x{5.8} 
\def\y{-5.5} 
\node[inner sep=0pt] (vid1) at (\x, \y)  
    {\includegraphics[width=.2\textwidth]{radar_ims/10_cam_pos.pdf}};
\node[inner sep=0pt] (vid2) at (\x + 0.2, \y-0.2)  
    {\includegraphics[width=.2\textwidth]{radar_ims/10_cam_pos.pdf}};
\node[inner sep=0pt] (vid2) at (\x + 0.4, \y-0.4)  
    {\includegraphics[width=.2\textwidth]{radar_ims/10_cam_pos.pdf}};
\node[inner sep=0pt] (vid2) at (\x + 0.6, \y-0.6)  
    {\includegraphics[width=.2\textwidth]{radar_ims/10_cam_pos.pdf}};
\draw[line width=0.5mm, dashed, opacity=0.5] (9, -3.65) -- (9, -6.1);
\draw[-{Latex[width=1.3mm,]}, line width=0.5mm, dashed, opacity=0.5] (9, -6.1) -- (8.3, -6.1);
\node[align=center] at (6, -7.5)  {Interpretability evaluation};

\node[align=center] at (5.5, 5)  {\textit{Adversarial attack}};
\draw [->,
line join=round,
decorate, decoration={
    zigzag,
    segment length=4,
    amplitude=.9,post=lineto,
    post length=3pt,
}, color=red, line width=0.3mm, opacity=0.4]  (5.5, 4.8) -- (5.5, 2);

\node[align=center, rotate=270] at (10.25, -1.1)  {\textit{Adversarial attack}};
\draw [->,
line join=round,
decorate, decoration={
    zigzag,
    segment length=4,
    amplitude=.9,post=lineto,
    post length=3pt,
}, color=red, line width=0.3mm, opacity=0.4]  (10, 0) -- (9, 0);
\draw [->,
line join=round,
decorate, decoration={
    zigzag,
    segment length=4,
    amplitude=.9,post=lineto,
    post length=3pt,
}, color=red, line width=0.3mm, opacity=0.4]  (10, -2.2) -- (9, -2.2);
\end{tikzpicture}
\caption{A visual summary of the flow of events in a household when an action is performed, including the entry points for possible adversarial attacks.}
\label{fig:visual_summary}
\end{figure*}

\section*{Visual Summary of the Evaluated Scenario}

Fig.~\ref{fig:visual_summary} contains a visual description of the flow of events for the scenario evaluated in the main text. To that end, when any action is performed in a household environment, a radar sensor (in this case an FMCW radar device) is able to detect this movement, with further processing leading to a sequence of range-Doppler frames. These frames are then sent to an on-site server (small and portable hardware) that contains a CNN, with the CNN performing a prediction on the frames received. Given the prediction made by the CNN, either a functionality inside the house (e.g., lights) or outside the house (e.g., calling emergency services) may be triggered. We conjecture three possible entry points for adversarial attacks: (1) when the radar frames are generated, (2) when the radar frames are being transferred from the detector to the on-site server, and (3) right before the prediction. In the end, we evaluate the vulnerability of predictive models that may be deployed in households and perform an investigation of the relationship between adversarial examples and the interpretability of neural networks.  

\begin{table*}[t]
\centering
\caption{Median value (interquartile range) of the $L_2$ and $L_{\infty}$ distances obtained for $1000$ adversarial optimizations, as well as their success rate, for $\mathcal{L}$ and data sets described in Section~\ref{Framework} of the main text. For easier comprehension, the threat models are listed from most permissive to least permissive. $L_{\infty}$ values less than $0.003$ are rolled up to $0.003$ (this is approximately the smallest amount of perturbation required to change a pixel value by $1$, so to make discretization possible).}
\begin{tabular}{ccccccccc}
 \cmidrule[1pt]{1-9}
   \multirow{2}{*}{\shortstack{Threat\\model}}  & \multirow{2}{*}{\shortstack{Source\\model}} & \multirow{2}{*}{\shortstack{Target\\model}} & \multicolumn{3}{c}{BIM} & \multicolumn{3}{c}{CW}  \\
 \cmidrule[0.25pt]{4-9}
  ~ &   ~ & ~ & $L_2$ & $L_{\infty}$ & Success \% & $L_2$ & $L_{\infty}$ & Success \% \\
\cmidrule[0.4pt]{1-9}
\multirow{2}{*}{WB} & \multirow{2}{*}{$\mathcal{L}_{\{S_+,S_-\}}$} & \multirow{2}{*}{$\mathcal{L}_{\{S_+,S_-\}}$}	 & $0.26$ & $0.003$ & \multirow{2}{*}{$100\%$} & $0.30$ & $0.01$ & \multirow{2}{*}{$84\%$}\\
~ & ~ & ~ & $(0.29)$ & $(0.003)$ & ~ &  $(0.30)$ & $(0.01)$\\
\cmidrule[0.4pt]{1-9}
 
\multirow{2}{*}{BB:1} & \multirow{2}{*}{$\mathcal{L}_{\{S_+,S_-\}}$} & \multirow{2}{*}{$\mathcal{L}_{\{S_-,S_+\}}$}	 & $0.32$ & $0.003$ & \multirow{2}{*}{$13\%$} & $0.33$ & $0.01$ & \multirow{2}{*}{$11\%$}\\
~ & ~ & ~ & $(0.16)$ & $(0.003)$ & ~ &  $(0.47)$ & $(0.02)$\\
\cmidrule[0.4pt]{1-9}

\multirow{4}{*}{BB:2} & \multirow{2}{*}{$\mathcal{A}_{\{S_+,S_-\}}$}  & \multirow{2}{*}{$\mathcal{L}_{\{S_+,S_-\}}$} 	 & $1.96$ & $0.009$ & \multirow{2}{*}{$35\%$} & $0.71$ & $0.02$ & \multirow{2}{*}{$32\%$}\\
~ & ~ & ~ & $(2.15)$ & $(0.006)$ & ~ & $(1.09)$ & $(0.001)$ \\
 \cmidrule[0.25pt]{2-9}
~ & \multirow{2}{*}{$\mathcal{R}_{\{S_+,S_-\}}$}  &  \multirow{2}{*}{$\mathcal{L}_{\{S_+,S_-\}}$} & $0.28$ & $0.006$ &  \multirow{2}{*}{$38\%$} & $0.27$ & $0.03$ &  \multirow{2}{*}{$34\%$} \\
~ & ~ & ~ & $(0.31)$ & $(0.003)$ & ~ & $(0.23)$ & $(0.02)$ & ~ \\ 
\cmidrule[0.4pt]{1-9}
 
\multirow{4}{*}{BB:3} & \multirow{2}{*}{$\mathcal{A}_{\{S_+,S_-\}}$}& \multirow{2}{*}{$\mathcal{L}_{\{S_-,S_+\}}$}	 & $2.07$ & $0.012$ & \multirow{2}{*}{$26\%$} & $0.84$ & $0.03$ & \multirow{2}{*}{$23\%$}\\
~ & ~ & ~ & $(1.45)$ & $(0.011)$ & ~ &  $(0.93)$ & $(0.07)$ \\
 \cmidrule[0.25pt]{2-9}
~ &\multirow{2}{*}{$\mathcal{R}_{\{S_+,S_-\}}$} & \multirow{2}{*}{$\mathcal{L}_{\{S_-,S_+\}}$}	 & $0.36$ & $0.009$ &  \multirow{2}{*}{$24\%$} & $0.28$ & $0.02$ &  \multirow{2}{*}{$19\%$} \\
~ & ~ & ~ & $(0.52)$ & $(0.007)$ & ~ & $(0.17)$ & $(0.03)$ & ~ \\ 

 \cmidrule[1pt]{1-9} 
\end{tabular}
\label{tbl:attack-table-lstm}
\end{table*}

\section*{Threat Model}
The configuration of threats and their taxonomy used in this study is listed in Fig.~\ref{fig:threat_model}. In terms of the amount of knowledge the adversary can make use of, we explain the different scenarios, from the most permissive scenario to the scenario that is the most limiting:

\begin{itemize}
    \item \textbf{Training data and model}\,\textendash\,The adversary has access to the model that is performing the classification task, as well as the data used to train this model.
    \item \textbf{Only trained model}\,\textendash\,The adversary has access to the model that is performing the classification, but not to the underlying training data.
    \item \textbf{Only training data}\,\textendash\,The adversary has access to the training data that the underlying model has been trained with, allowing the adversary to leverage knowledge about the underlying distribution of the data.
    \item \textbf{Only architecture}\,\textendash\,The adversary has access to the architecture that is performing the classification task, but not to the training data. This means that the attack must be implemented using data that have not been seen by the model before.
    \item \textbf{Surrogate}\,\textendash\,The adversary has access to a model (i.e., a surrogate) that has been trained with similar data that the underlying system has been trained with. This allows the adversary to leverage knowledge about a model that has been trained on data that are similar in terms of distribution to the training data of the underlying system.
\end{itemize}

The attacks originating from the first two cases are usually referred to as \textit{white-box} attacks, which means that the attacker has access to the underlying system, whereas the last three cases are referred to as \textit{black-box} attacks, which means the attacker does not have access to the underlying system. In the main text, we analyzed a white-box scenario, as well as several black-box scenarios.

For the same threat configuration, attacks that can be performed based on the knowledge of the adversary discussed above can be listed from easier to harder in the following way:

\begin{itemize}
    \item \textbf{Confidence reduction}\,\textendash\,Reduce the output confidence of the prediction.
    \item \textbf{Misclassification}\,\textendash\,Change the prediction from a correct one to an (unspecified) incorrect one.
    \item \textbf{Targeted misclassification}\,\textendash\,Force the prediction to become a specified class that is different from the correct one.
    \item \textbf{Targeted misclassification with a localized attack}\,\textendash\,Force the prediction to become a specified class that is different from the correct one and, while doing so, limit the attack (i.e., the perturbation) to selected regions of the input.
\end{itemize}

\section*{Constraints on Adversarial Attacks}
For the evaluation performed in the main text, we imposed the following constraints on the generation of adversarial examples:
\begin{itemize}
\item \textbf{Box constraint}\,\textendash\, In order to ensure that the generated adversarial example is a valid image, its values are constrained as follows: $\mathbf{X}_n \in [0, 1]$, with $0$ denoting black and $1$ denoting white. However, different from the image domain, the radar data we use in this study always contain a portion of noise, which limits the values even further when the radar signal is converted to a sequence of RD frames. Thus, for the radar signal, we select the box constraint as $\mathbf{X}_n \in [0.31, 0.83]$, with $0.31$ and $0.83$ representing the smallest and the largest value present in our data set, respectively.

\item \textbf{Time constraint}\,\textendash\, Threat scenarios that are tackled in this study consider data obtained from sensors manipulated by an adversary. However, we only assume the adversary to be capable of manipulating the frames (i.e., adversarial attacks). In doing so, we assume there is no delay between frame capturing and the transfer of these frames to the underlying model. As a result, in order to work with a realistic attack scenario, we assume there is a limited amount of time available to implement perturbations. In particular, we restrict the amount of time available for adversarial optimization to one second. This limitation approximately corresponds to $200$ optimization iterations for BIM and $72$ iterations for the CW attack on a single Titan-X GPU for model $\mathcal{A}$.

\item \textbf{Discretization}\,\textendash\, 
As described above, the input data are bounded between $0.31$ and $0.83$. However, when the input is represented as a grayscale image, these values must be represented as integers between $0$ and $255$. Thus, if a value does not have a direct integer correspondance, it is rounded to the closest integer. Studies that investigate adversarial examples often disregard the discretization property of the produced adversarial examples, hereby providing results for images that are impossible to represent in reality. This topic is discussed in more detail in~\citet{CW_Attack}. In this study, we make sure that the generated adversarial examples can be represented as valid grayscale images.
\end{itemize}

\section*{CNN-LSTM Experiments}

\begin{table}[!t]
\centering
\caption{Median value (interquartile range) of the $L_2$ and $L_{\infty}$ distances obtained for $1000$ adversarial optimizations, as well as their success rate, for $\mathcal{L}$ and data sets described in Section~\ref{Framework}, hereby using the padding attack described in Section~\ref{Adversarial Padding for Radar Data} of the main text.}
\begin{tabular}{ccccc}
 \cmidrule[1pt]{1-5}
    \multirow{2}{*}{\shortstack{Source\\model}} & \multirow{2}{*}{\shortstack{Target\\model}} & \multicolumn{3}{c}{Padding attack}  \\
\cmidrule[0.25pt]{3-5}
   ~ & ~ & $L_2$ & $L_{\infty}$ & Success \% \\
\cmidrule[0.4pt]{1-5}
\multirow{2}{*}{$\mathcal{L}_{\{S_+,S_-\}}$} & \multirow{2}{*}{$\mathcal{L}_{\{S_+,S_-\}}$}	 & $4.47$ & $0.015$ & \multirow{2}{*}{$78\%$} \\
 ~ & ~ & $(0.72)$ & $-$ & ~ \\
\cmidrule[0.4pt]{1-5}
 
\multirow{2}{*}{$\mathcal{L}_{\{S_+,S_-\}}$} & \multirow{2}{*}{$\mathcal{L}_{\{S_-,S_+\}}$}	 & $3.94$ & $0.015$ & \multirow{2}{*}{$73\%$} \\
 ~ & ~ & $(0.91)$ & $-$ & ~ \\
  \cmidrule[0.25pt]{1-5}

\multirow{2}{*}{$\mathcal{A}_{\{S_+,S_-\}}$}  & \multirow{2}{*}{$\mathcal{L}_{\{S_+,S_-\}}$} 	 & $3.58$ & $0.012$ & \multirow{2}{*}{$57\%$} \\
 ~ & ~ & $(1.16)$ & $-$ & ~ \\
  \cmidrule[0.25pt]{1-5}
 \multirow{2}{*}{$\mathcal{R}_{\{S_+,S_-\}}$}  &  \multirow{2}{*}{$\mathcal{L}_{\{S_+,S_-\}}$} & $4.11$ & $0.012$ &  \multirow{2}{*}{$63\%$} \\
 ~ & ~ & $(1.23)$ & $-$ & ~ \\ 
\cmidrule[0.4pt]{1-5}
 
 \multirow{2}{*}{$\mathcal{A}_{\{S_+,S_-\}}$}& \multirow{2}{*}{$\mathcal{L}_{\{S_-,S_+\}}$}	 & $4.81$ & $0.012$ & \multirow{2}{*}{$37\%$}\\
 ~ & ~ & $(1.46)$ & $-$ & ~  \\
  \cmidrule[0.25pt]{1-5}
\multirow{2}{*}{$\mathcal{R}_{\{S_+,S_-\}}$} & \multirow{2}{*}{$\mathcal{L}_{\{S_-,S_+\}}$}	 & $4.99$ & $0.012$ &  \multirow{2}{*}{$33\%$} \\
~ & ~ & $(0.93)$ & $-$ & ~  \\ 

 \cmidrule[1pt]{1-5} 
\end{tabular}
\label{tbl:pad-attack-table-lstm}
\end{table}

In Table~\ref{tbl:attack-table-lstm}, we provide the results (success rate and both $L_2$ and $L_{\infty}$ distances of the adversarial examples produced) for adversarial attacks performed on the CNN-LSTM architecture $\mathcal{L}$, for attack scenarios described in Section~\ref{Threat model} of the main text. In Table~\ref{tbl:pad-attack-table-lstm}, we present the same statistics for the padding attack explained in Section~~\ref{Adversarial Padding for Radar Data} of the main text.

Based on the results presented in Table~\ref{tbl:attack-table-lstm} and Table~\ref{tbl:pad-attack-table-lstm}, our observations are as follows:

\begin{itemize}
\item The first three observations we made in the main text for the models $\mathcal{A}$ and $\mathcal{R}$ regarding the properties of the selected attacks also hold true for attacks against the CNN-LSTM model $\mathcal{L}$. We also observe that adversarial examples originated from fully convolutional architectures are able to deceive the CNN-LSTM model without requiring an extra effort (i.e., another specialized attack).

\item We do not observe a significant difference in robustness of the CNN-LSTM model compared to the other two fully convolutional models.

\item Compared to Table~2 in the main text, the biggest difference is the low success rate of attacks for the case of threat scenario BB:1. The reason for this low success rate is the number of parameters in the model $\mathcal{L}$, which significantly increases the time to generate adversarial examples. When using this model for generating adversarial examples under limited time settings such as the ones we employ, the success rate drops significantly.  

\item The proposed adversarial padding attack is also able to create adversarial examples that transfer to CNN-LSTM models. In particular, even though the attack success rate is on average slightly lower compared to the fully convolutional models, this difference is not large, considering the difference in the initial accuracy of the models.

\end{itemize}

Hence, we observe that in this particular case, employing a CNN-LSTM model does not significantly improve (or reduce) adversarial robustness compared to fully convolutional models.

\section*{Additional Figures}

\begin{figure*}[t!]
\centering
{Original Frame}\phantom{--------------------------------}{Perturbed Frame}
\\
\rotatebox[origin=l]{90}{\phantom{-----}$L_2$ Distance: $0$}
\hspace{0.5em}
\subfloat{{\includegraphics[width=6cm, height=3.5cm]{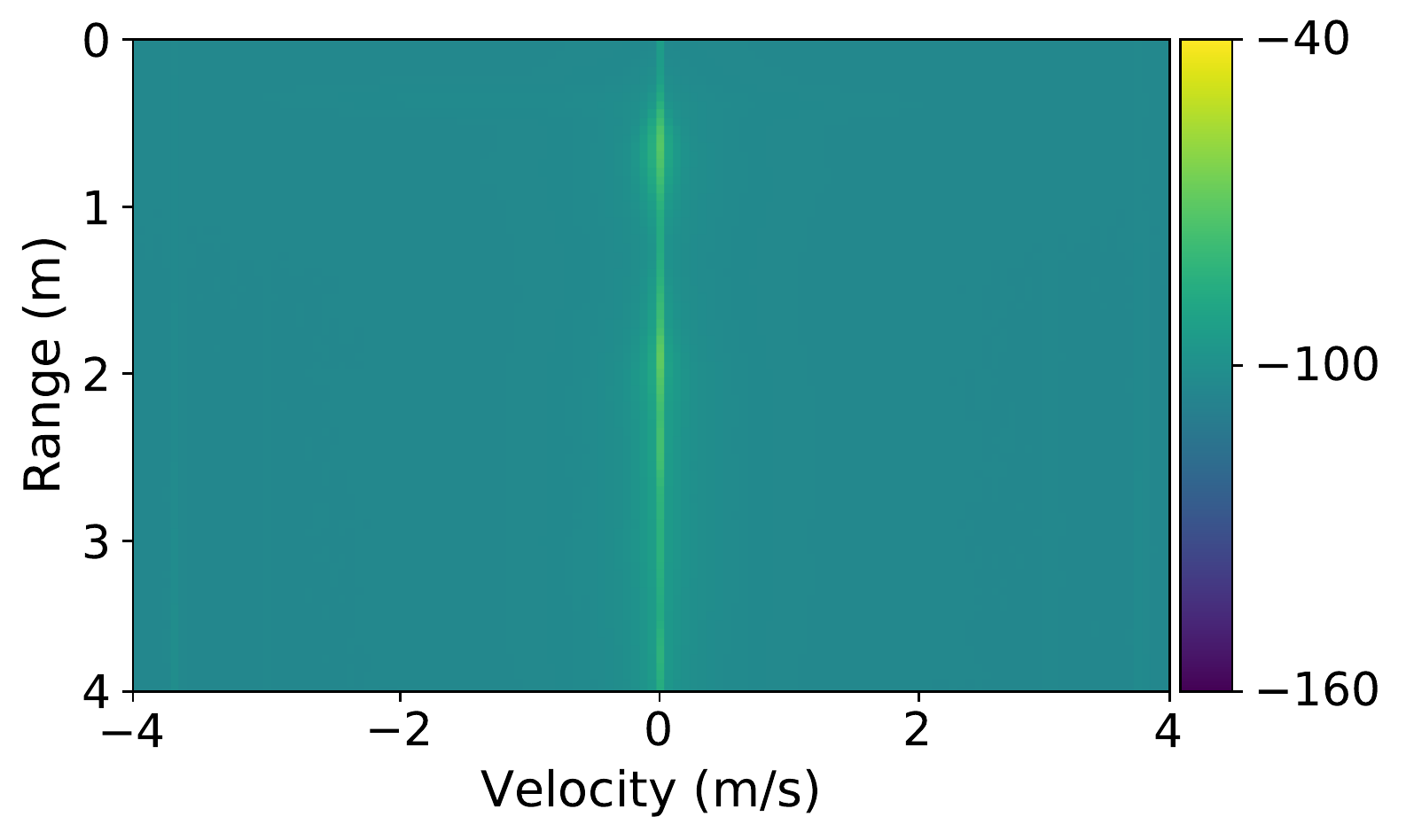}}}
\subfloat{{\includegraphics[width=6cm, height=3.5cm]{other_images/l2_images/im5_48_org_frame_hmap.pdf}}}
\\
\rotatebox[origin=l]{90}{\phantom{-----}$L_2$ Distance: $0.054$}
\hspace{0.5em}
\subfloat{{\includegraphics[width=6cm, height=3.5cm]{other_images/l2_images/im5_48_org_frame_hmap.pdf}}}
\subfloat{{\includegraphics[width=6cm, height=3.5cm]{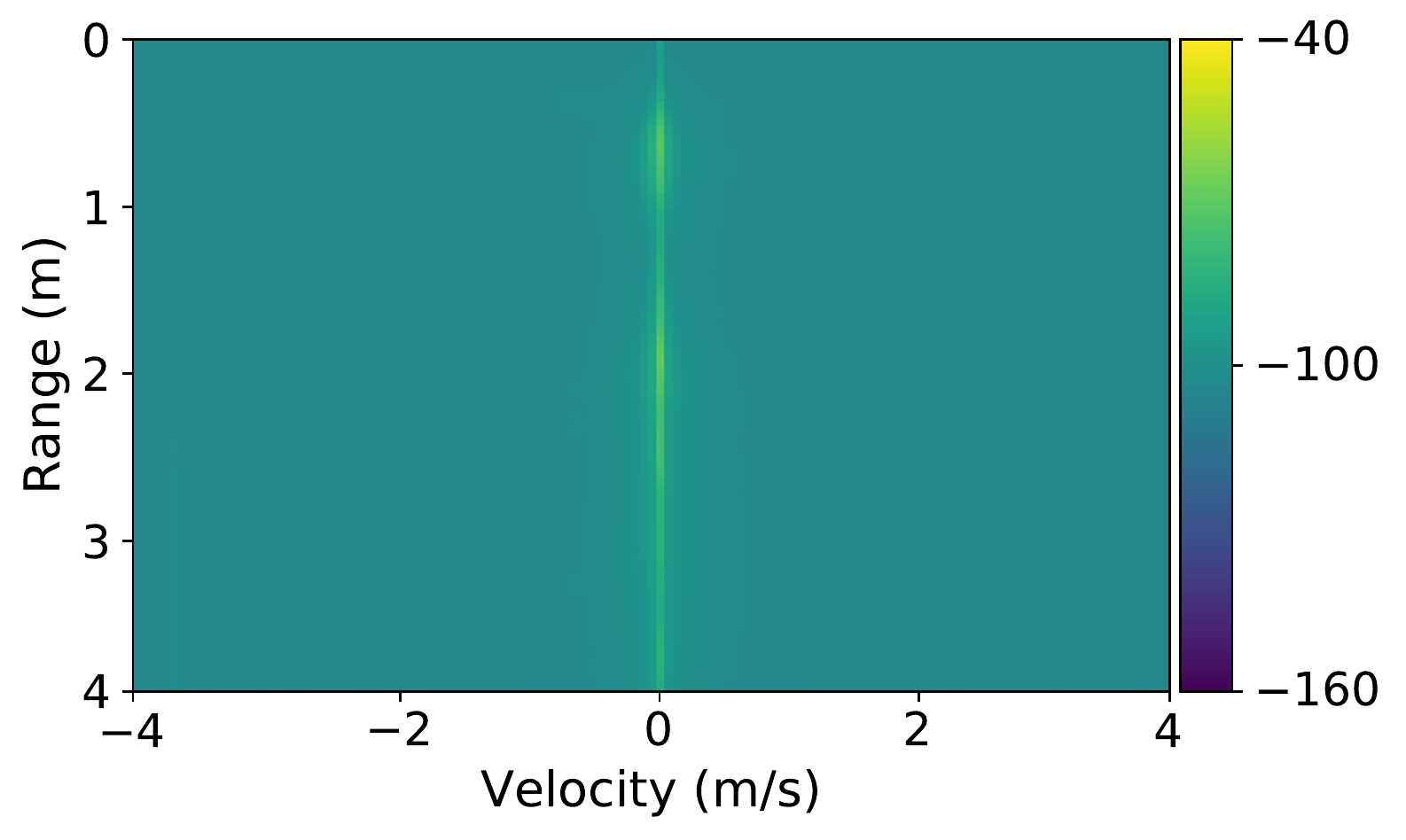}}}
\\
\rotatebox[origin=l]{90}{\phantom{-----}$L_2$ Distance: $0.869$}
\hspace{0.5em}
\subfloat{{\includegraphics[width=6cm, height=3.5cm]{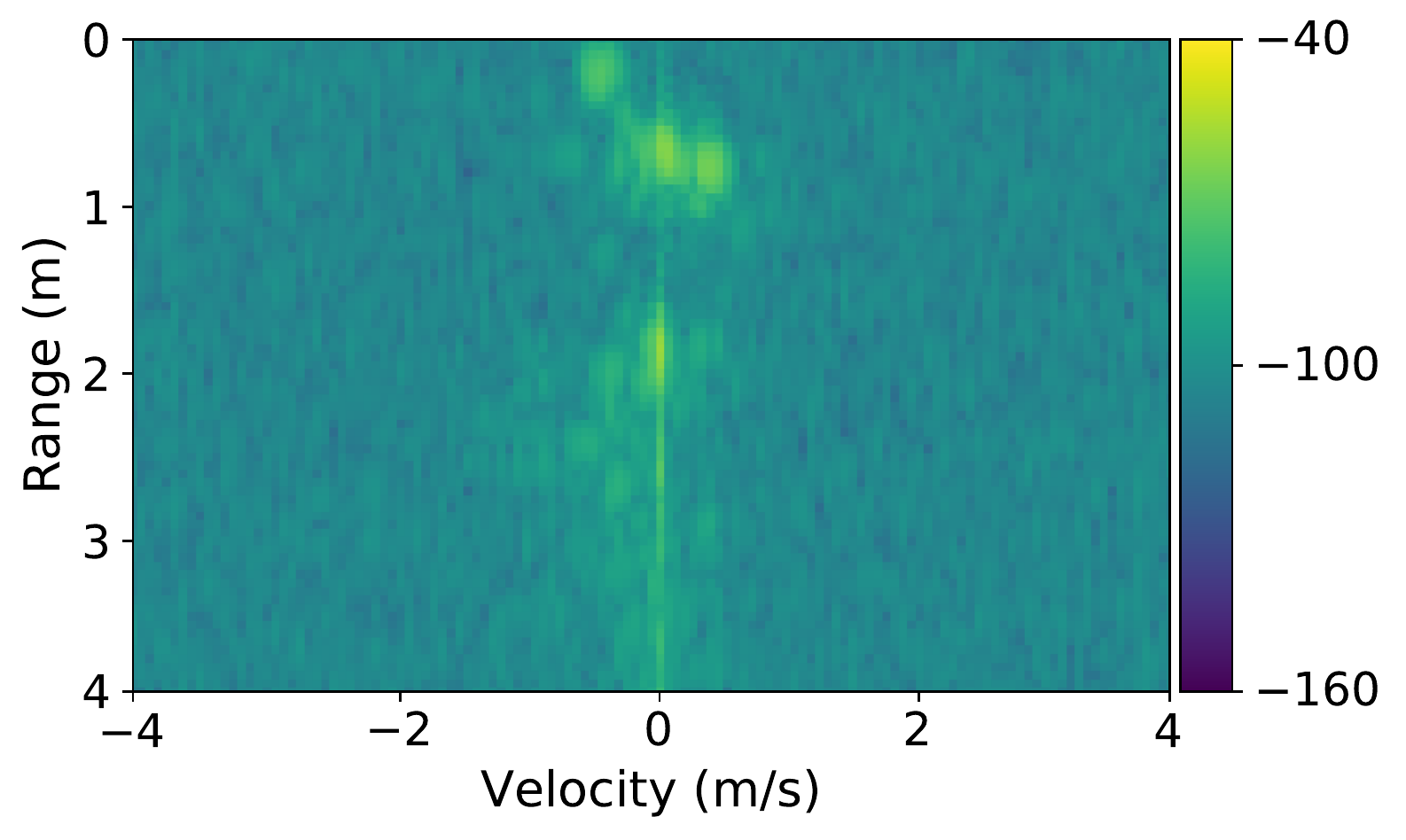}}}
\subfloat{{\includegraphics[width=6cm, height=3.5cm]{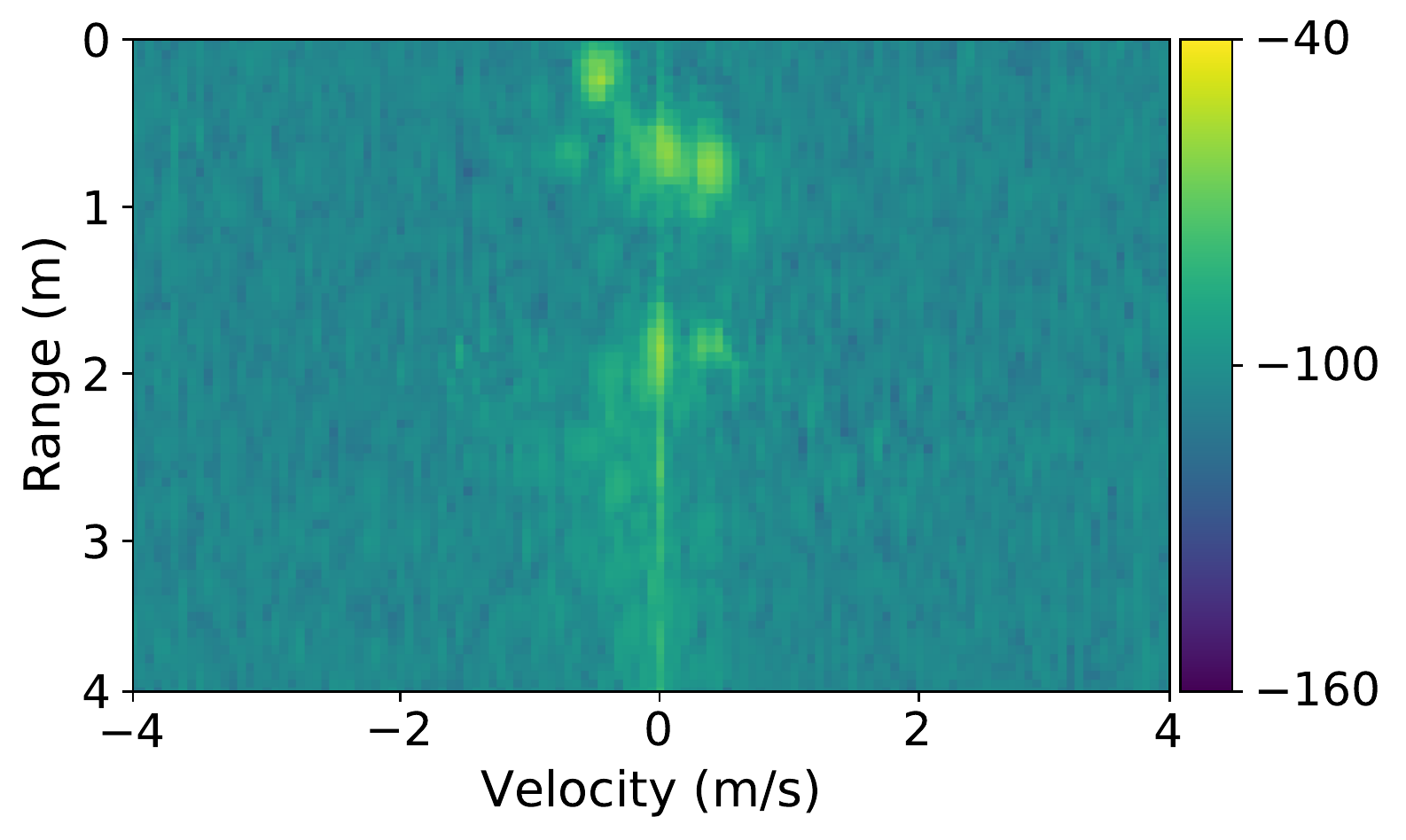}}}
\\
\rotatebox[origin=l]{90}{\phantom{-----}$L_2$ Distance: $1.638$}
\hspace{0.5em}
\subfloat{{\includegraphics[width=6cm, height=3.5cm]{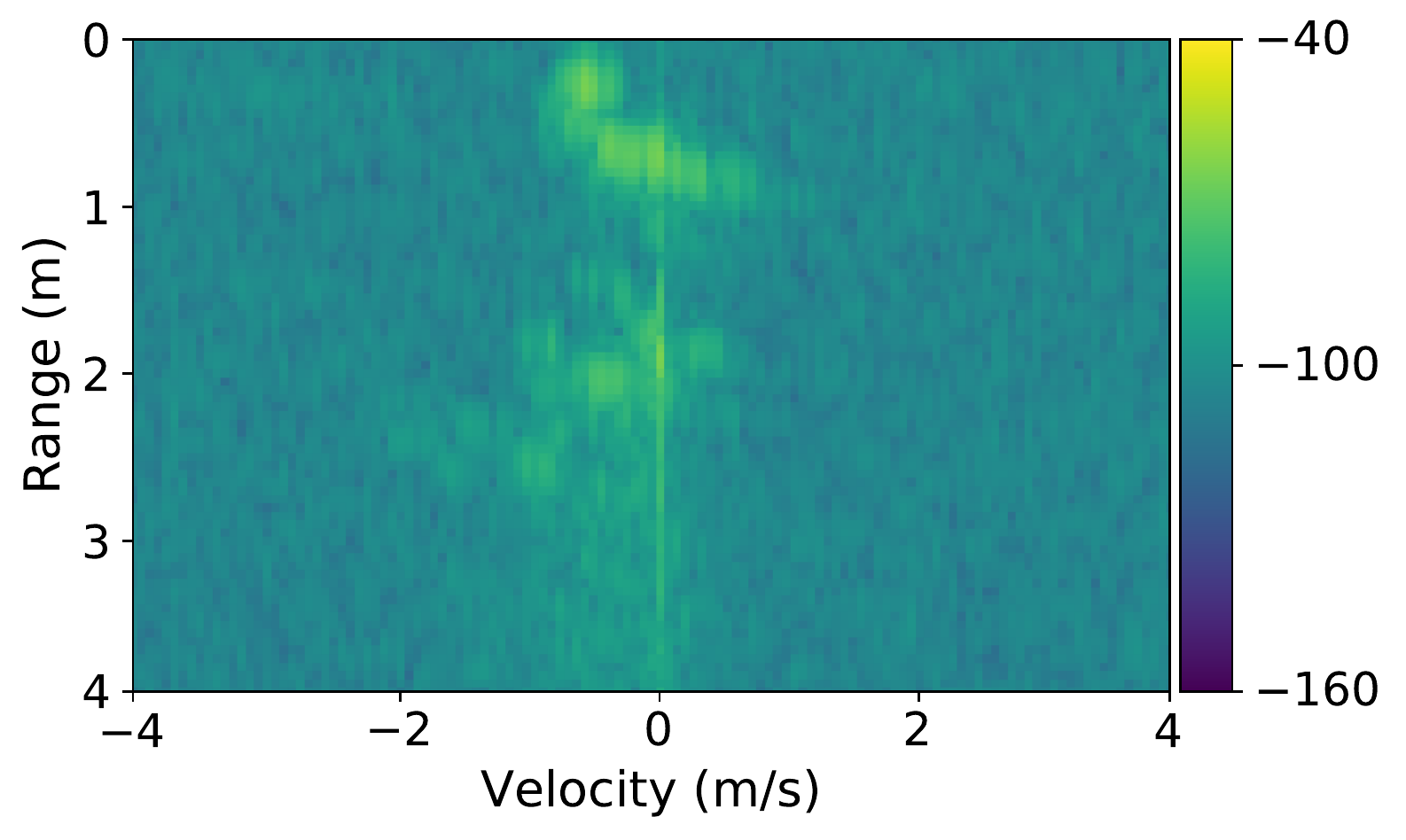}}}
\subfloat{{\includegraphics[width=6cm, height=3.5cm]{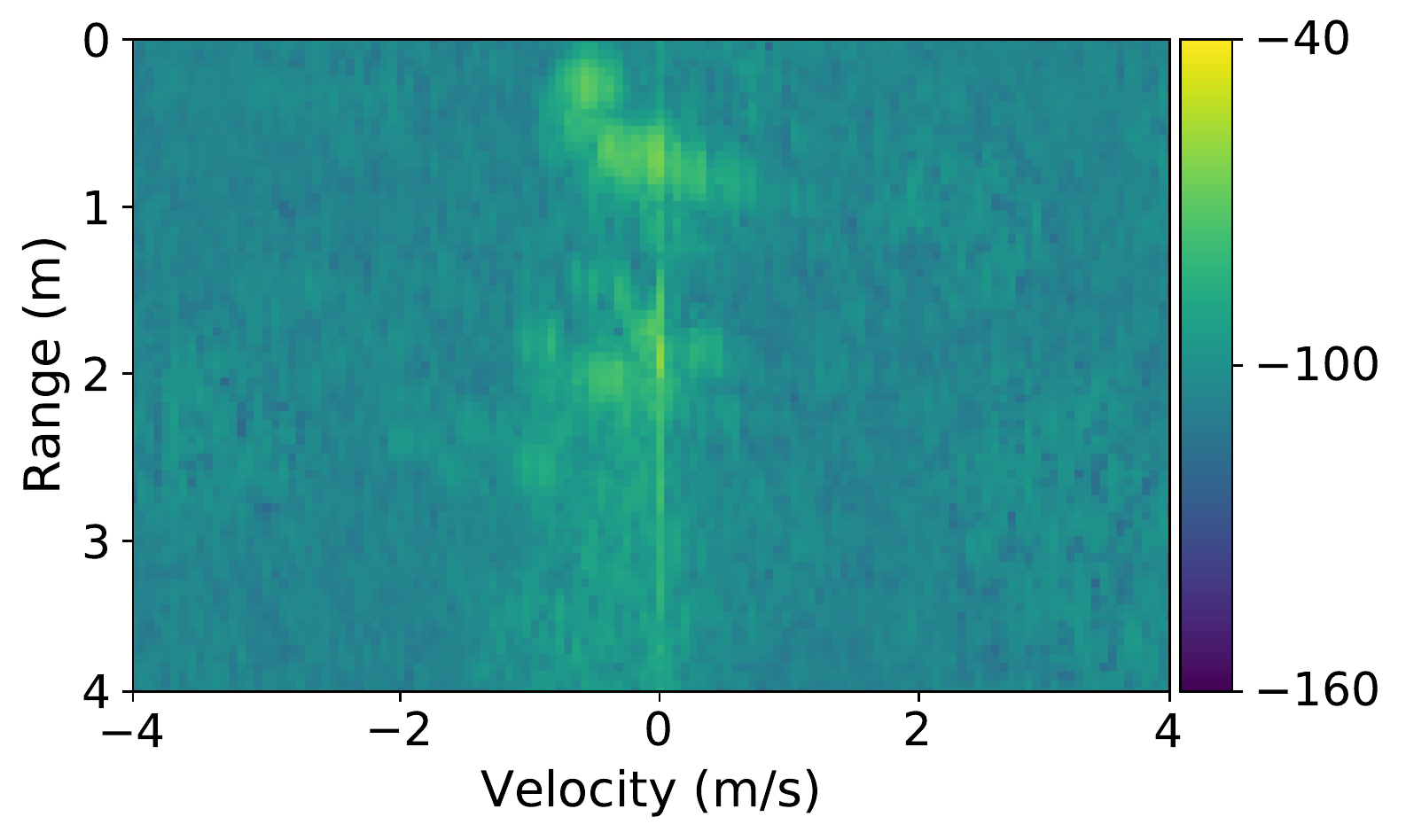}}}
\\
\rotatebox[origin=l]{90}{\phantom{-----}$L_2$ Distance: $2.919$}
\hspace{0.5em}
\subfloat{{\includegraphics[width=6cm, height=3.5cm]{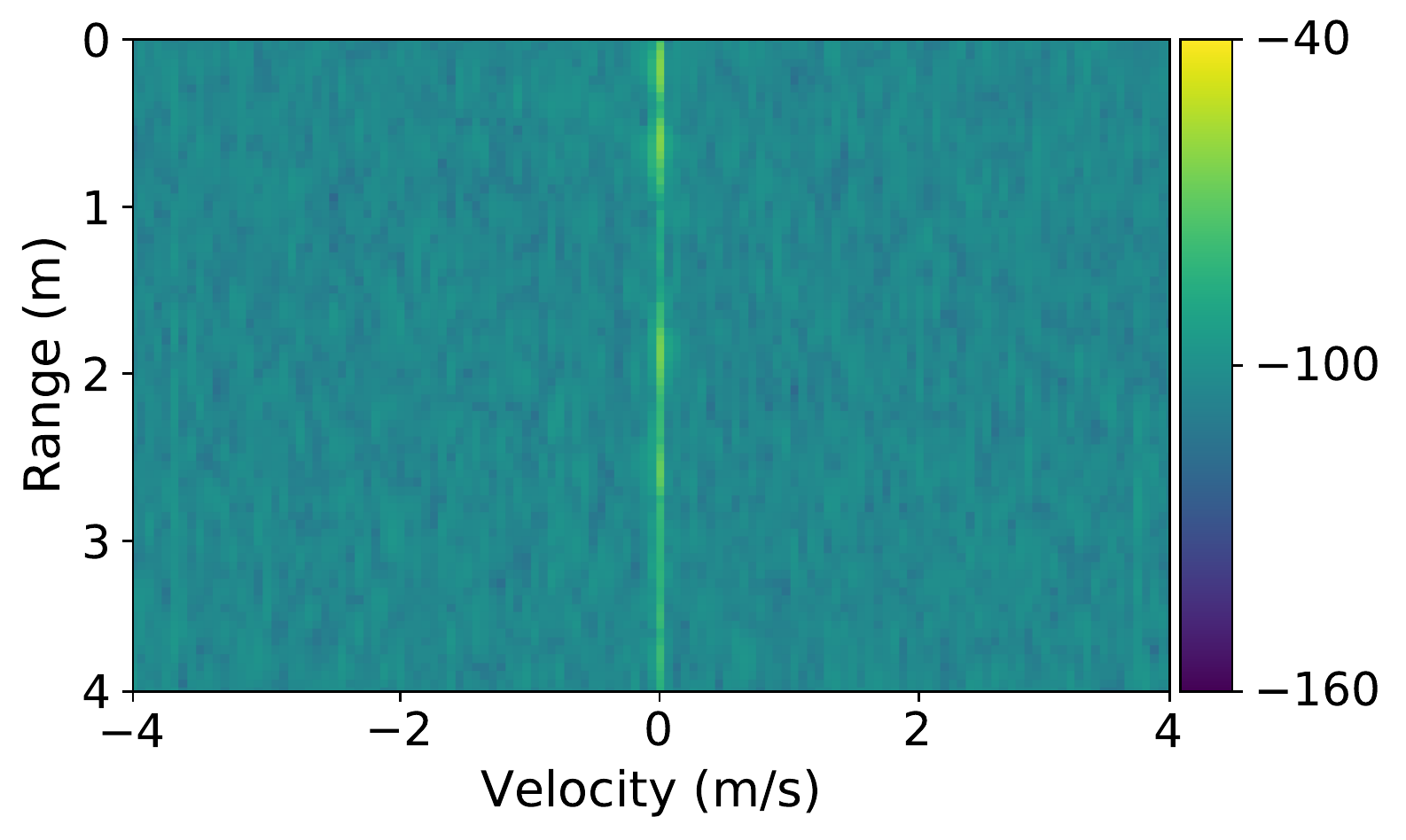}}}
\subfloat{{\includegraphics[width=6cm, height=3.5cm]{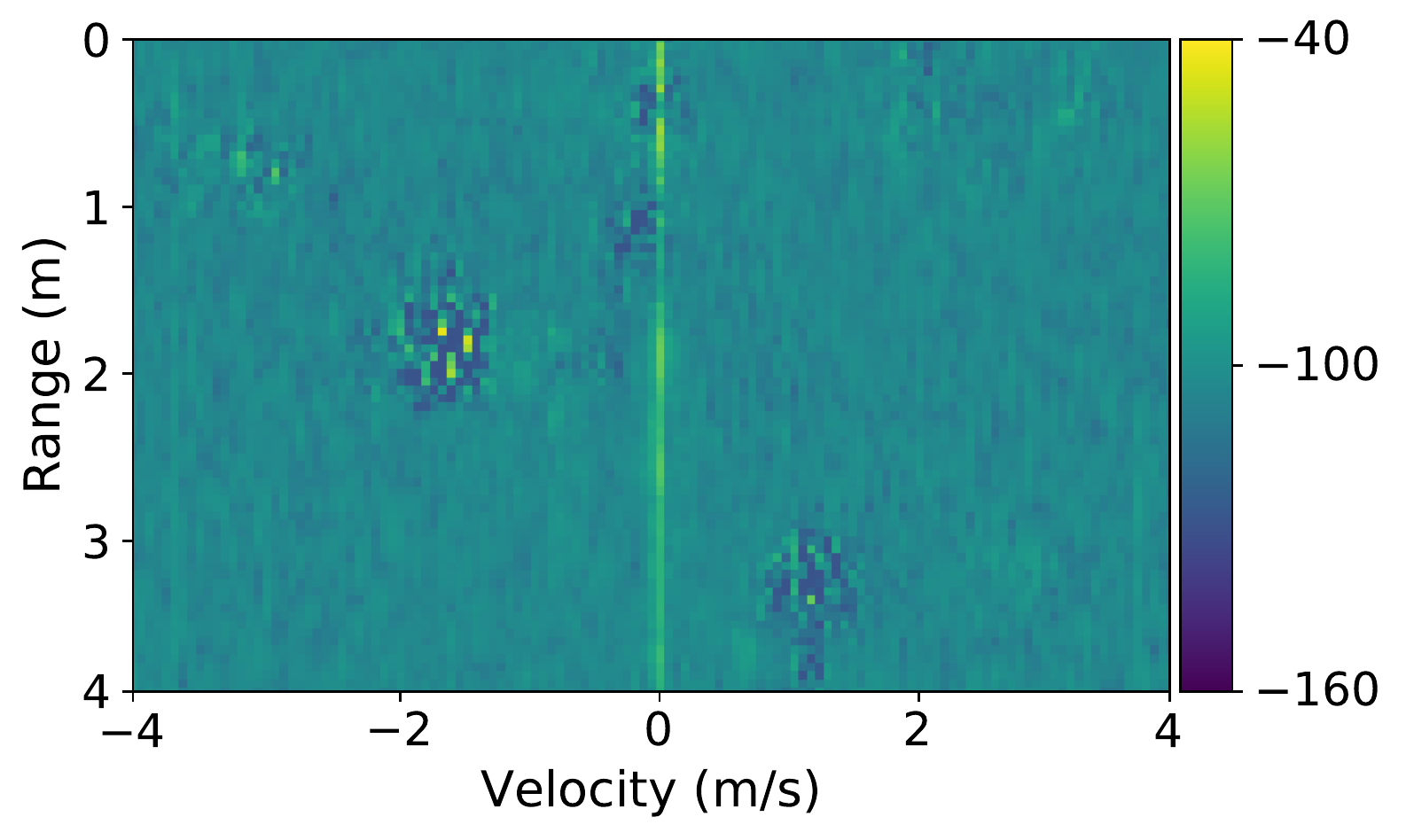}}}
\\
\rotatebox[origin=l]{90}{\phantom{-----}$L_2$ Distance: $3.457$}
\hspace{0.5em}
\subfloat{{\includegraphics[width=6cm, height=3.5cm]{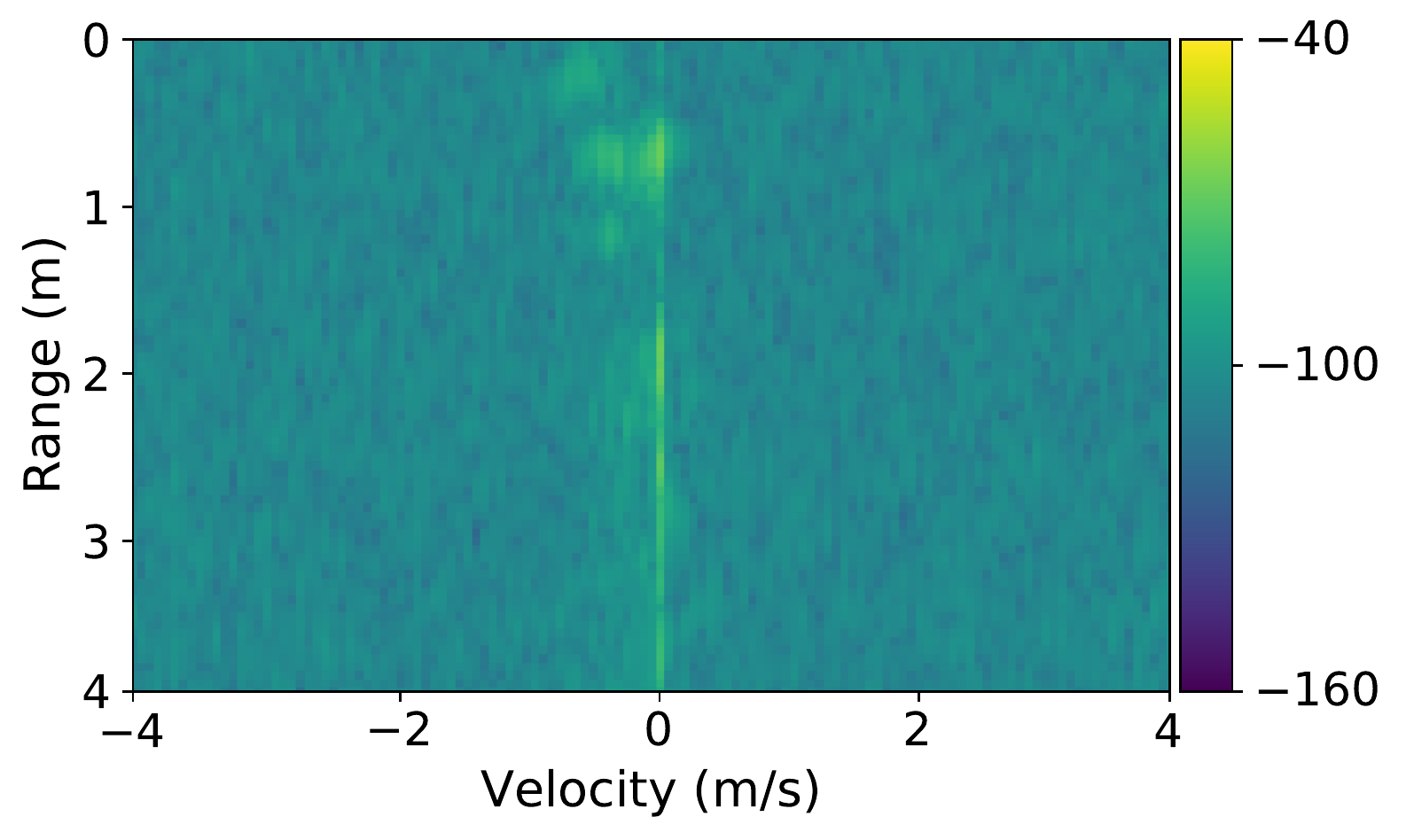}}}
\subfloat{{\includegraphics[width=6cm, height=3.5cm]{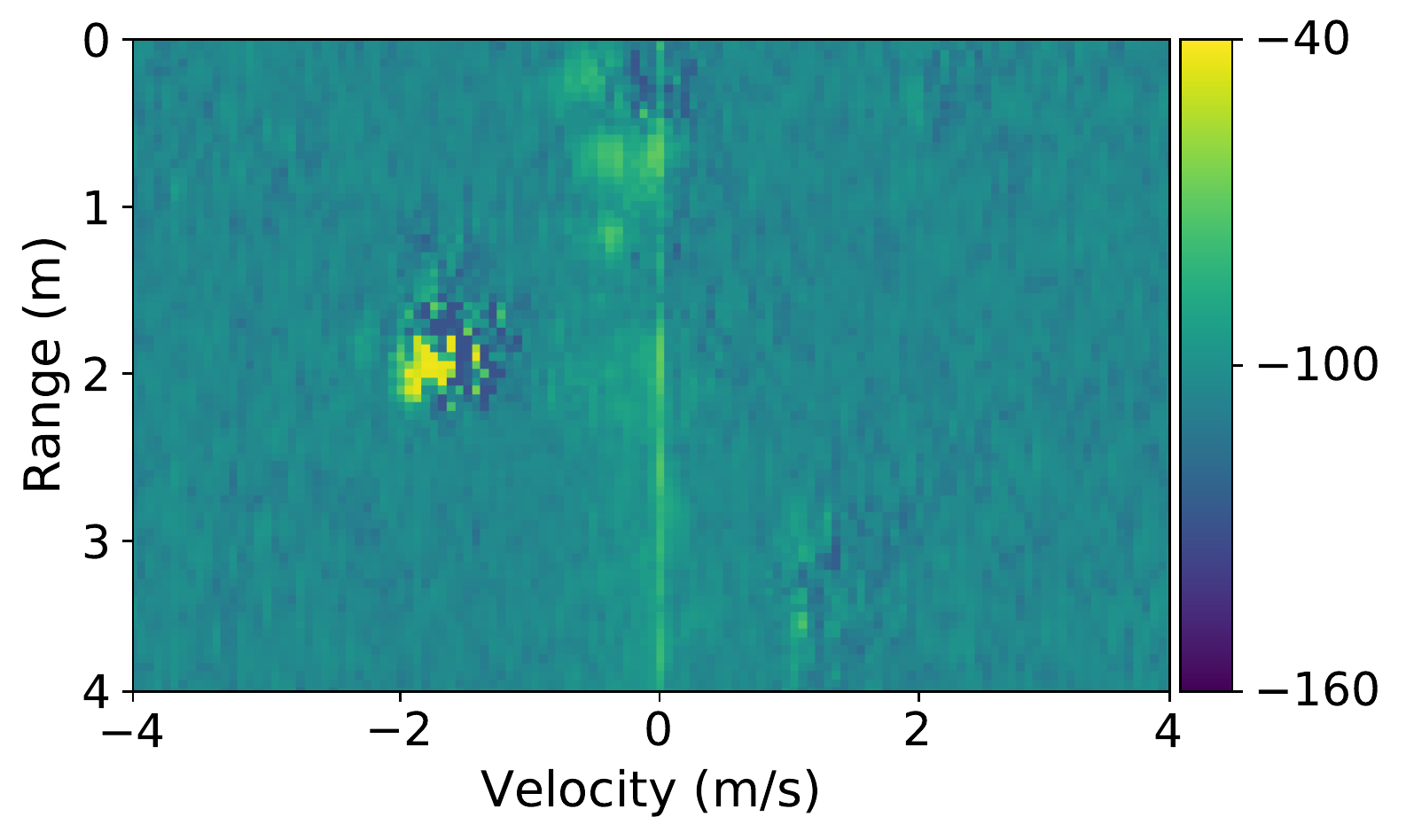}}}
\caption{(1) $L_2$ distances, (2) original frames, and (3) perturbed frames are given in order to improve the visual understanding of various degrees of perturbation.}
\label{fig:l2_comparison}
\centering
\end{figure*}

\textbf{$L_2$ distance and perturbation visibility }\,\textemdash\,In order to provide a better visual understanding of the $L_2$ distance comparisons made in the main text, we provide a detailed comparison of the $L_2$ distances between the original and the perturbed frames in Fig.~\ref{fig:l2_comparison}. 

The first two examples provided in Fig.~\ref{fig:l2_comparison} are examples taken from adversarial optimizations that we evaluated in our study. The latter three are extreme cases, characterized by excessive application of perturbation (i.e., the changes are becoming visible to the bare eye). Specifically, the last two examples show that, when the adversarial optimization is not controlled properly, the generated adversarial noise is concentrated in localized regions of the image at hand. The reason we call the last three examples \textit{extreme} is because the $L_2$ distance is calculated for a single frame. However, $L_2$ distances provided in the main text are calculated from $50$ frames. As a result, even though an $L_2$ perturbation of $3.45$ corresponds to visually identifiable perturbation for a single frame, when this perturbation is spread over $50$ frames, it becomes much less identifiable. For example, if this amount of perturbation is spread over $30$ frames (all of the frames that contain the action are perturbed, all of the padding frames are not perturbed), then an $L_2$ perturbation of $3.45$ becomes, on average, an $L_2$ perturbation of $0.115$ per frame. 



\newpage

\end{document}